\pdfoutput=1

\documentclass[11pt]{article}

\usepackage{EMNLP2023}

\usepackage{times}
\usepackage{latexsym}

\usepackage[T1]{fontenc}

\usepackage[utf8]{inputenc}

\usepackage{microtype}

\usepackage{inconsolata}

\usepackage{graphicx}
\usepackage{algorithm,algorithmic}
\usepackage{booktabs} 
 
\usepackage{hyperref}

\usepackage{natbib}

\usepackage{amsmath}
\usepackage{amssymb}
\usepackage{mathtools}
\usepackage{amsthm}
\allowdisplaybreaks

\usepackage{mathrsfs}

\usepackage[capitalize,noabbrev]{cleveref}

\theoremstyle{plain}
\newtheorem{theorem}{Theorem}[section]

\newtheorem{hypothesis}[theorem]{Hypothesis}

\theoremstyle{definition}

\theoremstyle{remark}

\usepackage[textsize=tiny]{todonotes}

\usepackage{hyperref}       
\usepackage{cleveref}       
\usepackage{url}            
\usepackage{booktabs}       
\usepackage{nicefrac}       
\usepackage{microtype}      
\usepackage{xcolor}         

\usepackage{microtype}
\usepackage{graphicx}
\usepackage{makecell}
\usepackage{caption}
\usepackage{subcaption}
\usepackage{multirow}
\usepackage{rotating}

\usepackage{colortbl}
\definecolor{bgcolor}{rgb}{0.66,0.88,1.00}

\usepackage{tablefootnote}
\usepackage{threeparttable}

\usepackage{xspace}

\xspaceaddexceptions{[]\{\}}

\newcommand{\dataset}[1]{{\tt #1}\xspace}

\usepackage{thm-restate}

\usepackage{eqparbox}

\usepackage{etoolbox}
\usepackage{enumitem}
\patchcmd{\quote}{\rightmargin}{\leftmargin 1em \rightmargin}{}{}

\makeatletter
\allowdisplaybreaks

\newcommand{\dotp}[2]{\left\langle {#1}, {#2}, \right\rangle}
\newcommand{\norm}[1]{\left\|{#1}\right\|}

\usepackage{amsmath,amsfonts,bm}

\def\eqref#1{equation~\ref{#1}}

\def\1{\bm{1}}

\def\vmu{{\bm{\mu}}}

\def\ve{{\bm{e}}}

\def\vv{{\bm{v}}}

\def\vx{{\bm{x}}}

\def\mA{{\bm{A}}}

\def\mE{{\bm{E}}}

\def\mI{{\bm{I}}}

\def\mK{{\bm{K}}}

\def\mM{{\bm{M}}}

\def\mP{{\bm{P}}}
\def\mQ{{\bm{Q}}}

\def\mV{{\bm{V}}}
\def\mW{{\bm{W}}}
\def\mX{{\bm{X}}}

\DeclareMathAlphabet{\mathsfit}{\encodingdefault}{\sfdefault}{m}{sl}
\SetMathAlphabet{\mathsfit}{bold}{\encodingdefault}{\sfdefault}{bx}{n}

\def\gG{{\mathcal{G}}}

\def\gI{{\mathcal{I}}}

\def\gN{{\mathcal{N}}}

\def\gP{{\mathcal{P}}}

\def\gT{{\mathcal{T}}}

\newcommand{\R}{\mathbb{R}}

\newcommand{\haoyu}[1]{{\color{orange}[HZ: #1]}}

\newcommand{\rong}[1]{{\color{green}[RG: #1]}}

\title{Do Transformers Parse while Predicting the Masked Word?}
\author{
Haoyu Zhao\textsuperscript{1,2}\footnotemark[1]\quad Abhishek Panigrahi\textsuperscript{1,2}\footnotemark[1]\quad Rong Ge\textsuperscript{3}\quad Sanjeev Arora\textsuperscript{1,2} \\
\textsuperscript{1}Department of Computer Science, Princeton University \\
\textsuperscript{2} Princeton Language and Intelligence, Princeton University \\
\textsuperscript{3}Department of Computer Science, Duke University \\
\texttt{\{haoyu,ap34,arora\}@cs.princeton.edu, rongge@cs.duke.edu}
}
\date{}

\begin{document}

\maketitle

\renewcommand{\thefootnote}{\fnsymbol{footnote}}
\footnotetext[1]{Equal contribution.}

\begin{abstract}

\looseness=-1 Pre-trained language models have been shown to encode linguistic structures like parse trees in their embeddings while being trained unsupervised. Some doubts have been raised whether the models are doing parsing or only some computation weakly correlated with it. Concretely: (a) Is it possible to explicitly describe transformers with realistic embedding dimensions, number of heads, etc. that are {\em capable} of doing parsing ---or even approximate parsing? (b) Why do pre-trained models capture parsing structure? 
This paper takes a step toward answering these questions in the context of generative modeling with PCFGs. We show that masked language models like BERT or RoBERTa of moderate sizes can approximately execute the Inside-Outside algorithm for the English PCFG~\citep{marcus1993building}. We also show that the Inside-Outside algorithm is optimal for masked language modeling loss on the PCFG-generated data. 
We conduct probing experiments on models pre-trained on PCFG-generated data to show that this not only allows recovery of approximate parse tree, but also recovers marginal span probabilities computed by the Inside-Outside algorithm, which suggests an implicit bias of masked language modeling towards this algorithm.

\end{abstract}

\section{Introduction}
\looseness=-1One of the surprising discoveries about transformer-based language models like BERT~\citep{devlin2019bert} and RoBERTa~\citep{liu2019roberta} was that contextual word embeddings encode information about parsing, which can be extracted using a simple ``linear probing''  to yield approximately correct  dependency parse trees for the text~\citep{hewitt2019structural,manning2020emergent}.
Subsequently, \citet{vilares2020parsing,wu2020perturbed,arps2022probing} employed linear probing also to recover information about constituency parse trees.
Investigating the parsing capability of transformers is of significant interest, as incorporating (the awareness of) syntax in large language models has been shown to enhance the final performance on various downstream tasks~\citep{xu2021syntax,bai2021syntax}. Additionally, it can contribute to the ongoing exploration of the ``mechanistic interpretability'' for reverse engineering the inner workings of pre-trained large language models~\citep{elhage2021mathematical,olsson2022context,nanda2023progress}.

\looseness=-1The current paper focuses on the ability of BERT-style transformers to do constituency parsing, specifically for PCFGs. Prior studies~\citep{bhattamishra2020computational,perez2021attention} established that transformers are Turing complete, suggesting their potential for parsing. But do they actually parse while trying to do masked-word prediction?  
One reason to be cautiously skeptical is that naive translation of constituency parsing algorithms into a transformer results in transformers with number of heads that scales with the size of the grammar (\Cref{sec:construct-io}), whereas BERT-like models have around a dozen heads. This leads to the following question.  

\begin{quote}
    \centering
    {\em (Qs 1): Are BERT-like models capable of parsing with realistic number of heads?}
\end{quote}

\looseness=-1 This is not an idle question as \citet{maudslay2021syntactic} suggested that linear probing relies on semantic cues for parsing. They created syntactically correct but semantically meaningless sentences and found a significant drop in parsing performance compared to previous studies.  
\begin{quote}
\centering
    {\em (Qs 2): Do BERT-like models trained for masked language modeling (MLM) encode syntax, and if so, how and why?}
\end{quote}

\subsection{This paper}

\looseness=-1To address Qs 1, we construct a transformer that executes the Inside-outside algorithm for PCFG (\Cref{sec:construct-io}). If the PCFG has $N$ non-terminals and the length of the sentence is $L$, our constructed transformer has $2L$ layers in total, $N$ attention heads, and $2NL$ embedding dimensions in each layer. However, this is massive compared to BERT.
For PCFG learned on Penn Treebank (\dataset{PTB})~\citep{marcus1993building}, $N=1600$, average $L \approx 25$, which leads to a transformer with $80$k embedding dimension, depth $50$, and $1.6$k attention heads per layer. By contrast, BERT has $768$ embedding dimensions, $12$ layers, and $12$ attention heads per layer! 

\looseness=-1 One potential explanation could be that BERT does not do exact parsing but merely computes {\em some} information related to parsing. After all, linear probing didn't recover complete parse trees. It recovered trees with modest F1 score, such as $78.2\%$ for BERT~\citep{vilares2020parsing} and $82.6\%$ for RoBERTa~\citep{arps2022probing}. 
To the best of our knowledge, no study has investigated parsing methods that  strategically discard information to do more efficient approximate parsing. Toward this goal, we design an approximate version of the Inside-Outside algorithm (\Cref{sec:approx-overview}), executable by a transformer with $2L$ layers, $15$ attention heads, and $40L$ embedding dimensions, while still achieving $>70\%$ F1 score for constituency parsing on \dataset{PTB} dataset~\citep{marcus1993building}. 

\looseness=-1 Although realistic models can capture a fair amount of parsing information, it is unclear whether they need to do so for masked language modeling (MLM). After all, \citet{maudslay2021syntactic} suggested that linear probing picks up on semantic information that happens to correlate with parse trees.
To further explore this, we trained a (masked) language model on the synthetic text generated from a PCFG tailored to English text, separating syntax from semantics in a more rigorous manner than \citet{maudslay2021syntactic}.
\Cref{sec:mlmandio} notes that given such synthetic text, the Inside-Outside algorithm will minimize MLM loss. 
Note that parsing algorithms like CYK~\citep{kasami1966efficient} could be used instead of Inside-Outside, but they do not have an explicit connection to MLM (\Cref{sec:mlmandio}).
Experiments with pre-trained models on synthetic PCFG data (\Cref{sec:pretrain-pcfg}) reveal the existence of syntactic information inside the models: simple probing methods recover reasonable parse tree structure (\Cref{sec:parse}).
Additionally, probes of contextualized embeddings reveal correlations with the information computed by the Inside-Outside algorithm (\Cref{sec:probe-marginal-probs}).  This 
suggests transformers implicitly engage in a form of approximate parsing, in particular a process related to the Inside-Outside algorithm, to achieve low MLM loss.

\section{Preliminaries}\label{sec:preliminary}

\subsection{Attention}

\looseness=-1 We focus on encoder-only transformers like BERT and RoBERTa~\citep{devlin2019bert,liu2019roberta}, which stack identical layers with an attention module followed by a feed-forward module. Each attention module has multiple heads, represented by three matrices $\mQ_{h}, \mK_{h}, \mV_{h} \in \mathbb{R}^{d \times d}$.
For an input sequence of length $L$, we use $\mE^{(\ell)} \in \R^{L \times d}$ to denote contextual embeddings after layer $\ell$'s computations, where $\ve_i^{(\ell)}$ is the embedding of the $i^{th}$ token. The output of the attention head $h$ at layer $\ell$ is $\vv^{(\ell)}_{i, h} = \sum_{j \in [L]} a^h_{i, j} \mV_h \ve^{(\ell)}$, where $a^h_{i, j}$ is the attention score between $\ve_i$ and $\ve_j$ for head $h$:

{\small
\vspace{-1mm}
\begin{align}
    a^h_{i, j} = f_{\text{attn}} ( \mE^{(\ell)} \mK_h^{\top}, \mQ_h \ve_i^{(\ell)} )_j \label{def:attention}.
\end{align}
}

\looseness=-1 $f_{attn}$ is a non-linear function and is generally used as softmax on $\mE^{(\ell)} \mK_h^{\top} \mQ_h \ve_i^{(\ell)}$. Finally, the output of the attention module is given by $\sum_h \vv^{(\ell)}_{i, h}.$ This is a general definition of the attention module and captures the split and merge of the embeddings across the attention heads used in practice.



\subsection{PCFG and parsing} \label{sec:pcfg_def}
\looseness=-1 \paragraph{PCFG model} A probabilistic context-free grammar (PCFG) is a language generative model. It is defined as a 5-tuple $\gG = (\gN, \gI, \gP, n, p)$, where
\begin{itemize}
[leftmargin=*]
\setlength\itemsep{0.05em}
    \item $\gN$ is the set of non-terminal. $\gI,\gP\subset\gN$ are sets of \emph{in-terminals} and \emph{pre-terminals} respectively. $\gN = \gI\cup\gP$, and $\gI\cap\gP = \phi$.
    \item $[n]$ is the set of all possible words.
    \item $\forall A\in\gI, B,C\in\gN$, there is a rule $A\to BC$.
    \item For rule $A\to BC$ where $A\in\gI, B, C\in\gN$, there is a probability $\Pr[A\to BC]$ satisfying for all $A$, $\sum_{B,C}\Pr[A\to BC] = 1$.
    \item For all $A\in\gP, w\in [n]$, a rule $A\to w$.
    \item For each rule $A\to w$ where $A\in\gP, w\in [n]$, a probability $\Pr[A\to w]$, which satisfies for all $A$, $\sum_{w}\Pr[A\to w] = 1$.
    \item A non-terminal $\text{Root}\in\gI$. 
\end{itemize}

\looseness=-1 \paragraph{Data generation from PCFG} Strings are generated from the PCFG $\gG = (\gN, \gI, \gP, n, p)$ as follows: we maintain a string $s_t \in ([n]\cup\gN)^*$ at step $t$ with $s_1 = \text{ROOT}$. At step $t$, if all characters in $s_t$ belong to $[n]$, the generation process ends, and $s_t$ is the resulting string. Otherwise, we pick a character $A\in s_t$ such that $A\in\gN$. If $A\in\gP$, we replace the character $A$ to $w$ with probability $\Pr[A\to w]$. If $A\in\gI$, we replace the character $A$ to two characters $B, C$ with probability $\Pr[A\to BC]$.

\looseness=-1 \paragraph{Parse trees and parsing} For a sentence $s = w_1\dots w_L$ with length $L$, a labeled parse tree represents the likely derivations of a sentence under PCFG $\gG$. It is defined as a list of spans with non-terminals $\{(A, i, j)\}$ that forms a tree. An unlabelled parse tree is a list of spans that forms a tree. 

\looseness=-1To find the unlabelled parse tree for a sentence $s$ under the PCFG model, the Labelled-Recall algorithm~\citep{goodman1996parsing} is commonly used. This algorithm searches for the tree $T = \{(i,j)\}$ that maximizes $\sum_{(i,j)\in T} \mathrm{score}(i,j)$, where $\mathrm{score}(i,j)=\max_{A\in \gN} \Pr[A\Rightarrow w_iw_{i+1} \cdots w_j, \text{Root}\Rightarrow s|\gG]:=\max_{A\in \gN}\mu(A, i, j)$ is the marginal probability of span $w_iw_{i+1} \cdots w_j$ under non-terminal $A$.

\looseness=-1Marginal probabilities are computed by Inside-Outside algorithm~\citep{baker1979trainable}, with the inside probabilities $\alpha(A, i,j)$
and the outside probabilities $\beta(A, i, j)$
computed by the following recursion

{\small
\begin{align}
    & \alpha(A,i,j) \nonumber \\
    & = \sum_{B,C}\sum_{k=i}^{j-1} \Pr[A\to BC]\alpha(B,i,k)\alpha(C,k+1,j), \label{eq:inside_probability} \\
    & \beta(A,i,j) \nonumber \\ 
    &= \sum_{B, C} \sum_{k=1}^{i-1}\Pr[B \to C A] \alpha(C, k, i-1) \beta(B, k, j) \label{eq:outside_probability} \\  & \quad + \sum_{B, C} \sum_{k=j+1}^{L}\Pr[B \to A C] \alpha(C, j+1, k) \beta(B, i, k) \nonumber
\end{align}
}

\looseness=-1 with the base cases $\alpha(A,i,i) = \Pr[A\to w_i]$ for all $A,i$ and $\beta(\text{Root},1,L)=1$ for all $A$. The marginal probabilities are then computed as
{\small
\begin{equation}\label{eq:marginal_probability}
   \mu(A,i,j) = \alpha(A,i,j)\times\beta(A,i,j). 
\end{equation}
}

\looseness=-1 Parsing performance is evaluated by two types of unlabelled F1 scores, which depend on the average method: Sentence F1 (average of F1 scores for each sentence) and Corpus F1 (considers total true positives, false positives, and false negatives).

\subsection{Probing}

\looseness=-1 A probe $f(\cdot)$ is a supervised model that predicts a target $\text{tar}(\vx)$ for a given input $\vx$ \cite{alain2017understanding,hupkes2018visualisation,conneau-etal-2018-cram}. As an example, \citet{hewitt2019structural} used a probe $f(\cdot)$ to predict the tree distance $\text{tar}(i,j) = d_{\gT}(i,j)$ between words in a dependency parse tree $\gT$. 
Although mathematically equivalent, probes and supervised models have different goals. The latter aims for high prediction scores, while the former seeks to identify certain intrinsic information in embeddings~\citep{maudslay2020tale,chen2021probing}. Probes should be limited to only detect the desired information, with low performance on uncontextualized embeddings and high performance on contextualized ones.

\section{Parsing using Transformers}\label{sec:construction}
\looseness=-1We design transformers with moderate layers and heads for parsing and masked language modeling. In \Cref{sec:construct-io}, we prove that transformers can execute the Inside-Outside algorithm for bounded-length sentences with any PCFG. In \Cref{sec:mlmandio}, we connect our construction with masked language modeling and demonstrate the optimality of the Inside-Outside algorithm for MLM on PCFG-generated data. Finally, in \cref{sec:approx-overview}, we demonstrate the ability to reduce the size of these constructions while retaining their parsing performance.

\subsection{Transformers can execute Inside-Outside algorithm}\label{sec:construct-io}

\looseness=-1 We first give a construction (\cref{thm:hard_attnt}) that relies on {\em hard attention}, where only one of the attended positions will have positive attention score. For this construction,
we define $f_{\text{attn}}: \mathbb{R}^{L \times d} \times \mathbb{R}^{d}$ such that the attention scores in eq.~\ref{def:attention} are given by

{
\small
\begin{align}
    a^h_{i, j} = \text{ReLU} (  (\mK_h \ve_j^{(\ell)})^{\top} \mQ_h \ve_i^{(\ell)} ). \label{eq:hard_attention} 
\end{align}
}

This is similar to softmax attention used in practice, with softmax replaced by $\text{ReLU}$ activation. 


\begin{theorem}[Hard attention]\label{thm:hard_attnt}
    There exists a model with hard attention modules (\ref{eq:hard_attention}), $(4|\gN| + 1)L$ embeddings, $2L-1$ layers, and $4|\gN|$ attention heads in each layer that simulates the Inside-Outside algorithm on all sentences with length at most $L$ generated by PCFG $\gG = (\gN, \gI, \gP, n, p)$ and embed all inside and outside probabilities.
\end{theorem}

\looseness=-1 \begin{proof}[Proof sketch] We give the proof sketch and defer details to \cref{sec:hard_attnt_proof}. The core idea is to use the first $L$ layers to compute the inside probabilities with the recursive eq. \ref{eq:inside_probability}. Each layer $\ell \le L$ computes $\alpha(A,i,j)$ for all position pairs $(i, j)$ with $j-i = \ell$ and all non-terminals $A$. The next $L$ layers compute the outside probabilities with the recursive eq. \ref{eq:outside_probability}. Each layer $L+\ell > L$ computes $\beta(A,i,j)$ for all position pairs $(i, j)$ with $j-i=L-\ell$ and all non-terminals $A$. 

\looseness=-1 At any position $i$ in a layer $\ell \le L$, the input token embeds inside probabilities of all spans with a maximum length of $\ell$, starting and ending at $i$: $\alpha(A, i, j)$ and $\alpha(A, k, i)$ for all non-terminals $A$ and position tuples $(i, j, k)$ where $j-i<\ell$, $i-k<\ell$. To compute $\alpha(A, i, i+\ell)$ at each position $i$ for each non-terminal $A$, we use an attention head that calculates an inner product between the embeddings at positions $i$ and $i+\ell$, weighted by the matrix containing ${\Pr[A \to BC]}_{B, C \in \gN}$. The token at position $i$ attends only to the token at $i+\ell$ thanks to the position embeddings and hard attention. We use another attention head to compute $\alpha(A, i-\ell, i)$, and store the new inside probability terms along with the previous ones in the embeddings. We use a similar technique to compute the outside probabilities in the next $L$ layers. In layer $L+\ell$, we use two attention heads to compute $\beta(A, i, i+L-\ell)$ for each non-terminal $A$ and position $i$, as there are two terms to compute in \ref{eq:outside_probability}. We use two additional attention heads to compute $\beta(A, i-L+\ell, i)$, resulting in four attention heads for each non-terminal.
\end{proof}

\looseness=-1 To further reduce embedding size and attention heads, we introduce relative positions and use soft attention. We introduce $2L + 1$ relative position vectors $ \{ p_{ t } \in \mathbb{R}^d \}_{-L \le t \le L},$ and relative position biases $\{ b_{t, \ell} \in \mathbb{R} \}_{-L \le t \le L, 1 \le \ell \le 2L-1}$ that modify the key vectors depending on the relative position of the query and key tokens. For an attention head $h$ in layer $\ell$, the attention score $a_{i, j}^h$ is given by

{
\small
\begin{equation}
    a_{i, j}^h =  \text{ReLU}( \mK_h \ve_j^{(\ell)} + p_{j - i} - b_{j-i, \ell} )^\top \mQ_h \ve_i^{(\ell)}. \label{eq:soft_attention}
\end{equation}
}

\begin{theorem}[Relative positional embeddings] \label{thm:soft_attnt}
    There exists a model with attention module (\ref{eq:soft_attention}), $2|\gN| L + 1$ embeddings, $2L-1$ layers, and $|\gN|$ attention heads in each layer that simulate the Inside-Outside algorithm on all sentences with length at most $L$ generated by PCFG $\gG = (\gN, \gI, \gP, n, p)$ and embed all inside and outside probabilities.
\end{theorem}

\looseness=-1 The proof is deferred to \cref{sec:soft_attnt_proof}. \Cref{thm:soft_attnt} uses one attention head to compute layer-wise inside/outside probabilities per non-terminal, and only requires $|\gN|$ heads in each layer. 
Once we have the inside and outside probabilities for spans, we can directly build the parse tree using the Labelled-Recall algorithm, which acts as a ``probe'' on the contextual representations of the model.

\subsection{Masked language modeling for PCFG}\label{sec:mlmandio}

\looseness=-1 The Inside-Outside algorithm not only can parse but also has a connection to masked language modeling (MLM), the pre-training loss used by BERT. The following theorem shows that, if the language is generated from a PCFG, then the Inside-Outside algorithm achieves the optimal MLM loss.

\begin{theorem}\label{thm:io-optimal-mlm}
    Assuming language is generated from a PCFG, the Inside-Outside algorithm reaches the optimal MLM loss.
\end{theorem}

\looseness=-1The Inside-Outside algorithm optimizes MLM loss on PCFG data, suggesting that pre-training on such data enables implicit learning of the algorithm  or its computed quantities. Consequently, intermediate layers can capture syntactic information for parsing, potentially explaining the presence of structural information in language models~\citep{hewitt2019structural,vilares2020parsing,arps2022probing}. We validate this conjecture in \Cref{sec:probe-marginal-probs}.

\subsection{Towards realistic size}\label{sec:approx-overview}

\looseness=-1 For PCFG learned on the \dataset{PTB} training set (\dataset{PTB} sections 02-21) with an average sentence length of 25~\citep{Spectral-Parser}, \Cref{sec:construct-io} requires $~1600$ attention heads, $~3200L$ embedding dimensions, and $2L$ layers to simulate the Inside-Outside algorithm for sentences of length $L$, which is much larger than BERT.
However, by utilizing the inherent sparsity in the English PCFG, we can reduce the number of attention heads and the width of the embeddings while maintaining decent parsing performance. The details are deferred to \Cref{sec:approx-detailed}.


\paragraph{First ingredient: finding important non-terminals}
\looseness=-1 In the constructions of \cref{thm:hard_attnt,thm:soft_attnt}, the number of attention heads and embedding dimensions depend on the number of non-terminals of the PCFG. Thus if we can find a smaller PCFG, we can make the model much smaller. Specifically, if we only compute the probabilities of a specific set of in-terminals $\tilde\gI$ and pre-terminals $\tilde\gP$ in eq.~\ref{eq:inside_probability} and~\ref{eq:outside_probability}, we can reduce the number of attention heads from $|\gN|$ to $\max\{|\tilde\gI|,|\tilde\gP|\}$.\footnote{When $|\tilde\gP| < c|\tilde\gI|$, we can simulate the computations in the final layer using $c$ layers with $|\tilde\gI|$ heads instead of $|\tilde\gP|$ heads. Additionally, we can decrease the embedding size by only storing probabilities for relevant non-terminals.}

\begin{table}[!t]
    \centering
    \footnotesize
    \begin{tabular}{|c|c|c|c|}
    \hline
         Approximation & Corpus F1 & Sent F1 & ppl. \\
         \hline
         \makecell{No approx.} & 75.90 & 78.77 & 50.80 \\
         \hline
         $|\tilde\gI| = 10,|\tilde\gP|=45$ & 57.14 & 60.32 & 59.57 \\
         $|\tilde\gI| = 20,|\tilde\gP|=45$ & 68.41 & 71.91 & 55.16 \\
         $|\tilde\gI| = 40,|\tilde\gP|=45$ & 72.45 & 75.43 & 54.09 \\
         \hline
    \end{tabular}
    \caption{Restricting computations of the Inside-Outside algorithm to the most frequent in(pre)-terminal subsets $\tilde\gI$ ($\tilde\gP$) in the \dataset{PTB} sections 02-21. We report the unlabelled F1 scores on \dataset{PTB} section 22 and the 1-masking perplexity on 200 sentences generated from the PCFG. 
    $|\tilde \gI|=20, |\tilde \gP|=45$ resulted in a $8.58\%$ increase in perplexity and $8.71\%$ decrease in parsing F1 scores.
    }
    \label{tab:few-nt-pcfg-global}
\end{table}

\looseness=-1 We sort the non-terminals in terms of their frequency of occurrence in the \dataset{PTB} training set and show that restricting the Inside-Outside computation to a few frequent non-terminals has a negligible drop in performance (\cref{tab:few-nt-pcfg-global}). The parsing score is still highly non-trivial, since the naive baseline, Right Branching (RB), can only get $<40\%$ sentence and corpus F1 scores on \dataset{PTB} dataset.

\paragraph{Second ingredient: utilizing structures across non-terminals}
\looseness=-1 We still use one attention head to represent the computation for a specific non-terminal, which does not utilize possible underlying correlations between different non-terminals.
Specifically, for \cref{thm:soft_attnt}, we use one attention head at layer $\ell < L$ to compute the inside probabilities $\alpha(A, i,j)$ with $j-i = \ell$. 
If $\alpha(A,i,j)$ for different non-terminals $A\in\tilde\gI$ lie in a $k^{(\ell)}$-dimensional subspace with $k^{(\ell)} < |\tilde\gI|$, we can compute all of the inside probabilities using only $k^{(\ell)}$ attention heads by computing the vector $\mW^{(\ell)}\bm{\alpha}(i,j)$, where $\mW^{(\ell)}\in\R^{k^{(\ell)}\times |\tilde\gI|}$ is the transformation matrix and $\bm{\alpha}(i,j)\in\R^{|\tilde\gI|}$ is the concatenation of all inside probabilties ${\alpha(A,i,j)}_{A\in\tilde\gI}$. The same procedure can also be applied to the computation of outside probabilities.
\footnote{The computation for $A\in\tilde\gP$ needs $|\tilde\gP|$ heads in the last layer and can be simulated by several layers with fewer heads.} 
Although the probabilities should not lie in a low dimensional subspace in reality, we can still try to learn a transformation matrix $\mW^{(\ell)}\in\R^{k^{(\ell)}\times |\tilde\gI|}$ and approximately compute the inside probabilities by $\bm\alpha(i,j) = (\mW^{(\ell)})^{\dagger}\mW^{(\ell)}\bm\alpha^*(i,j)$ for $j-i = \ell$, where $\bm\alpha^*(i,j)$ denotes the Inside probabilities for non-terminals in $\tilde\gI$. 
Please refer to \Cref{sec:approx-low-rank} for more details.

 \looseness=-1 \subparagraph{Learning the transformations} 
 For sentence $s$ and a span with length $\ell+1$, we compute the marginal probabilities of this span $\vmu_s^{i,j}\in\R^{|\tilde\gI|}$, that contains $\mu(A,i,j)$ for each non-terminal $A\in\tilde\gI$. 
We then compute the normalized correlation matrix
$\mX^{(\ell)} = \sum_{s} \mX_s^{(\ell)} / \|\mX_s^{(\ell)}\|_{\text{F}}$, where $\mX_s^{(\ell)} = \sum_{i,j:j-i=\ell} \vmu_s^{i,j}(\vmu_s^{i,j})^\top$, which captures the correlation of $\tilde\gI$ for spans with length $\ell+1$ in the entire corpus.
We apply the Eigen-decomposition on $\mX_\ell$ and set $\mW^{(\ell)}$ as the top $k^{(\ell)}$ Eigen-vectors. 

\looseness=-1 The parsing results and 1-masking perplexity using $\{\mW^{(\ell)}\}_{\ell\le L}$ with different $k^{(\ell)}$ are shown in \Cref{tab:learned-transformation-global}. Utilizing the linear transformations, we obtain $71.33\%$ and $65.31\%$ sentence F1 on \dataset{PTB} with only 15 and 10 attention heads respectively, whereas only computing probabilities for top-$10$ in-terminals gives $60.32\%$ sentence F1 on \dataset{PTB}. The following theorem summarizes the results.

\begin{table}[!t]
    \centering
    \scriptsize
    \begin{tabular}{|c|c|c|c|}
        \hline
        Approximation & Corpus F1 & Sent F1 & ppl. \\ 
        \hline
        \makecell{$|\tilde\gI| = 10,|\tilde\gP|=45$} & 57.14 & 60.32 & 59.57 \\
        $|\tilde\gI| = 20,|\tilde\gP|=45$ & 68.41 & 71.91 & 55.16 \\
        \hline
        \makecell{$k^{(\ell)} = 10,|\tilde\gI| = 20,|\tilde\gP|=45$} & 61.72 & 65.31 & 57.05 \\
        \makecell{$k^{(\ell)} = 15,|\tilde\gI| = 20,|\tilde\gP|=45$} & 68.20 & 71.33 & 55.52 \\
        \hline
    \end{tabular}
    \caption{Approximate Inside-Outside algorithm using linear transformations $\{\mW^{(\ell)} \in \mathbb{R}^{k^{(\ell)} \times |\tilde\gI|} \}$ on the inside/outside probabilities of the selected subset $\tilde{\gI}$. We report the F1 scores on \dataset{PTB} section 22 and the 1-masking perplexity on 200 sentences generated from the PCFG. Applying linear transformations can further reduce the number of attention heads in the constructed model to $15$ starting from $20$ frequent non-terminals subset $\tilde\gI$, while only changing the performance by at most $1\%$. 
    }
    \label{tab:learned-transformation-global}
\end{table}


\begin{theorem}[Informal]\label{thm:approx-low-rank-informal}
    There exists a model with attention module (\ref{eq:soft_attention}), $275+40L$ embeddings, $2L+1$ layers, and $15$ attention heads in each layer that can approximately execute  Inside-Outside algorithm on all sentences with length at most $L$ generated by English PCFG, introducing $8.6\%$ increase in average 1-mask perplexity and resulting in at most $9.45\%$ drop in the parsing performance of Labeled-Recall algorithm.
\end{theorem}

\section{Probing Masked Language Models for Parsing Information}\label{sec:mlmtoparsing}

\looseness=-1 \cref{sec:construction} shows that transformers can execute the Inside-Outside algorithm and contain syntactic information in their intermediate states. These results are existential, and it is unclear if models pre-trained under MLM possess similar information. 

\looseness=-1 One difficulty in answering this question is that syntactic probes on BERT-like models may leverage semantic cues to parse. To address this concern, we pre-train multiple RoBERTa models on synthetic datasets derived from English PCFG (\Cref{sec:pretrain-pcfg}), which eliminates semantic dependencies. We then probe the models for parse tree construction (\Cref{sec:parse}) and marginal probabilities (\Cref{sec:probe-marginal-probs}) to verify if they capture information computed by the Inside-Outside algorithm.

\subsection{Pre-training on PCFG}\label{sec:pretrain-pcfg}
\looseness=-1 We pre-train RoBERTa models with varying attention heads and layers on synthetic PCFG data. We denote the models with A$i$L$j$, where $i$ and $j$ indicate the number of attention heads and layers, respectively. Additional pre-training details are available in~\Cref{sec:pretraining-details}. \Cref{tab:pretraining-ppl} shows the perplexity for various models. We find that except for models with too few layers (A12L1) and too few attention heads (A3L12), other models have nearly the same perplexity. Further increasing depth and number of heads does not appear to improve the result.

\begin{table}[!t]
    \centering
    \footnotesize
    \begin{tabular}{|c|c|c|}
        \hline
        Model & Training ppl. & Validation ppl. \\
        \hline
        A12L12 & 106.16 & 106.68 \\
        A12L1 & 111.8 & 110.57 \\
        A12L3 & 108.09 & 105.79 \\
        A12L6 & 105.78 & 104.58 \\
        A3L12 & 120.52 & 117.39 \\
        A24L12 & 106.28 & 104.5 \\
        \hline
    \end{tabular}
    \caption{\looseness=-1 Perplexity of different models trained on synthetic PCFG data. A$i$L$j$  refers to a model with $i$ attention heads and $j$ layers. Except for models with few layers (A12L1) and  few attention heads (A3L12), trained models have nearly the same perplexity.}
    \label{tab:pretraining-ppl}
\end{table}

\begin{figure}[!t]
    \centering
    \includegraphics[width=0.7\linewidth]{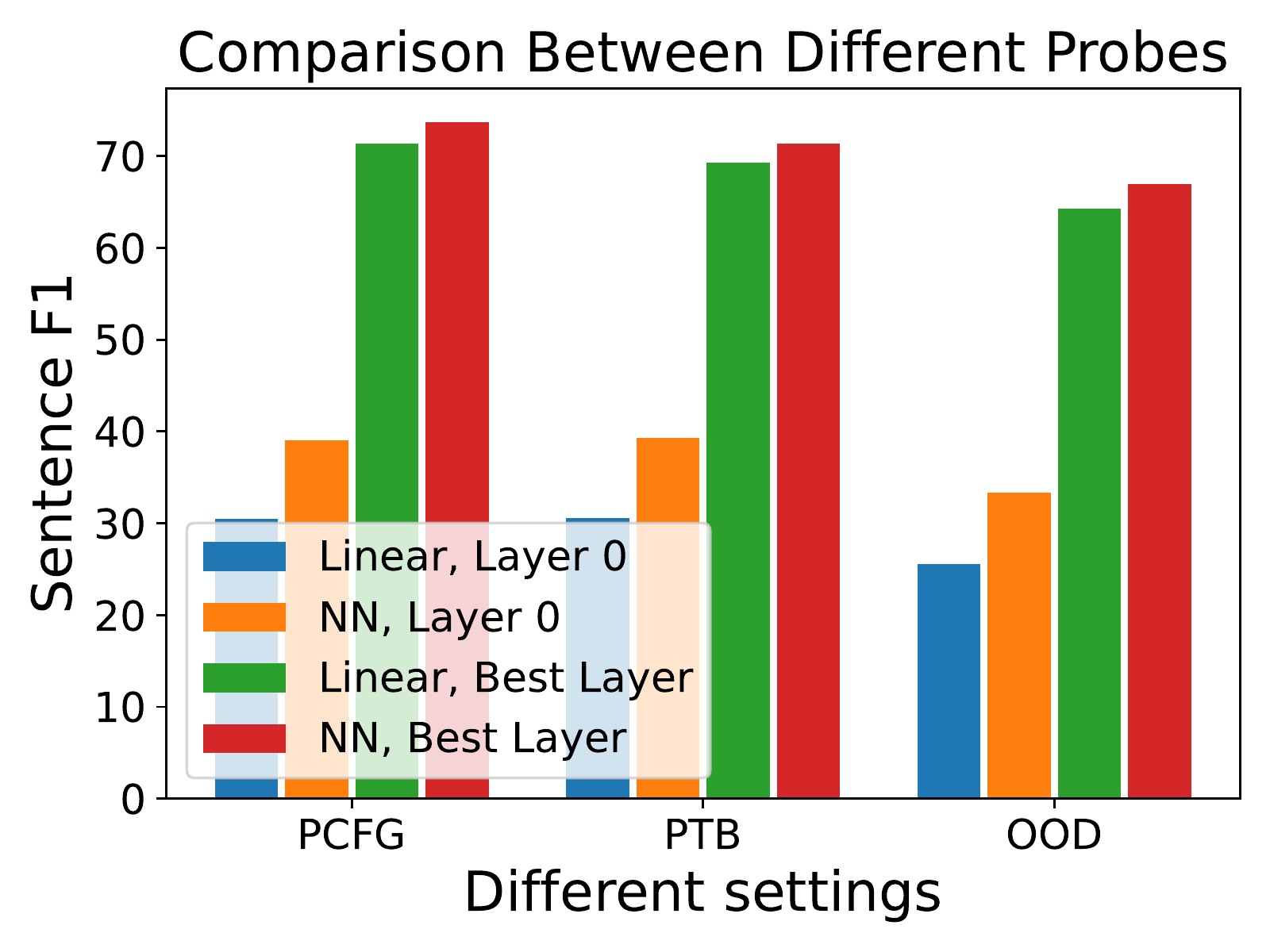}
    \caption{Comparison between different probes (linear or a 2-layer neural net) under different settings. 2-layer probes achieve better parsing performance, compared to linear probes. 
    The large performance gap of the probes on layer 0's embeddings from A12L12 and the best layer shows the existence of meaningful syntactic information in the contextualized embeddings.
    }
    \label{fig:probe-parsing-comparison}
\end{figure}

\begin{table*}[!th]
    \centering
    \footnotesize
    \begin{tabular}{|c|c|c|ccccccc|}
    \hline
         & & & IO & A12L12 & A12L1 & A12L3 & A12L6 & A3L12 & A24L12 \\
    \hline
        \multirow{6}{*}{\rotatebox[origin=c]{90}{Linear}}& \multirow{2}{*}{\begin{turn}{90} \dataset{PCFG} \end{turn}} & Sent. F1 & 81.61 &  \textbf{71.34} & 63.16 & 69.96 & \textbf{71.23} & 64.71 & \textbf{70.76} \\
        & & Corpus F1 & 71.65 & \textbf{63.01} & 54.24 & 61.54 & \textbf{62.57} & 55.36 & \textbf{62.56} \\
        \cline{2-10}
        & \multirow{2}{*}{\begin{turn}{90} \dataset{PTB} \end{turn}} & Sent. F1 & 78.77 &  \textbf{69.31} & 62.99 & 68.22 & 68.13 & 61.56 & \textbf{68.79} \\
        & & Corpus F1 & 75.90 & \textbf{65.01} & 59.96 & \textbf{65.21} & \textbf{65.01} & 58.31 & \textbf{65.97} \\
        \cline{2-10}
        & \multirow{2}{*}{\begin{turn}{90} OOD \end{turn}} & Sent. F1 & 81.61 & \textbf{64.26} & 57.96 & 63.22 & \textbf{63.89} & 58.00 & \textbf{63.88} \\
        & & Corpus F1 & 71.65 & \textbf{60.98} & 54.29 & 59.79 & \textbf{60.58} & 54.39 & \textbf{60.62} \\
        \hline
        \multirow{6}{*}{\rotatebox[origin=c]{90}{2-layer NN}}& \multirow{2}{*}{\begin{turn}{90} \dataset{PCFG} \end{turn}} & Sent. F1 & 81.61 &  \textbf{73.71} & 64.80 & 72.62 & \textbf{73.60} & 62.55 & \textbf{73.27} \\
        & & Corpus F1 & 71.65 & \textbf{66.18} & 57.16 & \textbf{65.36} & \textbf{66.01} & 53.36 & \textbf{65.92} \\
        \cline{2-10}
        & \multirow{2}{*}{\begin{turn}{90} \dataset{PTB} \end{turn}} & Sent. F1 & 78.77 & \textbf{71.32} & 64.89 & 70.15 & \textbf{70.33} & 63.23 & \textbf{70.59} \\
        & & Corpus F1 & 75.90 & \textbf{68.07} & 62.09 & \textbf{67.25} & \textbf{67.31} & 60.59 & \textbf{67.93} \\
        \cline{2-10}
        & \multirow{2}{*}{\begin{turn}{90} OOD \end{turn}} & Sent. F1 & 81.61 & \textbf{66.99} & 59.89 & \textbf{66.21} & \textbf{66.56} & 57.60 & \textbf{67.18} \\
        & & Corpus F1 & 71.65 & \textbf{63.89} & 56.74 & 63.30 & 63.81 & 54.60 & \textbf{64.54} \\
        \hline
    \end{tabular}
    \caption{Parsing results for different models under different settings using Linear and 2-layer neural net probes, when compared to Inside-Outside algorithm (IO). We report the best F1 score achieved using any of the layer's embeddings. 
    Scores within 1\% of the max (except IO) in each row are highlighted. 
    Models except A12L1 and A3L12 give decent parsing F1 scores, and models with more layers or heads tend to get better F1 scores in general.
    }
    \label{tab:parsing-results}
\end{table*}

\begin{table*}
    \centering
    \footnotesize
    \begin{tabular}{|c|cccccc|}
    \hline
         \makecell{Span \\Length} & A12L12 & A12L1 & A12L3 & A12L6 & A3L12 & A24L12 \\
    \hline
        $\ell = 2$ &  .88 / \textbf{.93} & .83 / .88 &  .88 / .91  &  .88 / \textbf{.92}  &  .86 / .88 & .87 / \textbf{.92} \\
        $\ell = 3$ &  .79 / \textbf{.90} & .74 / .84 &  .80 / .88  &  .79 / \textbf{.89}  &  .77 / .84 & .79 / \textbf{.89}  \\
        $\ell = 4$ &  .69 / \textbf{.86} & .65 / .77 &  .69 / .82  &  .69 / .84  &  .66 / .78 & .69 / \textbf{.85}  \\
        $\ell = 5$ &  .62 / .79 & .57 / .70 &  .62 / .77   &  .61 / \textbf{.81} &  .58 / .69 & .62 / .79  \\
        $\ell = 10$ & .51 / \textbf{.77} & .48 / .68 & .51 / .75 & .51 / \textbf{.78} & .51 / .61 & .51 / .73 \\
        \hline
    \end{tabular}
    \caption{Probing for the ``normalized'' marginal probabilities of spans at different lengths on different pre-trained models. We report the Pearson correlation between the predicted probabilities  and the span marginal probabilities computed by the Inside-Outside algorithm on \dataset{PTB} datasets, for both the linear and the 2-linear net probes (separated by /). 
    The high correlation indicates that the MLM pre-trained models approximately encode the marginal span probabilities of the Inside-Outside algorithm during pre-training. 
    }
    \label{tab:probe-probs-ptb}
\end{table*}

\subsection{Probing for constituency parse trees}\label{sec:parse}

\looseness=-1 We probe the language models pre-trained on synthetic PCFG data and show that these models indeed capture the ``syntactic information'', in particular, the structure of the constituency parse trees. 


\looseness=-1 \paragraph{Experiment setup} We mostly follow the probing procedure in \citet{vilares2020parsing} that predicts the relative depth of common ancestors between different token pairs and then constructs the constituency tree. Given a sentence $w_1w_2\dots w_L$ with parse tree $T$, we denote $\text{depth}(i,i+1)$ the depth of the least common ancestor of $w_i,w_{i+1}$ in the parse tree $T$. We want to find a probe $f^{(\ell)}$ to predict the relative depth $\text{tar}(i) = \text{depth}(i,i+1) - \text{depth}(i-1,i)$ for position $i$. In \citet{vilares2020parsing}, the probe $f^{(\ell)}$ is linear, and the input to the probe $f^{(\ell)}$ at position $i$ is the concatenation of the embeddings at position $i$ and the BOS (or EOS) token. Besides the linear probe $f^{(\ell)}$, we also experiment with the probe where $f^{(\ell)}$ is a 2-layer neural network with 16 hidden neurons. We consider three settings for probing: train and test the probe on synthetic PCFG data (\dataset{PCFG}); train and test on \dataset{PTB} dataset (\dataset{PTB}); and train on the synthetic PCFG data while test on \dataset{PTB} (out of distribution, OOD). The OOD setting serves as a baseline for a syntactic probe on \dataset{PTB} since semantic relations do not appear in the pre-trained model or the probe.

\looseness=-1 \paragraph{Experiment results}  \Cref{fig:probe-parsing-comparison} reveals a substantial difference between the probing outcomes of layer 0 embeddings and those of the best layer in all settings.  Both probing approaches profit greatly from the representations of subsequent layers. 

\looseness=-1\Cref{tab:parsing-results} shows probing results for different settings (\dataset{PCFG}, \dataset{PTB}, and OOD), different probes (linear or a 2-layer neural net) on different models. Except for A12L1 and A3L12, the linear and neural net probes give decent parsing scores (> 70\% sentence F1 for neural net probes) in both \dataset{PCFG} and \dataset{PTB} settings. As for the OOD setting, the performances achieved by the best layer drop by about 5\% compared with \dataset{PCFG} and \dataset{PTB}, but they are still much better than the performance achieved by the $0$-th layer embeddings. In this setting, there is no semantic information even in the probe itself and thus gives a baseline for the probes on \dataset{PTB} dataset that only uses syntactic information. As a comparison, the naive baseline, Right-branching (RB), reaches $<40\%$ for both sentence and corpus F1 score~\citep{li2020empirical} on \dataset{PTB} dataset, and if we use layer 0's embeddings to probe, the sentence F1 is $<41\%$ in all settings for all models. Our positive results on syntactic parsing support the claim that pre-training language models using MLM loss can indeed capture the structural information of the underlying constituency parse tree.







\begin{figure}[!th]
\begin{subfigure}[t]{0.49\textwidth}
    \centering
    \includegraphics[width=0.7\linewidth]{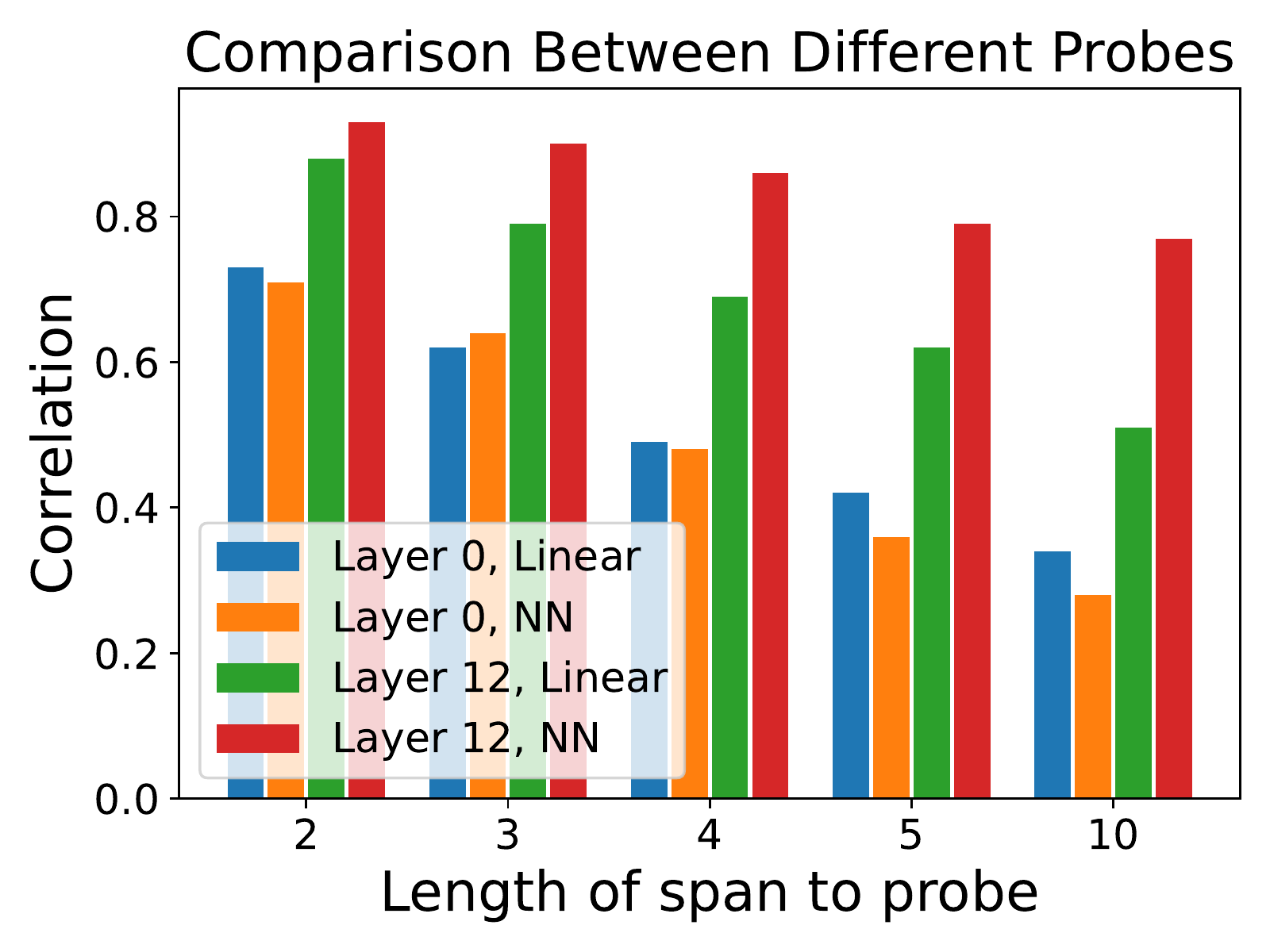}
    \caption{Compare linear/2-layer NN probes under \dataset{PTB} setting. We observe: (a) 2-layer NN probe has better performance, and (b) the probes give better performance on 12th-layer embeddings.} 
    \label{fig:probe-prob-comparison}
\end{subfigure}
\hfill
\begin{subfigure}[t]{0.49\textwidth}
    \centering
    \includegraphics[width=0.7\linewidth]{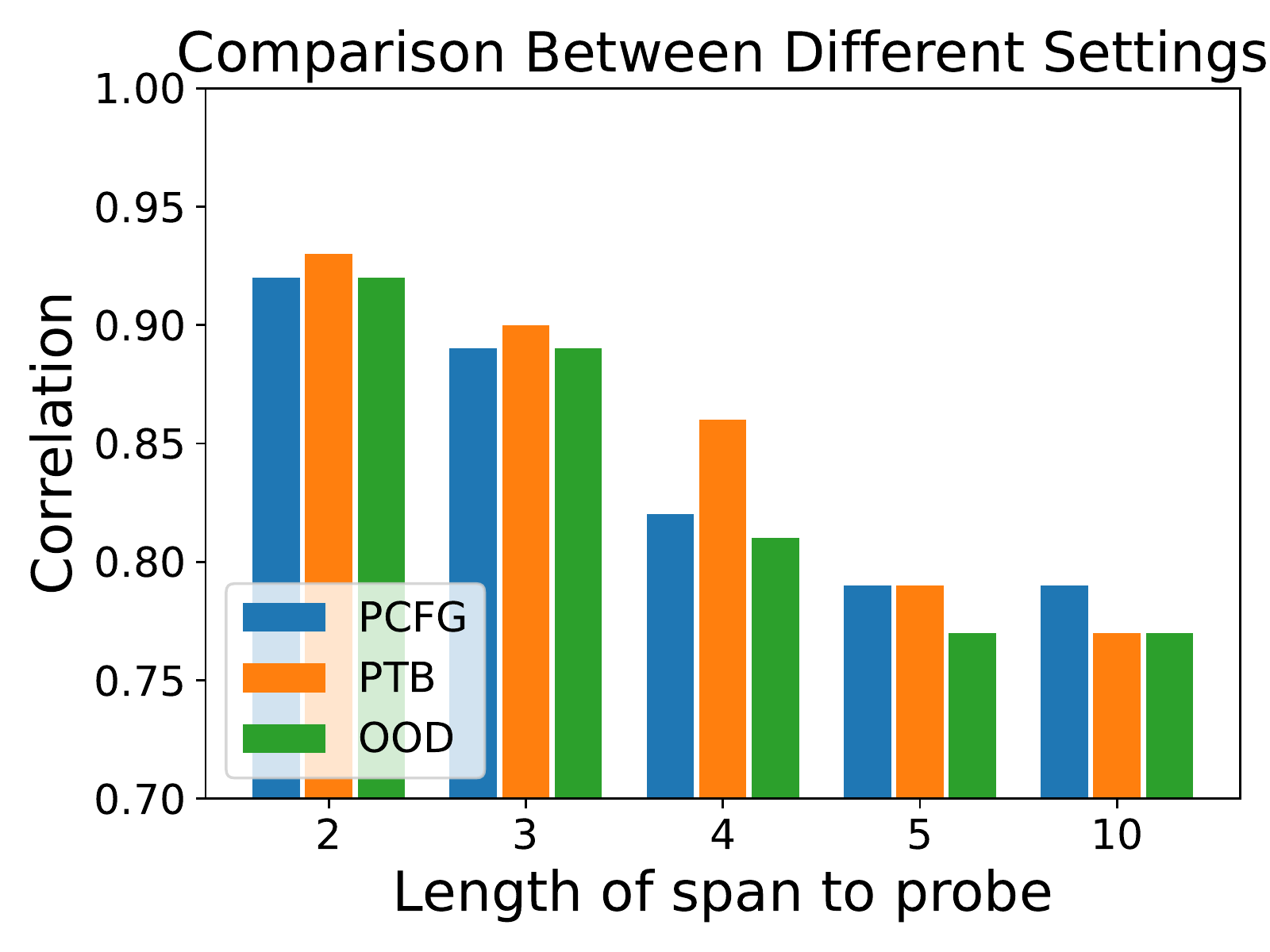}
    \caption{Performance of 2-layer neural net probe on the $12$-th layer embeddings under different settings. The closer correlation performance of the probe across  settings (including OOD) indicates true marginal probabilities captured by the trained probe. }
    \label{fig:probe-prob-setting-comparison}
\end{subfigure}
\caption{Comparison between different probes for marginal probabilities on the A12L12 model. The y-axis denotes correlation between the prediction and the target, and the x-axis denotes probes for different lengths.}
\end{figure}

\subsection{Probing for the marginal probabilities}\label{sec:probe-marginal-probs}

\looseness=-1\Cref{sec:parse} verifies that language models can capture structure information of the parse trees, but we still don't know if the model executes the Inside-Outside algorithm proposed in \Cref{sec:construct-io,sec:mlmandio}.
In this subsection, we test if model representations can be used to predict marginal probabilities computed in the Inside-Outside algorithm. 

\looseness=-1\paragraph{Experiment setup} We train a probe to predict the normalized marginal probabilities for spans with a specific length. Fix the span length $\ell$, for each sentence $w_1w_2\dots w_L$, denote $\ve_1, \ve_2,\dots,\ve_L$ the embeddings from the last layer of the pre-trained language model. We want to find a probe $f^{(\ell)}$ such that for each span $[i,i+\ell-1]$ with length $\ell$, the probe $f^{(\ell)}([\ve_i;\ve_{i+\ell-1}])$ predicts the normalized marginal probability of span $[i,i+\ell-1]$, i.e. $\text{tar}(i,i+\ell-1) = s(i,i+\ell-1) / \max_{j,j'}s(j,j')$,
where $s(i,j) = \max_A \mu(A,i,j)$ is the marginal probability of span $[i,j]$ and $\mu(A,i,j)$ is given by eq.~\ref{eq:marginal_probability}.
The input to the probe $[\ve_i;\ve_{i+\ell-1}]\in\R^{2d}$ is the concatenation of $\ve_i$ and $\ve_{i+\ell-1}$. To test the sensitivity of our probe, we also take the embeddings from the $0$-th layer as input to the probe $f^{(\ell)}$.

\looseness=-1We give two options for the probe $f^{(\ell)}$: (1) linear, and (2) a 2-layer neural network with 16 hidden neurons, since the relation between the embeddings and the target may not be a simple linear function. Similar to the \Cref{sec:parse}, we also consider three settings: \dataset{PCFG}, \dataset{PTB}, and OOD.

\paragraph{Experiment results} 

\looseness=-1\Cref{fig:probe-prob-comparison} reports the correlation between the span marginal probabilities and the predictions of the 4 different probes for A12L12 model. For both linear and 2-layer neural net probes, changing the input from layer 0 to layer 12 drastically increases the predicted correlation, which again suggests that the uncontextualized embeddings don't contain enough information about the marginal probabilities. Besides, the neural net can predict better on layer 12 embeddings, but performs nearly the same on layer 0, suggesting that the neural network is a better probe in this setting. 

\looseness=-1\Cref{fig:probe-prob-setting-comparison} compares the probing results under three different settings. Surprisingly, we find that the probe can achieve high correlation with the real marginal probabilities under all settings. Furthermore, we observe that there is almost no drop in performance when changing the test dataset from \dataset{PCFG} to \dataset{PTB} (\dataset{PCFG} setting and OOD setting). This result implies that the probe, along with the embeddings, indeed contains the syntactic information computed by the Inside-Outside algorithm and is not overfitting to the training dataset.

\looseness=-1\Cref{tab:probe-probs-ptb} shows the probing results on different pre-trained models. The results show that the neural network probe is highly correlated with the target for most pre-trained models, except for A12L1 and A3L12 models. Surprisingly, even for length $10$ spans, the neural network probe still achieves an F1 score of up to 78\% for the best model. The high correlation suggests that the pre-trained models contain certain syntactic information computed by the Inside-Outside algorithm. Overall, the results indicate that MLM training may incentivize the model to approximate the Inside-Outside algorithm, thus validating our constructions in \Cref{sec:construction}.

\subsection{Control tasks}\label{sec:control-task-main}

\begin{table*}
    \footnotesize
    \centering
    \begin{tabular}{|c|c|ccccccccccccc|}
    \hline
         & & L0 & L1 & L2 & L3 & L4 & L5 & L6 & L7 & L8 & L9 & L10 & L11 & L12 \\
         \hline
        \multirow{3}{*}{\rotatebox[origin=c]{90}{Linear}} & pred. rel. depth & .606 & .760 & .789 & .796 & .800 & .803 & .803 & .803 & .802 & .801 & .800 & .800 & .799 \\
         & control task & .758 & .677 & .645 & .626 & .620 & .610 & .608 & .617 & .599 & .595 & .612 & .606 & .608 \\
         & selectivity & -.152 & .083 & .144 & .170 & .180 & .193 & .195 & .186 & .203 & \textbf{.206} & .188 & .194 & .191 \\
         \hline
        \multirow{3}{*}{\rotatebox[origin=c]{90}{NN}} & pred. rel. depth & .616 & .771 & .804 & .810 & .814 & .807 & .815 & .802 & .795 & .810 & .806 & .803 & .776 \\
         & control task & .861 & .793 & .758 & .667 & .728 & .653 & .653 & .668 & .678 & .693 & .680 & .697 & .687 \\
         & selectivity & -.245 & -.022 & .046 & .143 & .086 & .154 & \textbf{.162} & .134 & .117 & .117 & .126 & .106 & .089 \\ 
         \hline
    \end{tabular}
    \caption{Computing the selectivity of constituency parsing probes with linear and 2-layer NN architectures (see \cref{sec:parse} and \cref{sec:control-task-main}). The ``pred. rel. depth'' rows denote the probing results for the relative depth of common ancestors in the constituency parse tree using different layers' representations of A12L12. We report the predicting accuracy under the \dataset{PTB} setting where the probe is trained and tested on \dataset{PTB} dataset. The ``control task'' rows denote the predicting accuracy for the control task on \dataset{PTB} dataset using different layers' representations of A12L12. The selectivity is the difference between the original task performance and the control task performance. We can observe that for all layers representations, the probe with a linear classifier has a larger selectivity.}
    \label{tab:parsing-control-task}
\end{table*}

\looseness=-1In probing experiments, it is crucial to ensure that the probing performance accurately reflects the presence of the specific information we intend to test. 
Consequently, it is undesirable for the probe to possess excessive power and be capable of learning all aspects (see \cref{sec:preliminary} for further discussions). 
\citet{chen2021probing} utilize ``sensitivity'' to assess the extent to which the probe captures the targeted information. The ``sensitivity'' of a probe is defined as the difference in probing performance between the layer of interest and the 0-th layer.
Intuitively, a large gap indicates that the probe fails to perform adequately using representations from the 0-th layer but achieves better performance when utilizing representations from a later layer, thus confirming the presence of the targeted information. 

\looseness=-1\citet{hewitt2019designing} introduced another metric, known as ``selectivity'', to assess the degree to which the probe captures the targeted information. Broadly speaking, \citet{hewitt2019designing} devised a specific task referred to as the ``control task'' to evaluate the probe's capability to align with specific types of random labels. Subsequently, ``selectivity'' is defined as the difference in performance between the probe for the original task, utilizing the layer of interest, and the probe for the control task, also utilizing the layer of interest. Intuitively, a large gap suggests that the probe lacks sufficient expressive power, resulting in the performance boost originating from the representations of the layer being probed. 

\looseness=-1Note that a probe with higher ``sensitivity'' does not necessarily imply larger ``selectivity''. Nevertheless, as demonstrated in the subsequent parts (and appendix), the metrics of ``sensitivity'' and ``selectivity'' align for both the constituency parsing probes and the marginal probability probes (\cref{sec:control-task}). We sketch the control task design and results for the constituency parsing probe, and defer the preliminaries of control tasks in~\citet{hewitt2019designing} and the control tasks experiments for marginal probabilities probe to \cref{sec:control-task}.

\paragraph{Control task for constituency parsing } For the constituency parsing in \cref{sec:parse}, we follow the design of control task for sequence labeling problems~\citep{hewitt2019designing}.
Specifically, we have $y_i = \text{tar}(i) = \text{depth}(i,i+1) - \text{depth}(i-1,i)$ for position $i$. Then for the control task, for each word $w$, we uniformly sample $\phi(w) \in \{-1,0,1\}$, and then define the labels for the control task as $\hat y_{1:T} = [\phi(x_1), \phi(x_2),\dots,\phi(x_T)]$.

\paragraph{\emph{Selectivity} is aligned with \emph{Sensitivity}}

\cref{tab:parsing-control-task} provides a summary of the performance of the constituency parsing probe, employing different architectures (linear classifier and a 2-layer neural network with 16 hidden neurons), on the original task, control task, as well as the selectivity.

\looseness=-1From \cref{tab:parsing-control-task}, the probe with a 2-layer NN achieves slightly higher accuracy in predicting the relative depth of common ancestors, leading to a higher F1 score in parsing. However, its performance on the control task surpasses that of the probe with a linear classifier by a significant margin. This suggests that when using the ``selectivity'' metric, the linear probe outperforms the 2-layer neural network probe in recovering the constituency parse tree, aligning with the conclusions drawn using the ``sensitivity metric'' (see Figure \ref{fig:probe-parsing-comparison}, where the sensitivity of the linear probe is greater than that of the 2-layer NN probe). Experiment results for marginal probability control task (\Cref{sec:control-task}) also support the alignment of \emph{Selectivity} and \emph{Sensitivity}.

\section{Related Works}

\paragraph{(Structural) probing}
\looseness=-1Several recent works on probing have aimed to study the encoded information in BERT-like models~\citep{rogers2020primer}. \citet{hewitt2019structural,reif2019visualizing,manning2020emergent,vilares2020parsing,maudslay2020tale,maudslay2021syntactic,chen2021probing,arps2022probing,jawahar2019does} have demonstrated that it is possible to predict various syntactic information present in the input sequence, including parse trees or POS tags, from internal states of BERT. 
In contrast to existing approaches that commonly employ a model pre-trained on natural language, we pre-train our model under PCFG-generated data to investigate the interplay between the data, the MLM objective, and the architecture's capacity for parsing. 
Besides syntax, probing has also been used to test other linguistic structures like semantics, sentiment, etc.~\citep{belinkov2017neural,reif2019visualizing,kim2020pre,richardson2020probing,vulic2020probing,conia-navigli-2022-probing}.


\paragraph{Expressive power of transformers}
\looseness=-1\citet{Yun2020Are,yun2020n} show that transformers are universal sequence-to-sequence function approximators. Later, \citet{perez2021attention,bhattamishra2020computational} show that attention models can simulate Turing machines, with \citet{wei2022statistically} proposing statistically meaningful approximations of Turing machines. 
To understand the behavior of moderate-size transformer architectures, many works have investigated specific classes of languages, e.g. bounded-depth Dyck languages~\citep{yao2021self}, modular prefix sums~\citep{anil2022exploring}, adders~\citep{nanda2023progress}, regular languages~\citep{bhattamishra2020ability}, and sparse logical predicates~\citep{edelman2022inductive}. \citet{merrill2022saturated} relate saturated transformers with constant depth threshold circuits, and \citet{liu2022transformers} provide a unified theory on understanding automata within transformers.
These works study expressive power under a class of synthetic language. Compared to the prior works, our results are more related to the natural language, as we consider not only a class of synthetic language (PCFG), but also a specific PCFG tailored to the natural language.



\section{Conclusion}

\looseness=-1 In this work, we show that MLM with moderate size has the capacity to parse decently well. We probe BERT-like models pre-trained (with MLM loss) on the synthetic text generated using PCFGs to verify that these models capture syntactic information.
Furthermore, we show that the models contain the marginal span probabilities computed by the Inside-Outside algorithm, thus connecting MLM and parsing. We hope our findings may yield new insights into large language models and MLM. 



\section*{Limitation}

We believe that the main limitations of our study are the transformer architecture and size. 

Due to limitations imposed by GPU resources, we assess encoder-only models with specific limitations: a maximum of 12 layers, 24 attention heads per layer, and 768 embedding dimensions.  Nevertheless, all the experiment results begin to stabilize for smaller models and generalize to the largest model we investigate. Hence, we believe that the results can generalize to even larger models. 

Our central theoretical discovery (Theorem 3.3) establishes a connection between the masked language modeling (MLM) loss and the Inside-outside algorithm.  Extending to auto-regressive models like GPT is an important theoretical question and is kept for future study.

\paragraph{Acknowledgement}
Haoyu Zhao, Abhishek Panigrahi, and Sanjeev Arora are supported by funding from NSF, ONR, Simons Foundation, DARPA, and SRC. Rong Ge is supported by NSF Award DMS-2031849, CCF-1845171 (CAREER), CCF-1934964 (Tripods), and a Sloan Research Fellowship.

\bibliographystyle{acl_natbib}
\bibliography{ref.bib}

\clearpage
\appendix

\section*{Appendix}

\section{More Experiment Results}
In this section, we provide more experiment results for RoBERTa pre-trained on PCFG-generated data. In \Cref{sec:more-parsing-result}, we show more structural probing results related to the experiments in \Cref{sec:parse}. In \Cref{sec:attn-patterns}, we do some simple analysis on the attention patterns for RoBERTa pre-trained on PCFG-generated data, trying to gain more understanding of the mechanism beneath large language models.

\subsection{Details for pre-training}\label{sec:pretraining-details}
\paragraph{Experiment setup}  We generate $10^7$ sentences for the training set from the PCFG, with an average length of $25$ words. The training set is roughly $10\%$ in size compared to the training set of the original RoBERTa which was trained on a combination of Wikipedia (2500M words) plus BookCorpus (800M words). We also keep a small validation set of $5 \times 10^4$ sentences generated from the PCFG to track the MLM loss. We follow \cite{izsak-etal-2021-train,wettig2022should} to pre-train all our models within a single day on a cluster of 8 RTX 2080 GPUs. Specifically, we train our models with AdamW \cite{loshchilov2017decoupled} optimization, using $4096$ sequences in a batch and hyperparameters $(\beta_1, \beta_2, \epsilon) = (0.9, 0.98, 10^{-6}).$ We follow a linear warmup schedule for $1380$ training steps with the peak learning rate of $2 \times 10^{-3}$, after which the learning rate drops linearly to $0$ (with the max-possible training step being $2.3 \times 10^{4}$). We report the performance of all our models at step $5 \times 10^{3}$ where the loss seems to converge for all the models.

\paragraph{Architecture} To understand the impact of different components in the encoder model, we pre-train different models by varying the number of attention heads and layers in the model. To understand the role of the number of layers in the model, we start from the RoBERTa-base architecture, which has 12 layers and 12 attention heads, and vary the number of layers to 1,3,6 to obtain $3$ different architectures. Similarily, to understand the role of the number of attention heads in the model, we start from the RoBERTa-base architecture and vary the number of attention heads to 3 and 24 to obtain $2$ different architectures.

\paragraph{Data generation from PCFG} Strings are generated from the PCFG $\gG = (\gN, \gI, \gP, n, p)$ as follows: We always maintain a string $s_t \in ([n]\cup \gN)^*$ at step $t$. The initial string $s_1 = \text{ROOT}$. At step $t$, if all characters in $s_t$ belong to $[n]$, the generation process ends, and $s_t$ is the resulting string. Otherwise, we pick a character $A\in s_t$ such that $A\in\gN$. If $A\in\gP$, we replace the character $A$ to $w$ with probability $\Pr[A\to w]$. If $A\in\gI$, we replace the character $A$ to two characters $BC$ with probability $\Pr[A\to BC]$.


\subsection{More results on constituency parsing}\label{sec:more-parsing-result}

\paragraph{More details on probing experiments}
In \Cref{sec:parse}, we mention that there are three settings: \dataset{PCFG}, \dataset{PTB}, and OOD. We generate two synthetic PCFG datasets according to the PCFG generation process: the first contains 10,000 sentences, which serves as the training set for probes, and the second contains 2,000 sentences, which serves as the test set for probes. As for the \dataset{PTB}, the training set for the probes consists of the first 10,000 sentences from sections 02-21, and we use \dataset{PTB} section 22 as the test set for the probes. In the \dataset{PCFG} setting, we train on the PCFG training set we generated, and test on the PCFG test set. In the \dataset{PTB} setting, we train on the \dataset{PTB} training set (10,000 sentences in sections 02-21) and test on the \dataset{PTB} test set (section 22). In the OOD setting, we train on the \dataset{PCFG} training set, while test on the \dataset{PTB} test set (section 22).

For the linear probe, we directly use Scikit-learn~\citep{scikit-learn}. For the 2-layer NN probe, we train the neural net with Adam optimizer with learning rate $1e-3$. We optimize for 800 epochs, and we apply a multi-step learning rate schedule with milestones $200,400,600$ and decreasing factor $0.1$. The batch size for Adam is chosen to be 4096.

\paragraph{Probing on embeddings from different layers} In \Cref{sec:parse}, we show the probing results on the embeddings either from $0$-th layer or from the best layer (the layer that achieves the highest F1 score) of different pre-trained models. In this section, we show how the F1 score changes with different layers.

\begin{figure*}[!th]
\begin{subfigure}[t]{0.48\textwidth}
    \centering
    \includegraphics[width=\textwidth]{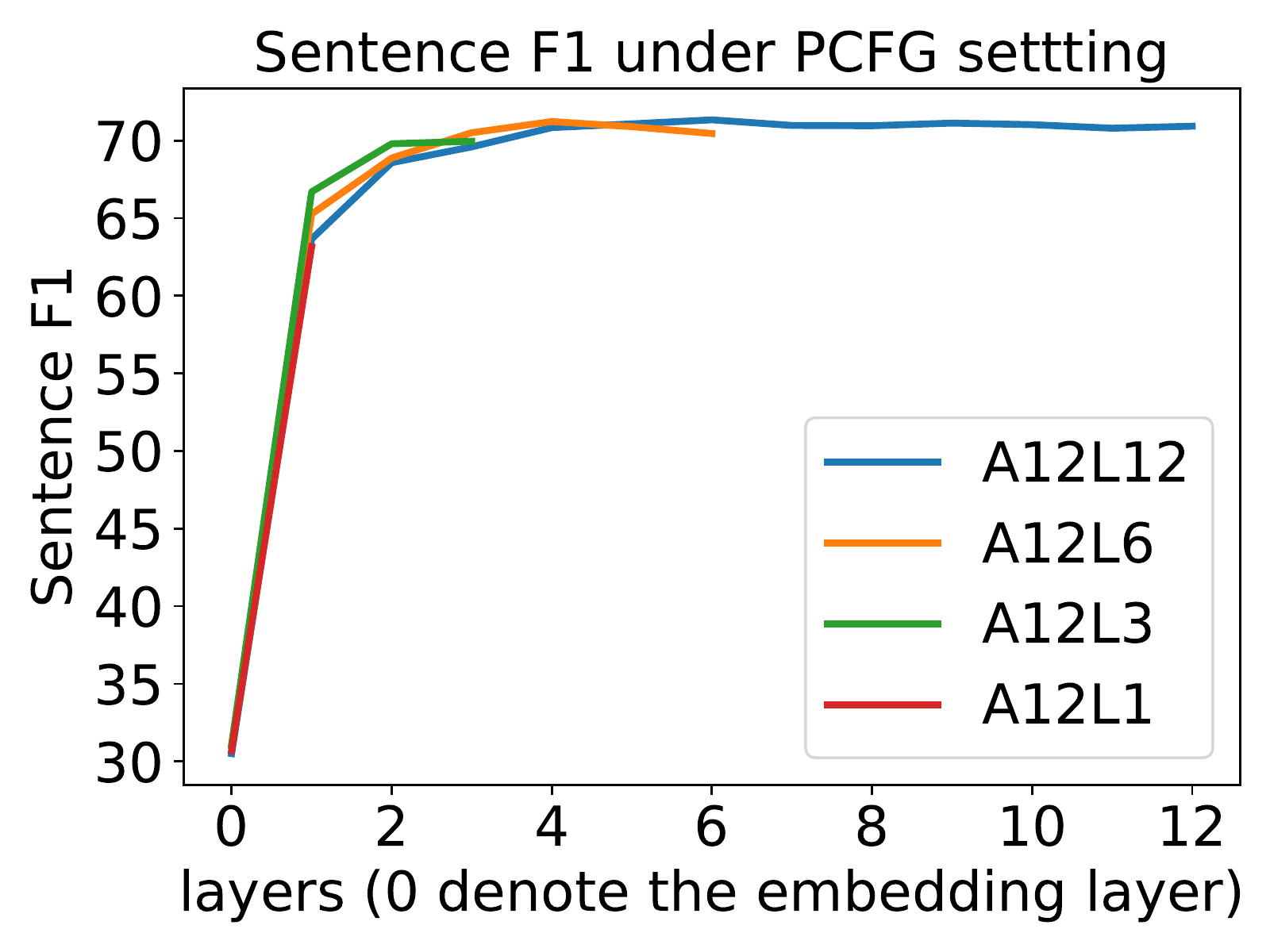}
    \caption{Comparison under \dataset{PCFG} setting. We compare the models with different number of layers.}
    \label{fig:parsing-pcfg-diff-layers-linear}
\end{subfigure}
\hfill
\begin{subfigure}[t]{0.48\textwidth}
    \centering
    \includegraphics[width=\textwidth]{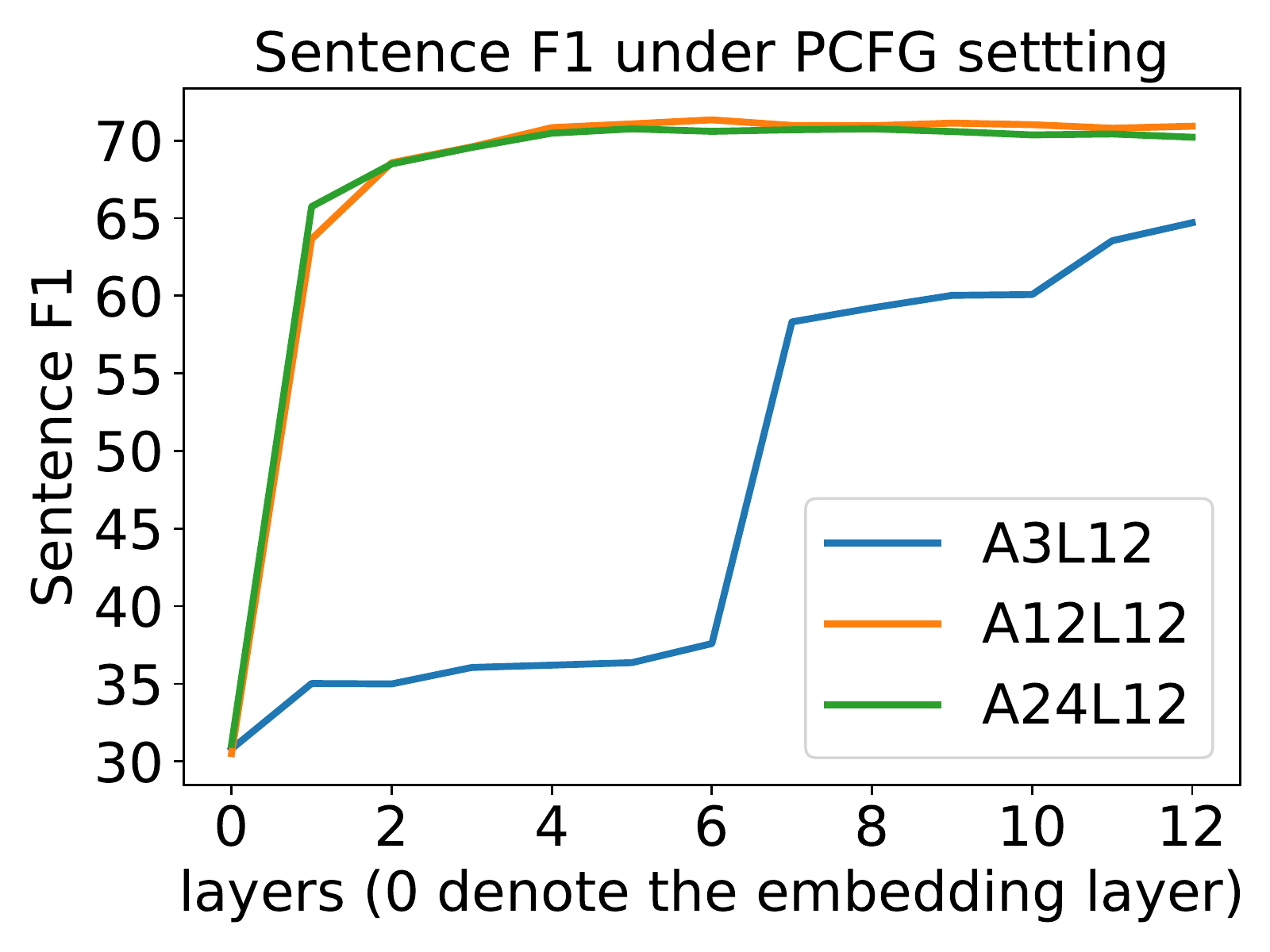}
    \caption{Comparison under \dataset{PCFG} setting. We compare the models with different number of attention heads.}
    \label{fig:parsing-pcfg-diff-attn-linear}
\end{subfigure}

\begin{subfigure}[t]{0.48\textwidth}
    \centering
    \includegraphics[width=\textwidth]{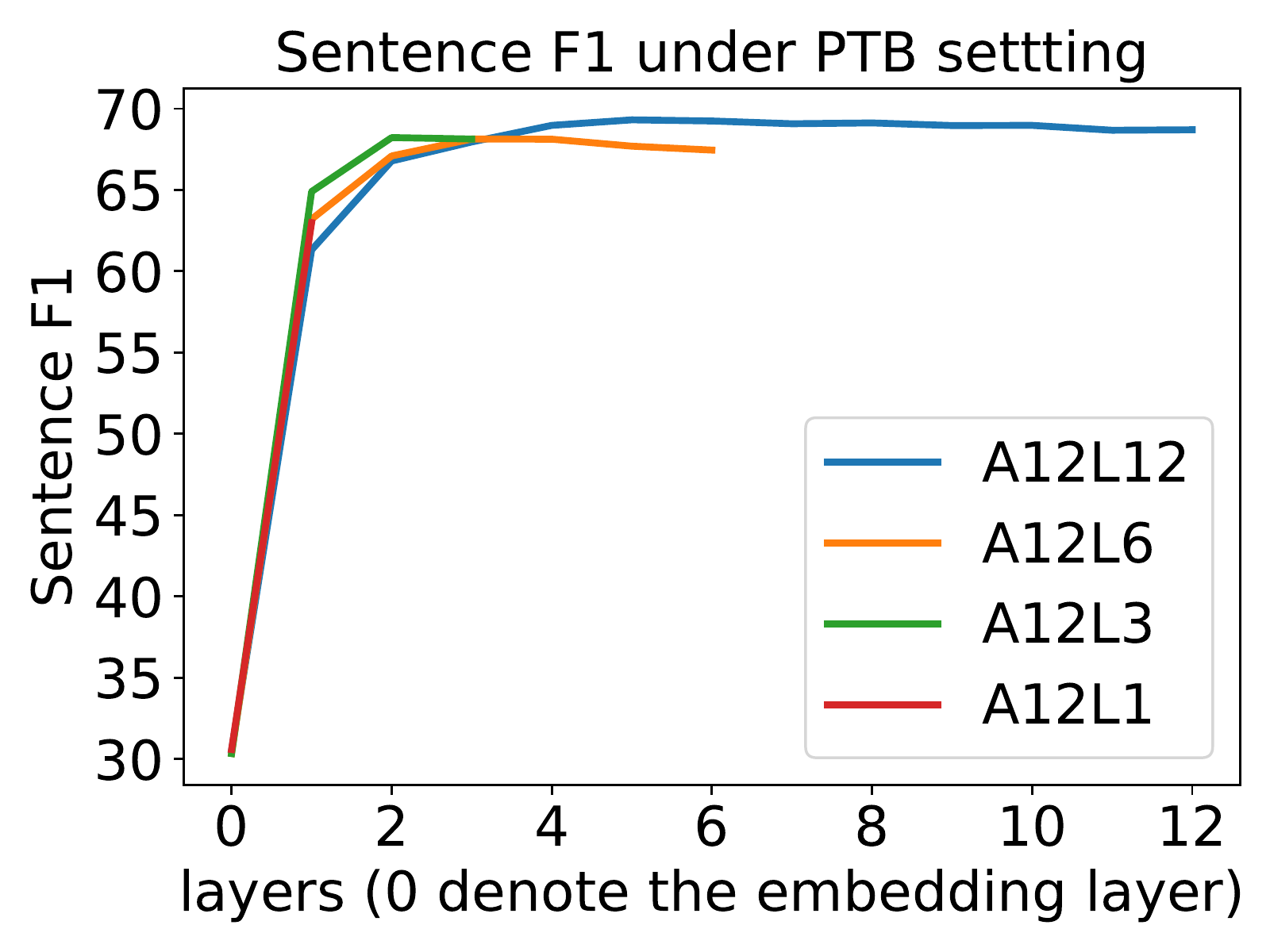}
    \caption{Comparison under \dataset{PTB} setting. We compare the models with different number of layers.}
    \label{fig:parsing-ptb-diff-layers-linear}
\end{subfigure}
\hfill
\begin{subfigure}[t]{0.48\textwidth}
    \centering
    \includegraphics[width=\textwidth]{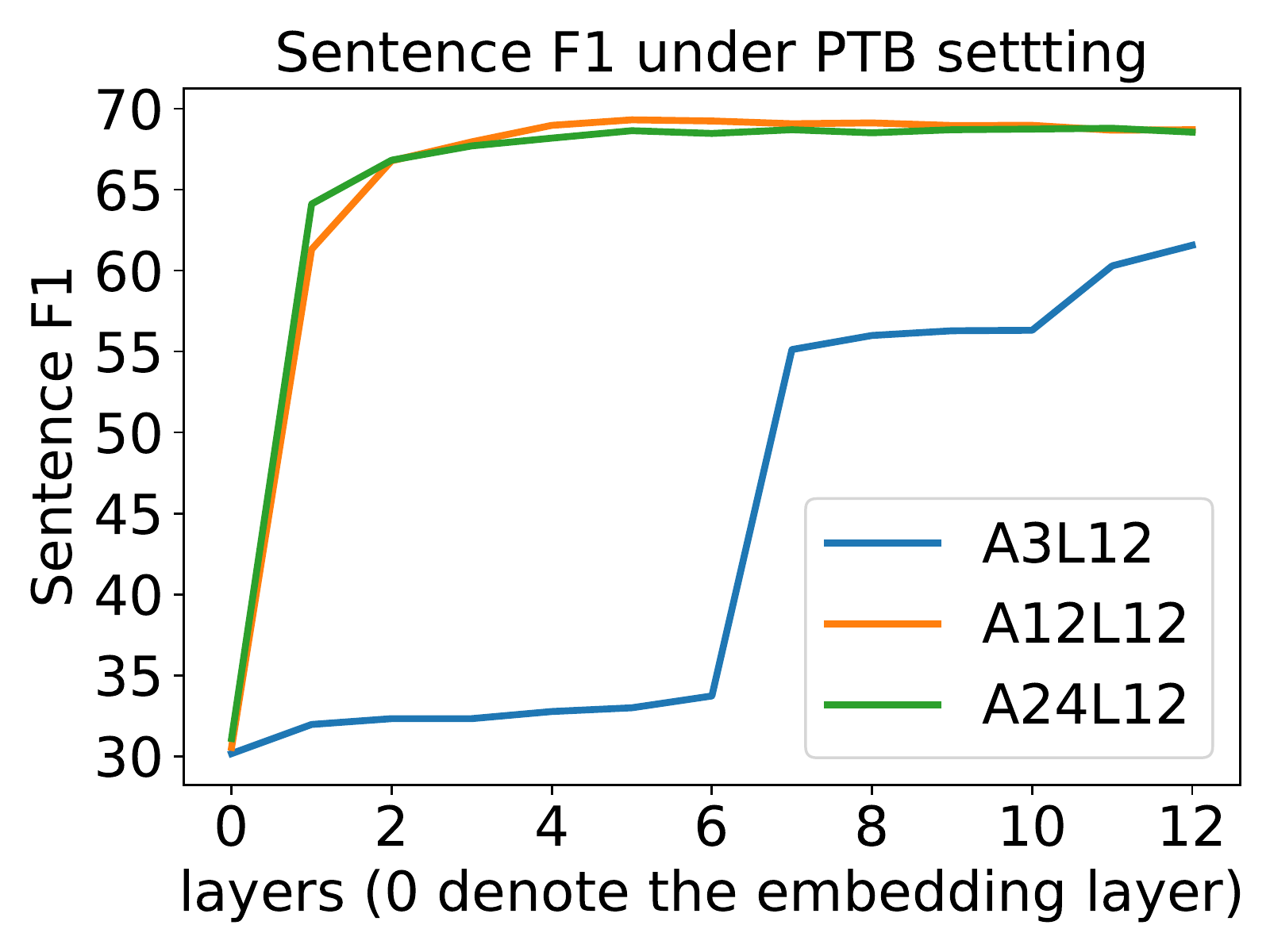}
    \caption{Comparison under \dataset{PTB} setting. We compare the models with different number of attention heads.}
    \label{fig:parsing-ptb-diff-attn-linear}
\end{subfigure}
    \caption{Sentence F1 for linear probes $f(\cdot)$ trained on different layers' embeddings for different pre-trained models. We show the results under \dataset{PCFG} and \dataset{PTB} settings. A$i$L$j$ denotes the pre-trained model with $i$ attention heads and $j$ layers.}
    \label{fig:f1-using-different-layers}
\end{figure*}

\Cref{fig:f1-using-different-layers} shows sentence F1 scores for linear probes $f(\cdot)$ trained on different layers' embeddings for different pre-trained models. We show the results under the \dataset{PCFG} and \dataset{PTB} settings. From \Cref{fig:f1-using-different-layers}, we observe that using the embeddings from the $0$-th layer can only get sentence F1 scores close to (or even worse than) the naive Right-branching baseline for all the pre-trained models. However, except for model A3L12, the linear probe can get at least $60\%$ sentence F1 using the embeddings from layer 1. Then, the sentence F1 score increases as the layer increases, and gets nearly saturated at layer 3 or 4. The F1 score for the latter layers may be better than the F1 score at layer 3 or 4, but the improvement is not significant. The observations still hold if we change the linear probe to a neural network, consider the OOD setting instead of \dataset{PCFG} and \dataset{PTB}, or change the measurement from sentence F1 to corpus F1.

Our observations suggest that most of the constituency parse tree information can be encoded in the lower layers, and a lot of the parse tree information can be captured even in the first layer. Although our constructions (\Cref{thm:hard_attnt,thm:soft_attnt}) and approximations (\Cref{thm:approx-few-nt-informal,thm:approx-low-rank-informal}) try to reduce the number of attention heads and the number of embedding dimensions close to the real language models, we don't know how to reduce the number of layers close to BERT or RoBERTa (although our number is acceptable since GPT-3 has 96 layers). More understanding of how language models can process such information in such a small number of layers is needed.


\paragraph{Comparison with probes using other input structures} In \Cref{sec:parse}, we train a probe $f(\cdot)$ to predict the relative depth $\text{tar}(i) = \text{depth}(i,i+1) - \text{depth}(i-1,i)$, and the input to the probe $f$ is the concatenation of the embedding $\ve^{(\ell)}_i$ at position $i$ and the embedding $\ve^{(\ell)}_{\text{EOS}}$ for the EOS token at some layer $\ell$. Besides taking the concatenation $[\ve^{(\ell)}_i; \ve^{(\ell)}_{\text{EOS}}]$ as the input structure of the probe, it is also natural to use the concatenation $[\ve^{(\ell)}_{i-1}; \ve^{(\ell)}_i; \ve^{(\ell)}_{i+1}]$ to predict the relative depth $\text{tar}(i)$. In this part, we compare the performances of probes with different input structures. We use EOS to denote the probe that takes $[\ve^{(\ell)}_i; \ve^{(\ell)}_{\text{EOS}}]$ as the input and predicts the relative depth, while ADJ (Adjacent embeddings) to denote the probe the takes $[\ve^{(\ell)}_{i-1}; \ve^{(\ell)}_i; \ve^{(\ell)}_{i+1}]$ as input.

\begin{figure*}
    \begin{subfigure}[t]{0.48\textwidth}
        \centering
        \includegraphics[width=\textwidth]{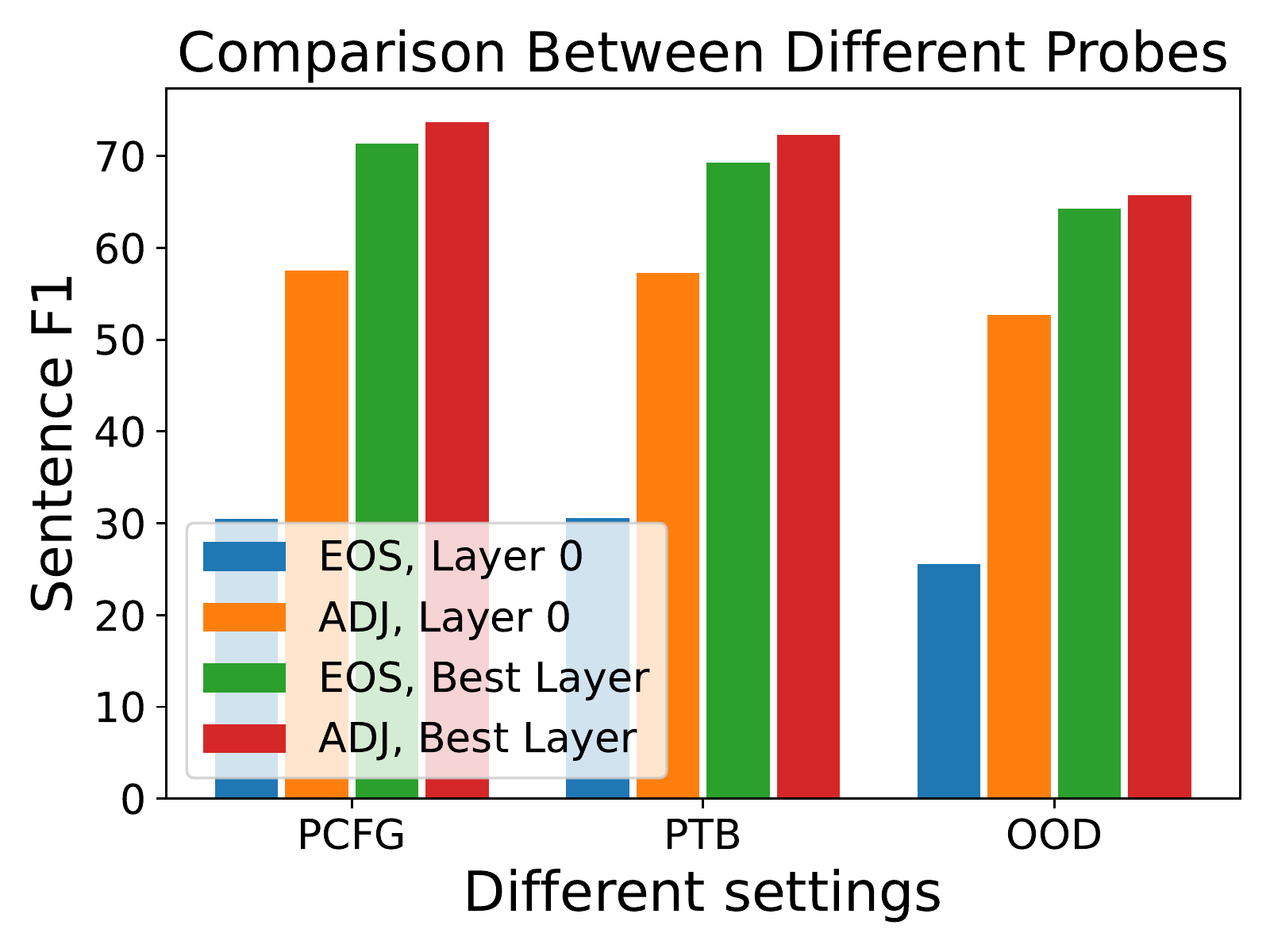}
        \caption{Comparison of different inputs under different settings when the probe $f(\cdot)$ is linear.}
        \label{fig:parsing-diff-inputs-linear}
    \end{subfigure}
    \begin{subfigure}[t]{0.48\textwidth}
        \centering
        \includegraphics[width=\textwidth]{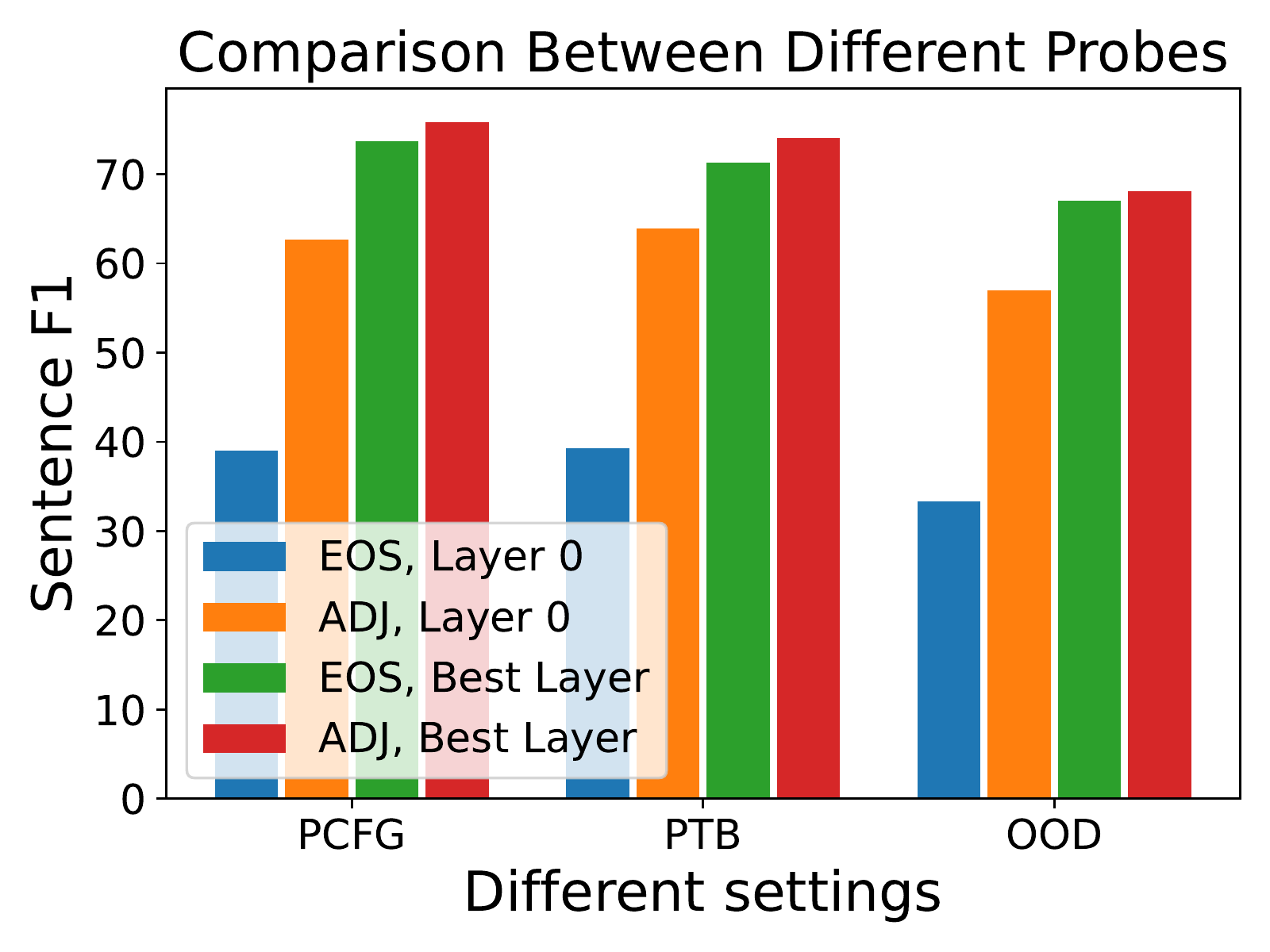}
        \caption{Comparison of different inputs under different settings when the probe $f(\cdot)$ is a 2-layer neural network.}
        \label{fig:parsing-diff-inputs-nn}
    \end{subfigure}
    \caption{Comparison of the probes with different inputs under different settings. We probe the model with 12 attention heads and 12 layers, and report the scores with $f(\cdot)$ taking embeddings from layer 0 or the embeddings from the best layer. EOS denotes the probe that takes $[\ve^{(\ell)}_i; \ve^{(\ell)}_{\text{EOS}}]$ as input and predicts the relative depth $\text{tar}(i)$, and ADJ (Adjacent embeddings) denotes the probe that takes $[\ve^{(\ell)}_{i-1}; \ve^{(\ell)}_i; \ve^{(\ell)}_{i+1}]$ as input.}
    \label{fig:parsing-diff-inputs}
\end{figure*}

\Cref{fig:parsing-diff-inputs} shows the probing results on A12L12, the model with 12 attention heads and 12 layers. We compare the probes with different inputs structure (EOS or ADJ), and the input embeddings come from different layers (the $0$-th layer or the layer that achieves the best F1 score). We observe that: (1) the probes using ADJ input structure have better parsing scores than the probes using EOS input structure, and (2) the sentence F1 for the probes using the ADJ input structure is high even if the input comes from layer 0 of the model ($>55\%$ for linear $f(\cdot)$ and $>60\%$ for neural network $f(\cdot)$). Although the probe using ADJ has better parsing scores than the probe using EOS, it is harder to test whether it is a good probe, since the concatenation of adjacent embeddings $[\ve^{(0)}_{i-1}; \ve^{(0)}_i; \ve^{(0)}_{i+1}]$ from layer $0$ is already contextualized, and it is hard to find a good baseline to show that the probe is \emph{sensitive} to the information we want to test. Thus, we choose to follow~\citet{vilares2020parsing,arps2022probing} and use the probe with input structure $[\ve^{(\ell)}_i; \ve^{(\ell)}_{\text{EOS}}]$ in \Cref{sec:parse}.

Nonetheless, the experiment results for probes taking $[\ve^{(0)}_{i-1}; \ve^{(0)}_i; \ve^{(0)}_{i+1}]$ as input are already surprising: by knowing three adjacent word identities and their position (the token embedding $\ve^{(0)}_i$ contains both the word embedding and the positional embedding) and train a 2-layer neural network on top of that, we can get $62.67\%, 63.91\%, 57.02\%$ sentence F1 scores under \dataset{PCFG}, \dataset{PTB}, and OOD settings respectively. As a comparison, the probe taking $[\ve^{(\ell)}_i; \ve^{(\ell)}_{\text{EOS}}]$ as input~\citep{vilares2020parsing,arps2022probing} only get $39.06\%, 39.31\%, 33.33\%$ sentence F1 under \dataset{PCFG}, \dataset{PTB}, and OOD settings respectively. It shows that lots of syntactic information (useful for parsing) can be captured by just using adjacent words without more context.

\paragraph{More discussion on probing measurement} (Unlabelled) F1 score is the default performance measurement in the constituency parsing and syntactic probing literature. However, we would like to point out that only focusing on the F1 score may cause some bias. Because all the spans have equal weight when computing the F1 score, and most of the spans in a tree have a short length (if the parse tree is perfectly balanced, then length 2 spans consist of half of the spans in the parse tree), one can get a decently well F1 score by only getting correct on short spans. Besides, we also show that by taking the inputs $[\ve^{(0)}_{i-1}; \ve^{(0)}_i; \ve^{(0)}_{i+1}]$ from layer 0 of the model (12 attention heads and 12 layers), we can already capture a lot of the syntactic information useful to recover the constituency parse tree (get a decently well F1 score). Thus, the F1 score for the whole parse tree may cause people to focus less on the long-range dependencies or long-range structures, and focus more on the short-range dependencies or structures.

To mitigate this problem, \citet{vilares2020parsing} computed the F1 score not only for the whole parse tree, but also for each length of spans. \citet{vilares2020parsing} showed that BERT trained on natural language can get a very good F1 score when the spans are short (for length 2 spans, the probing F1 is over $80\%$), but when the span becomes longer, the F1 score quickly drops. Even for spans with length 5, the F1 score is less than $70\%$, and for spans with length 10, the F1 score is less than $60\%$. Our experiments that probe the marginal probabilities for different lengths of spans (\Cref{sec:probe-marginal-probs}) can also be viewed as an approach to mitigate the problem.

\subsection{More results on probing marginal probabilities}

In \cref{sec:probe-marginal-probs}, we conduct probing experiments to demonstrate the predictability of the "normalized marginal probabilities" computed by the Inside-Outside algorithm using transformer representations. Our objective is to establish a strong correlation, measured through the Pearson correlation coefficient. However, we have not provided a comprehensive explanation for our preference for Pearson correlation over alternative metrics such as Spearman correlation. In the following section, we show the experiment results measured by the Spearman correlation, and give an explanation of why we prefer the Pearson correlation over the Spearman correlation.

\begin{table*}
    \centering
    \footnotesize
    \begin{tabular}{|c|cccccc|}
    \hline
         \makecell{Span \\Length} & A12L12 & A12L1 & A12L3 & A12L6 & A3L12 & A24L12 \\
    \hline
        $\ell = 2$ &  .71 / \textbf{.93} & .69 / .88 &  .75 / \textbf{.93}  &  .71 / \textbf{.93}  &  .76 / .86 & .75 / \textbf{.92} \\
        $\ell = 5$ &  .59 / \textbf{.82} & .54 / .64 &  .47 / .79  &  .49 / .79  &  .54 / .71 & .48 / .79 \\
        $\ell = 10$ & .43 / \textbf{.78} & .48 / .68 & .59 / .73 & .45 / .75 & .33 / .62 & .39 / .72 \\
        \hline
    \end{tabular}
    \caption{Probing for the ``normalized'' marginal probabilities of spans at different lengths on different pre-trained models. We report the Spearman and Pearson correlations (separated by /) between the predicted probabilities and the span marginal probabilities computed by the Inside-Outside algorithm on \dataset{PTB} datasets for the 2-linear net probe. 
    }
    \label{tab:probe-probs-ptb-spearman}
\end{table*}

\begin{figure*}
    \begin{subfigure}[t]{0.32\textwidth}
    \centering
    \includegraphics[width=\textwidth]{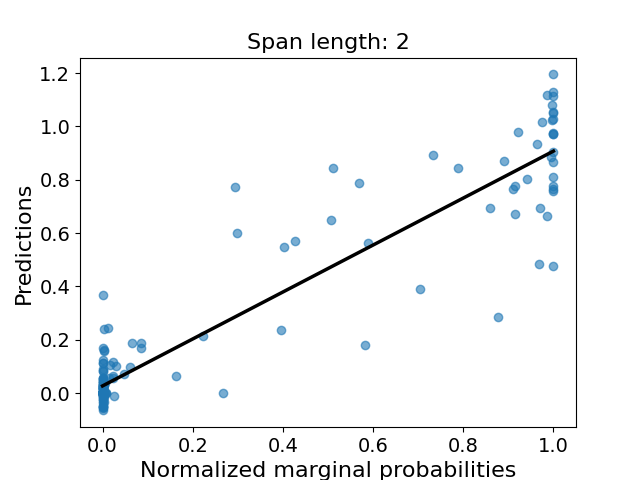}
    \caption{Span length to probe: $\ell = 2$.}
    \label{fig:probe-prob-l2}
\end{subfigure}
\hfill
\begin{subfigure}[t]{0.32\textwidth}
    \centering
    \includegraphics[width=\textwidth]{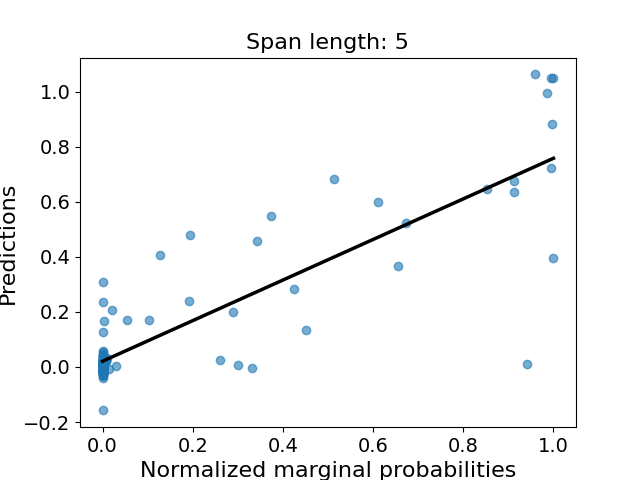}
    \caption{Span length to probe: $\ell = 5$.}
    \label{probe-prob-l5}
\end{subfigure}
\hfill
\begin{subfigure}[t]{0.32\textwidth}
    \centering
    \includegraphics[width=\textwidth]{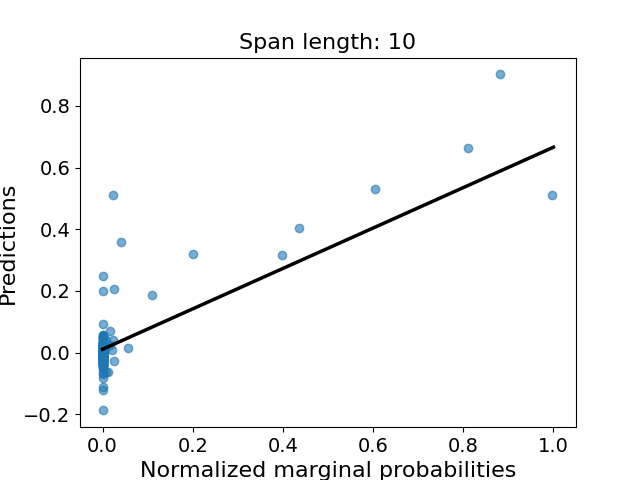}
    \caption{Span length to probe: $\ell = 10$.}
    \label{fig:probe-prob-l10}
\end{subfigure}
\caption{The predicted probability versus true normalized marginal probability plot for different span lengths $\ell$ using 2-layer NN probe with the 12-th layer's representations from A12L12 model. In each figure, we sample 200 points (each point corresponds to a span) to plot from the test set. The y-axis denotes the predicted probabilities and the x-axis denotes the true normalized marginal probabilities. The line shows the best linear fit for all the spans in the test set. We can observe that there are lots of points that have very small normalized marginal probabilities, and it is very hard to predict their rank correctly, thus resulting in a low Spearman correlation.}
\label{fig:probe-probe-pred-vs-true}
\end{figure*}

\paragraph{Measure with Spearman correlation} \cref{tab:probe-probs-ptb-spearman} summarizes the correlations between the predicted probabilities and the span marginal probabilities computed by the Inside-Outside algorithm on \dataset{PTB} datasets for the 2-linear net probes. It is evident that the Spearman correlation is significantly lower than the Pearson correlation, indicating that the probe primarily captures "linear" correlations rather than rank-based relationships.

In order to investigate the underlying cause of this phenomenon, we plot the predicted probabilities against the true normalized marginal probabilities, as shown in \cref{fig:probe-probe-pred-vs-true}. Numerous points have extremely small normalized marginal probabilities, particularly when the probe length $\ell$ is large (e.g., $\ell = 5, 10$). This observation aligns with the intuition that the probability of a randomly selected span existing in the constituency parse tree is low.

However, accurately predicting the exact rank for the points clustered near the origin proves to be extremely challenging, leading to a relatively low Spearman correlation. In contrast, when considering the Pearson correlation, the noise associated with predicting spans having low normalized marginal probabilities is relatively small compared to the overall "variance" of the data points. Furthermore, it is evident that the probe exhibits greater efficacy in capturing the "influential spans" characterized by large normalized marginal probabilities. Achieving relatively accurate predictions for these influential spans accounts for a significant portion of the observed variation, leading to a relatively high Pearson correlation.

\subsection{More details on control tasks}\label{sec:control-task}

In this section, we present more details for the design of the control task in~\citep{hewitt2019designing}, and also show the control task experiment for the marginal probability probes (\cref{sec:probe-marginal-probs}).

\paragraph{Control task} \citet{hewitt2019designing} considered control task for sequence labeling problems: Given a sentence $x_{1:T}$, the goal is to label each word $y_{1:T}$. For example, the Part-of-speech tagging problem and the dependency parsing all belong to the sequence labeling category, since for Part-of-speech tagging, $y_i$ is the POS tag of $x_i$, and for dependency parsing, $y_i$ is the parent of $x_i$ in the parse tree. For a sequence labeling problem, the control task for this sequence labeling problem consists of two key components:
\begin{enumerate}
    \item Structure: the output $\hat y_i$ of a word $x_i$ is a deterministic function of $x_i$, i.e., $\hat y_i = \phi(x_i)$.
    \item Randomness: The output $\hat y_i$ for each word $x_i$ is sampled independently at random.
\end{enumerate}

Then, the goal of the control task is to fit the labels $\hat y_{1:T}$ using the probe with the input $h_{1:T}$ where $h_{1:T}$ denote the hidden representations of the specific layer of the transformer. Please refer to Section 2 of \citet{hewitt2019designing} for more details and examples on control task.


\paragraph{Control task for marginal probability probe} For the marginal probability probe in \cref{sec:probe-marginal-probs}, we need to generalize the original control task from sequence labeling problem to span labeling problem. Given a span $x_{i:j}$, the original goal is to predict the normalized marginal probability $y_{i,j} = \text{tar}(i,j) = s(i,j) / \max_{j_1,j_2}s(j_1,j_2)$ where $s(i,j)$ is the marginal probability for span $i:j$ computed by the Inside-Outside algorithm. Now for each pair of words $w_1, w_2$, we uniformly sample $\phi(w_1,w_2)\in [0,1]$. Then for the sequence $x_{1:T}$, we have the label for the control task $\hat y_{i,j} = \phi(x_i, x_j)$.

\paragraph{\emph{Selectivity} is aligned with \emph{Sensitivity} for marginal probability probes}

In \cref{sec:control-task-main}, we design the control task for constituency parsing probes and show that \emph{selectivity} is aligned with \emph{sensitivity} (\cref{tab:parsing-control-task}). In this part, we show that \emph{selectivity} is aligned with \emph{sensitivity} for the marginal probability probes.
\cref{tab:marginal-prob-control-task} provides a summary of the performance of the constituency parsing probe and the marginal probability probes, employing different architectures (linear classifier and a 2-layer neural network with 16 hidden neurons), on the original task, control task, as well as the selectivity.


Based on the information presented in \cref{tab:marginal-prob-control-task}, it is evident that the probe utilizing a 2-layer neural network demonstrates superior performance in predicting span probabilities for the control task. Nonetheless, compared to the linear probe, the 2-layer neural network probe achieves significantly better results on the original task, resulting in a larger ``selectivity''. Analyzing \cref{fig:probe-prob-comparison}, we observe that the 2-layer NN probe exhibits significantly stronger predictive correlation than the linear probe at the 12-th layer of A12L12, while displaying similar performance at the 0-th layer, which contributes to a higher ``sensitivity''. Consequently, the ``selectivity'' metric aligns with the ``sensitivity'' metric for marginal probability probes, indicating that 2-layer NN probes capture a relatively greater amount of syntactic information.

\begin{table*}
    \footnotesize
    \centering
    \begin{tabular}{|c|c|ccccc|}
    \hline
         & Probe span length & 2 & 3 & 4 & 5 & 10 \\
         \hline
        \multirow{3}{*}{\rotatebox[origin=c]{90}{Linear}} & pred. marginal prob. & .88 & .79 & .69 & .62 & .51 \\
         & control task & .62 & .55 & .53 & .60 & .58 \\
         & selectivity & .26 & .24 & .16 & .02 & -.07 \\
         \hline
        \multirow{3}{*}{\rotatebox[origin=c]{90}{NN}} & pred. marginal prob. & .93 & .90 & .86 & .79 & .77 \\
         & control task & .66 & .66 & .69 & .66 & .68 \\
         & selectivity & .27 & .24 & .17 & .13 & .09 \\ 
         \hline
    \end{tabular}
    \caption{Computing the selectivity of marginal probability probes with linear and 2-layer NN architectures (see \cref{sec:probe-marginal-probs} and \cref{sec:control-task}). The ``pred. marginal prob.'' rows denote the probing results for the ``normalized'' marginal probabilities of spans at different lengths using the 12-th layer of A12L12. We report the Pearson correlation between the predicted probabilities and the span marginal probabilities computed by the Inside-Outside algorithm on \dataset{PTB} dataset. The ``control task'' rows denote the Pearson correlation between the predicted probabilities and the probabilities generated from the control task on \dataset{PTB} dataset using the 12-th layer of A12L12. The selectivity is the difference between the original task performance and the control task performance. We can observe that for spans with all lengths tested, the probe with 2-layer NN has a larger selectivity, especially when the probe length is large.}
    \label{tab:marginal-prob-control-task}
\end{table*}

\subsection{Analysis of attention patterns}\label{sec:attn-patterns}
In \Cref{sec:parse}, we probe the embeddings of the models pre-trained on synthetic data generated from PCFG and show that model training on MLM indeed \emph{captures} syntactic information that can recover the constituency parse tree. \Cref{thm:io-optimal-mlm} builds the connection between MLM and the Inside-Outside algorithm, and the connection is also verified in \Cref{sec:probe-marginal-probs}, which shows that the embeddings also contain the marginal probability information computed by the Inside-Outside algorithm. However, we only build up the correlation between the Inside-Outside algorithm and the attention models, and we still don't know the mechanism inside the language models: the model may be executing the Inside-Outside algorithm (or some approximations of the Inside-Outside algorithm), but it may also use some mechanism far from the Inside-Outside algorithm but happens to contain the marginal probability information. We leave for future work the design of experiments to interpret the content of the contextualized embeddings and thus ``reverse-engineer'' the learned model. In this section, we take a small step to understand more about the mechanism of language models: we need to \emph{open up the black box} and go further than probing, and this section serves as one step to do so.

\paragraph{General idea} The key ingredient that distinguishes current large language models and the fully-connected neural networks is the self-attention module. Thus besides probing for certain information, we can also look at the attention score matrix and discover some patterns. In particular, we are interested in how far an attention head looks at, which we called the "averaged attended distance".

\paragraph{Averaged attended distance} For a model and a particular attention head, given a sentence $s$ with length $L_s$, the head will generate an $L_s \times L_s$ matrix $\mA$ containing the pair-wise attention score, where each row of $\mA$ sums to 1. Then we compute the following quantity ``Averaged attended distance''
\[\text{AD}_s = \frac{1}{L_s}\sum_{1\le i,j \le L_s} |i-j|\cdot \mA_{i,j},\]
which can be intuitively interpreted as ``the average distance this attention head is looking at''. We then take the average of the quantity for all sentences. We compute ``Averaged attended distance'' for three models on the synthetic \dataset{PCFG} dataset and \dataset{PTB} dataset. The models all have 12 attention heads in each layer but have 12, 6, 3 layers respectively.

\begin{figure*}[!t]
\centering
\begin{subfigure}[b]{0.48\textwidth}
\centering
\includegraphics[width=\textwidth]{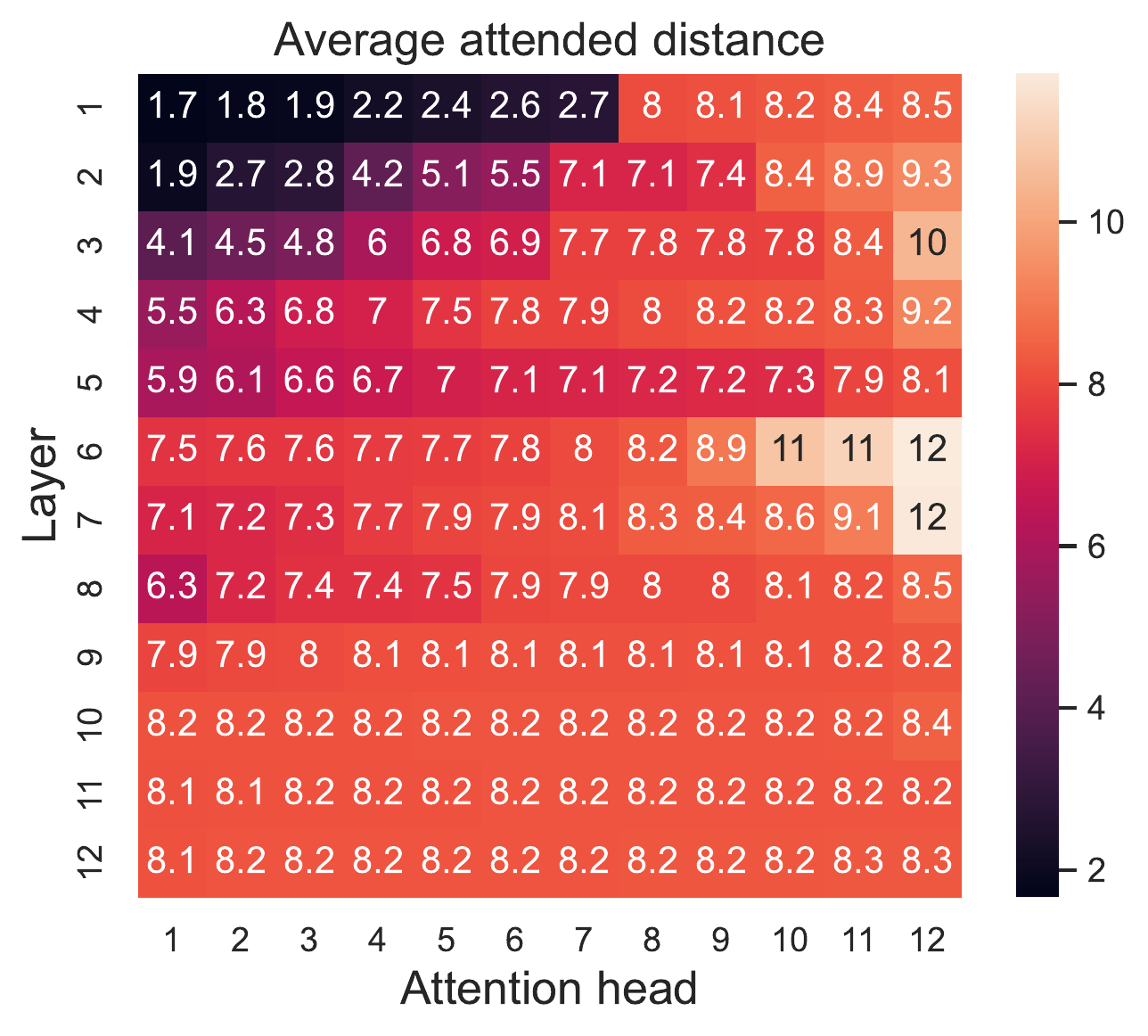}
\caption{12 attention heads and 12 layers, \dataset{PCFG} dataset.}
\label{fig:attn-dist-a12l12-pcfg}
\end{subfigure}
\hfill
\begin{subfigure}[b]{0.48\textwidth}
\centering
\includegraphics[width=\textwidth]{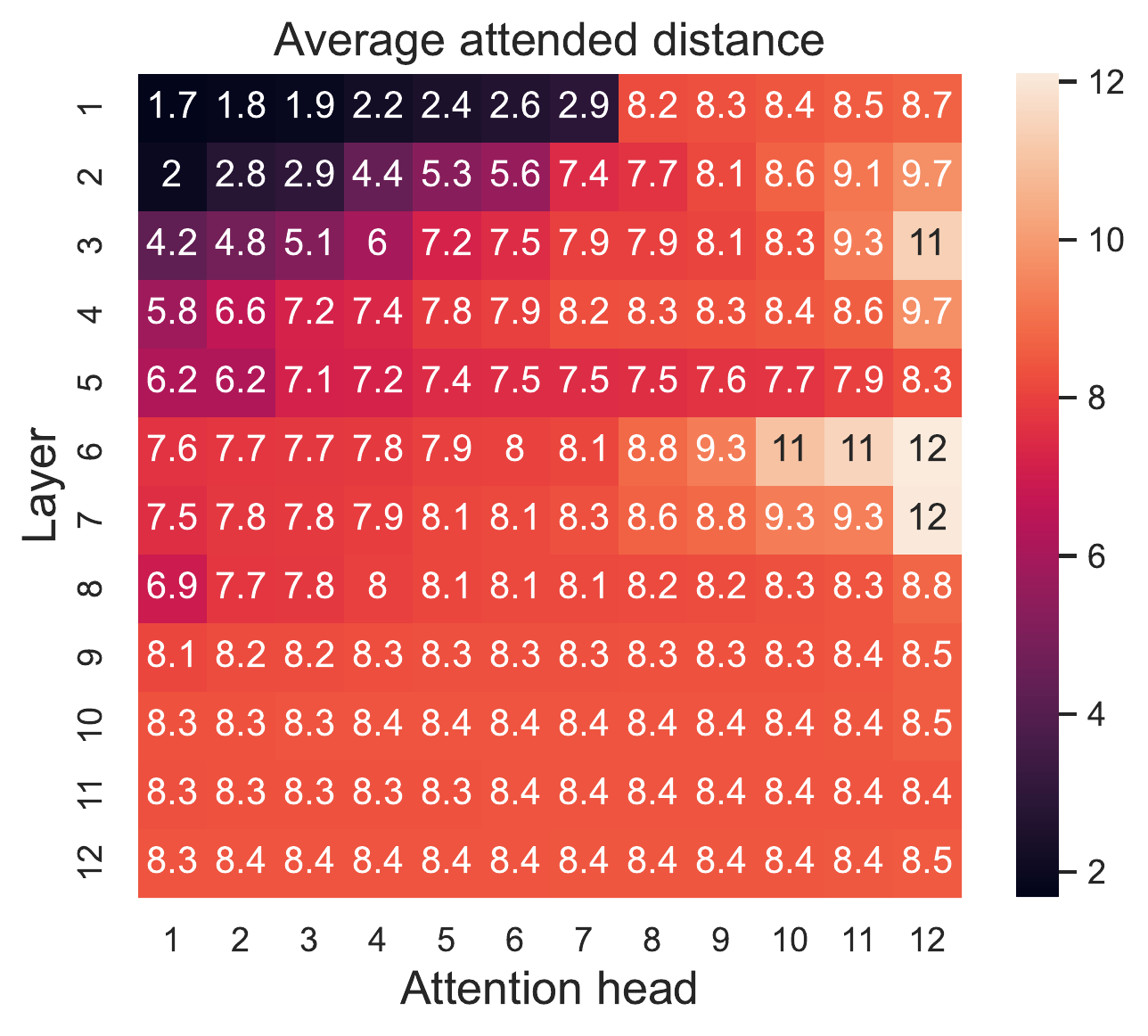}
\caption{12 attention heads and 12 layers, \dataset{PTB} dataset.}
\label{fig:attn-dist-a12l12-ptb}
\end{subfigure}

\begin{subfigure}[b]{0.48\textwidth}
\centering
\includegraphics[width=\textwidth]{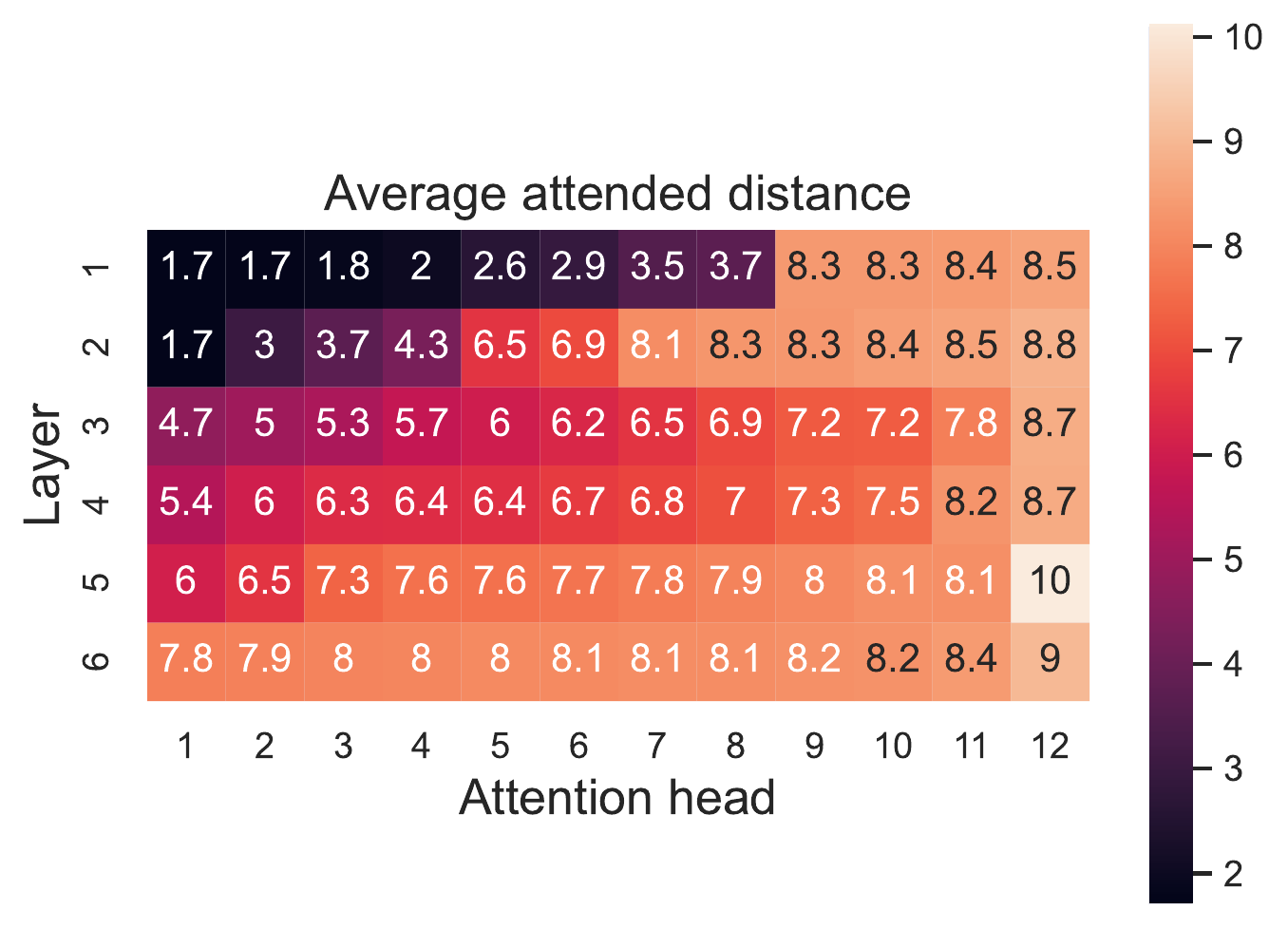}
\caption{12 attention heads and 6 layers, \dataset{PCFG} dataset.}
\label{fig:attn-dist-a12l6-pcfg}
\end{subfigure}
\hfill
\begin{subfigure}[b]{0.48\textwidth}
\centering
\includegraphics[width=\textwidth]{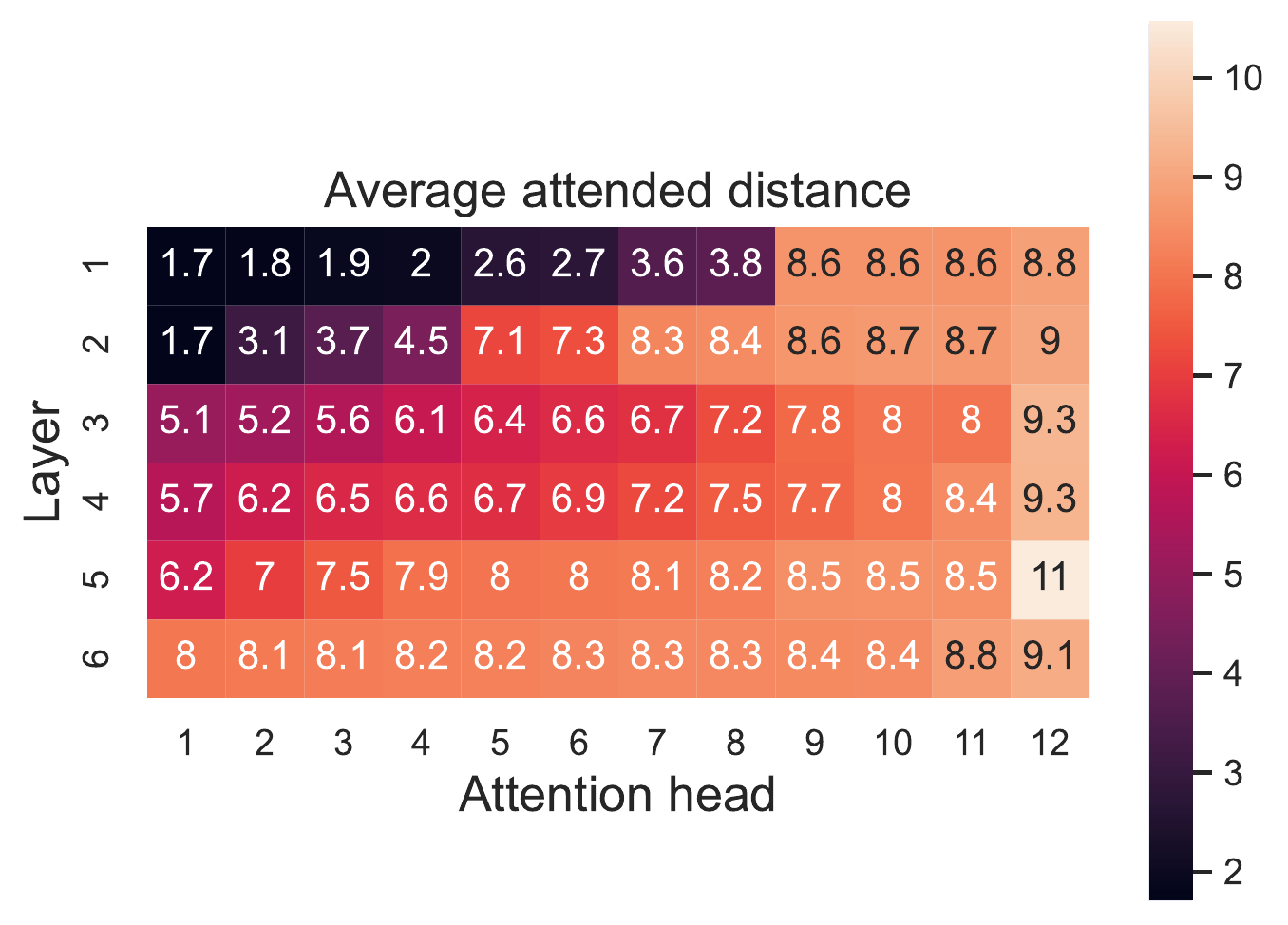}
\caption{12 attention heads and 6 layers, \dataset{PTB} dataset.}
\label{fig:attn-dist-a12l6-ptb}
\end{subfigure}

\begin{subfigure}[b]{0.48\textwidth}
\centering
\includegraphics[width=\textwidth]{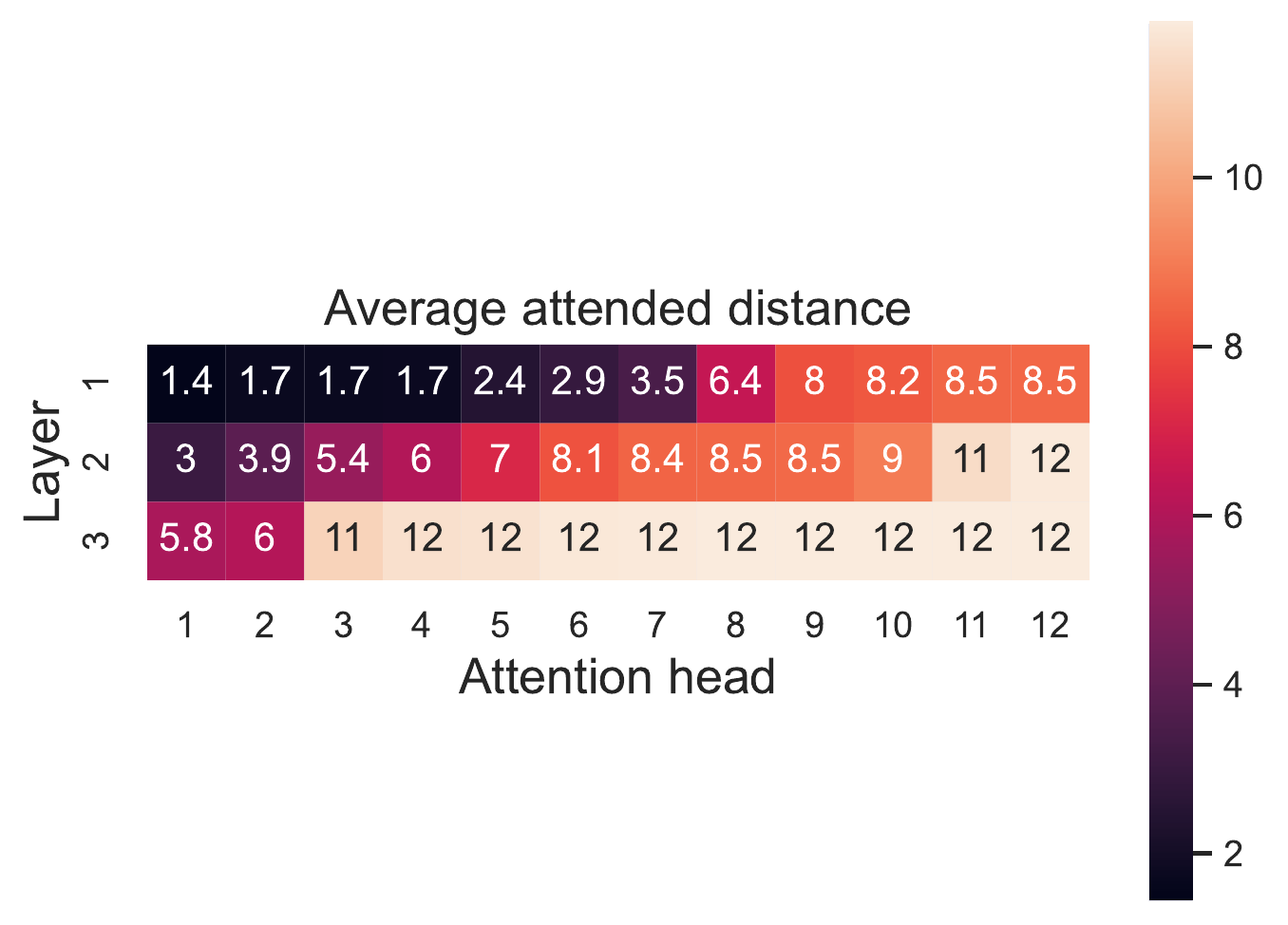}
\caption{12 attention heads and 3 layers, \dataset{PCFG} dataset.}
\label{fig:attn-dist-a12l3-pcfg}
\end{subfigure}
\hfill
\begin{subfigure}[b]{0.48\textwidth}
\centering
\includegraphics[width=\textwidth]{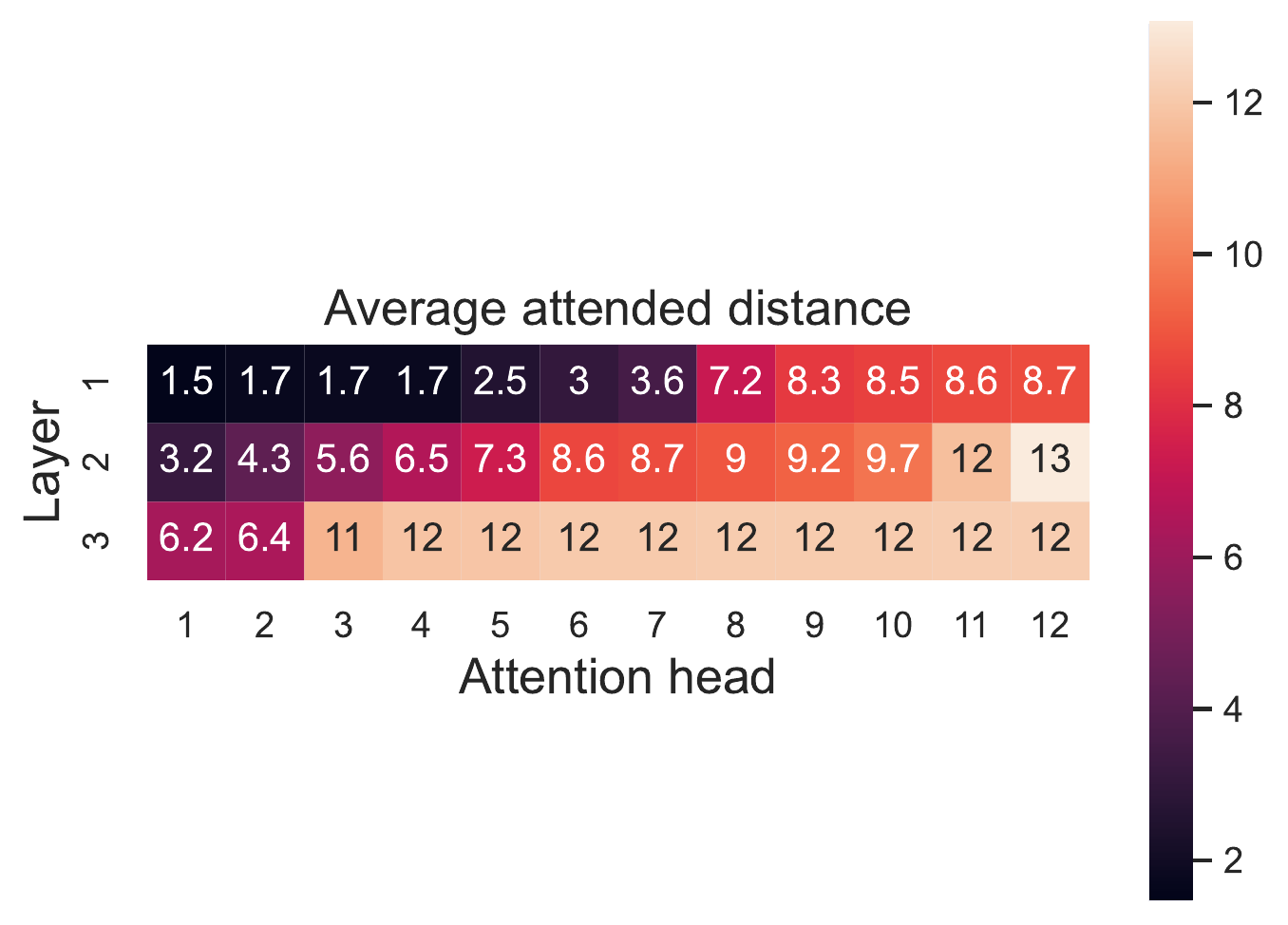}
\caption{12 attention heads and 3 layers, \dataset{PTB} dataset.}
\label{fig:attn-dist-a12l3-ptb}
\end{subfigure}

\caption{``Averaged attented distance'' of each attention heads for different models on \dataset{PCFG} and \dataset{PTB} datasets. \cref{fig:attn-dist-a12l12-pcfg,fig:attn-dist-a12l6-pcfg,fig:attn-dist-a12l3-pcfg} show the results on the synthetic \dataset{PCFG} dataset, and \cref{fig:attn-dist-a12l12-ptb,fig:attn-dist-a12l6-ptb,fig:attn-dist-a12l3-ptb} show the results on the \dataset{PTB} dataset.}
\label{fig:attn-dist}
\end{figure*}

\paragraph{Experiment results} \Cref{fig:attn-dist} shows the results of the ``Averaged attented distance'' for each attention head in different models. \cref{fig:attn-dist-a12l12-pcfg,fig:attn-dist-a12l6-pcfg,fig:attn-dist-a12l3-pcfg} show the results on the synthetic \dataset{PCFG} dataset, and \cref{fig:attn-dist-a12l12-ptb,fig:attn-dist-a12l6-ptb,fig:attn-dist-a12l3-ptb} show the results on the \dataset{PTB} dataset. We sort the attention heads in each layer according to the ``Averaged attended distance''.

From \cref{fig:attn-dist-a12l12-pcfg,fig:attn-dist-a12l6-pcfg,fig:attn-dist-a12l3-pcfg}, we can find that for all models, there are several attention heads in the first layer that look at very close tokens (``Averaged attended distance'' less than $3$). Then as the layer increases, the ``Averaged attended distance'' also increases in general, meaning that the attention heads are looking at further tokens. Then at some layer, there are some attention heads looking at very far tokens (``Averaged attended distance'' larger than 12).\footnote{Note that the average length of the sentences in the synthetic \dataset{PCFG} dataset is around 24, if the attention head gives 0.5 attention score to the first and the last token for every token, the ``Averaged attended distance'' will be 12.} This finding also gives some implication that the model is doing something that correlates with our construction: it looks longer spans as the layer increases. However, different from our construction that the attention head only looks at a fixed length span, models trained using MLM look at different lengths of spans at each layer, which cannot be explained by our current construction, and suggests a further understanding of the mechanism of large language models.

Besides, we can find that the patterns are nearly the same for the synthetic \dataset{PCFG} dataset and \dataset{PTB} dataset, and thus the previous finding can also be transferred to the \dataset{PTB} dataset.

\section{Missing Proofs in \Cref{sec:construction}}
In this section, we show the detailed proof for \Cref{thm:hard_attnt}, \Cref{thm:soft_attnt}, and \Cref{thm:io-optimal-mlm}.

\subsection{Proof of \Cref{thm:hard_attnt}}\label{sec:hard_attnt_proof}
\begin{proof} 
The first $L-1$ layers simulate the recursive formulation of the Inside probabilities from eq.~\ref{eq:inside_probability}, and the last $L-1$ layers simulate the recursive formulation of the outside probabilities from  eq.~\ref{eq:outside_probability}. The model uses embeddings of size $4|\gN| L + L$, where the last $L$ coordinates serve as one-hot positional embeddings and are kept unchanged throughout the model. 

\paragraph{Notations:} For typographical simplicity, we will divide our embeddings into 5 sub-parts. We will use the first $2 |\gN| L$ coordinates to store the inside probabilities, the second $2 |\gN| L$ coordinates to store the outside probabilities, and the final $L$ coordinates to store the one-hot positional encodings. For every position $i$ and span length $\ell+1$, we store the inside probabilities $\{\alpha(A, i, i+\ell)\}_{A \in \gN}$ after computation in its embedding at coordinates $[|\gN|\ell, |\gN|(\ell+1))$. Similarly
we store $\{\alpha(A, i-\ell, i)\}_{A \in \gN}$ at $[|\gN|(L+\ell), |\gN|(L+\ell+1))$, $\{\beta(A, i, i+\ell)\}_{A \in \gN}$ at $[|\gN|(2L+\ell), |\gN|(2L+\ell+1))$, and $\{\beta(A, i-\ell, i)\}_{A \in \gN}$ at $[|\gN|(3L+\ell), |\gN|(3L+\ell+1))$ respectively. For simplicity of presentation, we won't handle cases where $i+\ell$ or $i-\ell$ is outside the range of $1$ to $L$ - those coordinates will be fixed to 0. 


\paragraph{Token Embeddings:} The initial embeddings for each token $w$ will contain $\Pr[A \rightarrow w]$ for all $A \in \gP$. This is to initiate the inside probabilities of all spans of length $1$. 
Furthermore, the tokens will have a one-hot encoding of their positions in the input in the last $L$ coordinates. 


\paragraph{Inside probabilities:} The contextual embeddings at position $i$ after the computations of any layer $\ell < L$ contains the inside probabilities of all spans of length at most $\ell + 1$ starting and ending at position $i$, i.e.  $\alpha(A, i, i + k)$ and $\alpha(A, i-k, i)$ for all $A \in \gN$ and $k \le \ell$. The rest of the coordinates, except the position coordinates, contain $0$.




\paragraph{Layer $1 \le \ell < L$: }
At each position $i$, this layer computes the inside probabilities of spans of length $\ell+1$ starting and ending at $i$, using the recursive formulation from eq.~\ref{eq:inside_probability}. 

For every non-terminal $A \in \gN$, we will use a unique attention head to compute $\alpha(A, i, i + \ell)$ at each token $i$. Specifically, the attention head representing non-terminal $A \in \gN$ will represent the following operation at each position $i$:  

{\small
\begin{align}
    &\alpha(A, i, j) \nonumber \\
    =&  \sum_{B, C \in \gN} \sum_{k=i}^{j-1}\Pr[A \rightarrow B C]  \cdot \alpha(B, i, k) \cdot \alpha(C, k+1, j) \nonumber \\
    =& \sum_{B, C \in \gN} \sum_{ \substack{\ell_1, \ell_2 \ge 0\\ \ell_1 + \ell_2 = \ell-1}
     }\Pr[A \rightarrow B C] \nonumber \\
     &\quad \cdot \alpha(B, i, i+\ell_1) \cdot \alpha(C, j-\ell_2, j) \label{eq:construction-inside-computation},
\end{align}
}

where $j = i+\ell$. In the final step, we modified the formulation to represent the interaction of spans of different lengths starting at $i$ and ending at $j$. We represent this computation as the attention score $a_{i, j}$ using a key matrix $\mK_{A}^{(\ell)}$ and query matrix $\mQ_{A}^{(\ell)}$.

\paragraph{Computing Eq.~\ref{eq:construction-inside-computation}} We set the Key matrix $\mK_{A}^{(\ell)}$ as $\mI$. The Query matrix  $\mQ_{A}^{(\ell)}$ is set such that if we define $\mP_A\in \R^{|\gN|\times |\gN|}$ that contains $\{\Pr[A \to BC]\}_{B,C \in \gN},$ $\mP_A$ appears at positions $(|\gN| (L + \ell_2), |\gN|  \ell_1 )$ for all $\ell_1, \ell_2 \ge 0$ with $\ell_1 + \ell_2 = \ell - 1$. Finally, $\mQ_{A}^{(\ell)}$ contains $\mQ_p\in \R^{L \times L} $ at position $(4|\gN|L, 4|\gN|L)$, such that $\mQ_p[i,i+\ell] = 0$ for $0 \le i < L$, with the rest set to $-\zeta$ for some large constant $\zeta$. The rest of the blocks are set as $0$. We give an intuition behind the structure of $\mQ_{A}^{(\ell)}$ below.

\paragraph{Intuition behind $\mQ_{A}^{(\ell)}$:} For any position $i$ and range $\ell_1 \le \ell$,  $\ve_i^{(\ell-1)}$ contains the inside probabilities $\{ \alpha(C, i - \ell_1, i) \}_{C \in \gN}$ in the coordinates $[|\gN| (L+\ell_1), |\gN| (L+\ell_1+1) )$, while it contains the inside probabilities $\{ \alpha(B, i, i + \ell_1) \}_{B \in \gN}$ in the coordinates $[|\gN| \ell_1, |\gN| (\ell_1+1) ).$ Hence, if we set 
the block  at position $(|\gN| (L + \ell_2), |\gN| \ell_1)$ in $\mQ_{A}^{(\ell)}$
to $\mP_A$ for some $0 \le \ell_1, \ell_2 \le \ell$, with the rest set to $0$, we can get for any two positions $i, j$,

{\small
\begin{align*}
    & (\mK_{A}^{(\ell)} \ve_j^{(\ell-1)})^{\top}  \mQ_{A}^{(\ell)} \ve_i^{(\ell-1)}\\
    =&  \sum_{B, C \in \gN}   \Pr[A \to B C] \cdot \alpha(B, i, i+\ell_1) \cdot  \alpha (C, j -\ell_2, j) .
\end{align*}
}

Because we want to involve the sum over all $\ell_1, \ell_2$ pairs with $\ell_1 + \ell_2 = \ell - 1$, we will set blocks at positions $\{(|\gN| (L + \ell_2), |\gN| \ell_1 )\}_{\ell_1, \ell_2 : \ell_1 + \ell_2 = \ell-1}$ to $\mP_A$, while setting the rest to $0$. This gives us

{\small
\begin{align*}
    &(\mK_{A}^{(\ell)} \ve_j^{(\ell-1)})^{\top} \mQ_{A}^{(\ell)} \ve_i^{(\ell-1)} \\
    =& \sum_{B, C \in \gN} \sum_{\substack{\ell_1, \ell_2 \ge 0\\ \ell_1 + \ell_2 = \ell-1}}  \Pr[A \to B C] \cdot \alpha(B, i, i+\ell_1) \\
    &\quad \cdot  \alpha (C, j -\ell_2, j) .
\end{align*}
}

However, we want $(\mK_{A}^{(\ell)} \ve_j^{(\ell-1)})^{\top} \mQ_{A}^{(\ell)} \ve_i^{(\ell-1)}$ to compute $\alpha(A, i, j)$ iff $j = i + \ell$ and $0$ otherwise, so we will use the final block in $\mQ_{A}^{(\ell)}$ that focuses on the one-hot position encodings of $i$ and $j$ to differentiate the different location pairs. Specifically, the final block $\mQ_p$ will return $0$ if $j = i + \ell$, while it returns $-\zeta$ for some large constant $\zeta$ if $j \ne i + \ell$. This gives us

{\small
\begin{align}
    &(\mK_{A}^{(\ell)} \ve_j^{(\ell-1)})^{\top} \mQ_{A}^{(\ell)} \ve_i^{(\ell-1)} \nonumber \\
    =&  \zeta(\mathbb{I}[j - i = \ell] - 1) + \sum_{B, C \in \gN} \sum_{\substack{\ell_1, \ell_2 \ge 0\\ \ell_1 + \ell_2 = \ell-1}}  \Pr[A \to B C] \nonumber \\
    &\quad \cdot \alpha(B, i, i+\ell_1) \cdot  \alpha (C, j -\ell_2, j). \label{eq:attnt_head_inside_construct}
\end{align}
}
With the inclusion of the term $\zeta(\mathbb{I}[j - i = \ell ] - 1)$, we make $(\mK_{A}^{(\ell)} \ve_j^{(\ell-1)})^{\top} \mQ_{A}^{(\ell)} \ve_i^{(\ell-1)}$ positive if $j - i = \ell$, and negative if $j - i \ne \ell$. Applying a ReLU activation on top will zero out the unnecessary terms, leaving us with $\alpha(A, i, i+\ell)$ at each location $i$.



Similarly, we use another $|\gN|$ attention heads to compute $\alpha(A,i-\ell, i)$. In the end, we use the residual connections to copy the previously computed inside probabilities $\alpha(A,i-\ell', i)$ and $\alpha(A,i, i+\ell')$ for $\ell' < \ell$.





\paragraph{Outside probabilities:}


In addition to all the inside probabilities, the contextual embeddings at position $i$ after the computations of any layer $(2L - 1) - \ell$ ($\ge L$) contain the outside probabilities of all spans of length at least $\ell + 1$ starting and ending at position $i$, i.e.  $\beta(A, i, i + k)$ and $\beta(A, i - k, i)$ for all $A \in \gN$ and $k \ge \ell $. The rest of the coordinates, except the position coordinates, contain $0$.

\paragraph{Layer $L$}
In this layer, we initialize the outside probabilities $\beta(\text{ROOT}, 1, L) = 1$ and $\beta(A, 1, L) = 0$ for $A\neq \text{ROOT}$. Furthermore, we move the inside probabilities $\alpha(A,i+1,i+k)$ from position $i+1$ to position $i$, and $\alpha(A,i-k,i-1)$ from position $i-1$ to position $i$ using 2 attention heads. 

\paragraph{Layer $L + 1 \le \tilde{\ell} := (2L - 1) - \ell \le 2L - 1$:} 
At each position $i$, this layer computes the outside probabilities of spans of length $\ell + 1$ starting and ending at $i$, using the recursive formulation from eq.~\ref{eq:outside_probability}. The recursive formulation for $\beta(A, i, i + \ell)$ for a non-terminal $A \in \gN$ has two terms, given by

{\small
\begin{align}
     \beta(A,i,j) =& \beta_1(A,i,j) + \beta_2(A, i, j), \text{ with   } \nonumber \\
    \beta_1(A,i,j) =& \sum_{C,B \in \gN} \sum_{k=1}^{i-1} \Pr[B \to C A] 
    \nonumber \\
    &\quad\cdot \alpha(C, k, i-1) \beta(B, k, j), \text{ and } \label{eq:beta1} \\
    \beta_2(A, i, j) =& \sum_{B,C \in \gN} \sum_{k=j+1}^{L} \Pr[B \to A C] \nonumber \\
    &\quad \cdot \alpha(C, j+1, k) \beta(B, i, k), \label{eq:beta2}
\end{align}
}

where $j = i + \ell.$ For each non-terminal $A \in \gN$, we will use two unique heads to compute $\beta(A, i, i+\ell)$
, each representing one of the two terms in the above formulation. We outline the construction for $\beta_1$; the construction for $\beta_2$ follows similarly.

\paragraph{Computing Eq.~\ref{eq:beta1}} We build the attention head in the same way we built the attention head to represent the inside probabilities in eq.~\ref{eq:attnt_head_inside_construct}. Similar to \ref{eq:attnt_head_inside_construct}, we modify the formulation of $\beta_1$ to highlight the interaction of spans of different lengths.

{\small
\begin{align}
    \beta_1(A,i,j) =& \sum_{B, C \in \gN} \sum_{\substack{\ell_1, \ell_2 \ge 0\\ \ell_2 - \ell_1 = \ell }} \Pr[B \to C A] \nonumber \\
    & \quad \cdot \alpha(C, i-\ell_1, i-1) \beta(B, j-\ell_2, j), \label{eq:construction-beta1}
\end{align}
}

where $j = i+\ell$. We represent this computation as the attention score $a_{i, i+\ell}$ using a key matrix $\mK_{A, 1}^{(\tilde{\ell})}$ and query matrix $\mQ_{A, 1}^{(\tilde{\ell})}$. 
First, we set the Key matrix $\mK_{A, 1}^{(\tilde{\ell})}$ as $\mI$. If we define $\mP_{A, r} \in \R^{|\gN|\times |\gN|}$ as a matrix that contains $\{\Pr[B \to CA]\}_{B,C \in \gN},$ which is the set of all rules where $A$ appears as the right child, $\mQ_{A, 1}^{(\tilde{\ell})}$ is set such that $\mP_{A, r}$ appears at positions $[|\gN| (3L + \ell_2), |\gN| (L + \ell_1))$ for all $0 \le \ell_1, \ell_2 \le L$ that satisfy $\ell_2 - \ell_1 = \ell$. Finally, $\mQ_{A, 1}^{(\tilde{\ell})}$ contains $\mQ_p\in \R^{L \times L} $ at position $(4|\gN|L, 4|\gN|L)$, such that $\mQ_p[i,i+\ell] = 0$ for $0 \le i < L$, with the rest set to $-\zeta$ for some large constant $\zeta$. The rest of the blocks are set as $0$. We give an intuition behind the structure of $\mQ_{A,1}^{(\tilde{\ell})}$ below.

\paragraph{Intuition for $\mQ_{A, 1}^{(\tilde{\ell})}$:}  For position $i$ and any ranges $1 \le \ell_1 < L$, $\ell+1 \le \ell_2 \le L$, $\ve_i^{( \tilde{\ell} - 1 )}$ contains the inside probabilities $\{ \alpha(C, i - \ell_1, i-1) \}_{C \in \gN}$ in the coordinates $[ |\gN| (L+\ell_1), |\gN| (L+\ell_1+1) )$, while it contains the outside probabilities $\{ \beta(B, i - \ell_2, i) \}_{B \in \gN}$ in the coordinates $[|\gN| (3L+\ell_2), |\gN| (3L+\ell_2+1) ).$ Hence, if we set the block at position $(|\gN| (3L + \ell_2), |\gN| (L + \ell_1))$
to $\mP_A$ for some $0 \le \ell_1 \le L, \ell+1 \le \ell_2 \le L$, with the rest set to $0$, we can get for any two positions $i, j$,

{\small
\begin{align*}
    & (\mK_{A}^{(\tilde{\ell})} \ve_j^{(\tilde{\ell}-1)})^{\top}  \mQ_{A}^{(\tilde{\ell})} \ve_i^{(\tilde{\ell}-1)} \\
    =&  \sum_{B, C \in \gN}   \Pr[B \to CA] \cdot \alpha(C, i-\ell_1, i-1) \cdot  \beta (B, j -\ell_2, j) .
\end{align*}
}


Because we want to include the sum over $\ell_1, \ell_2$ pairs with $ \ell_2 - \ell_1 = \ell$, we will only set blocks at positions $[|\gN| (3L + \ell_2), |\gN| (L + \ell_1))$ for all $0 \le \ell_1, \ell_2 \le L$ that satisfy $\ell_2 - \ell_1 = \ell$ to $\mP_{A, r}$, while setting the rest to $0$. This gives us

\begin{align*}
    &(\mK_{A}^{(\tilde{\ell})} \ve_j^{(\tilde{\ell}-1)})^{\top}  \mQ_{A}^{(\tilde{\ell})} \ve_i^{(\tilde{\ell}-1)} \\
    =& \sum_{B, C \in \gN}  \sum_{\substack{\ell_1, \ell_2 \ge 0 \\ \ell_2 - \ell_1 = \ell}}  \Pr[B \to CA] \\
    &\quad\cdot \alpha(C, i-\ell_1, i-1) \cdot  \beta (B, j -\ell_2, j).
\end{align*}

Because we want $(\mK_{A}^{(\tilde{\ell})} \ve_j^{(\tilde{\ell}-1)})^{\top}  \mQ_{A}^{(\tilde{\ell})} \ve_i^{(\tilde{\ell}-1)}$ to compute $\beta_1(A, i, j)$ with $j = i + \ell$ and $0$ otherwise, we will use the final block in $\mQ_{A}^{(\ell)}$ that focuses on the one-hot position encodings of $i$ and $j$ to differentiate the different location pairs. Specifically, the final block $\mQ_p$ will return $0$ if $j = i + \ell$, while it returns $-\zeta$ for some large constant $\zeta$, if $j \ne i + \ell$. This gives us

{\small
\begin{align*}
    & (\mK_{A}^{(\tilde{\ell})} \ve_j^{(\tilde{\ell}-1)})^{\top}  \mQ_{A}^{(\tilde{\ell})} \ve_i^{(\tilde{\ell}-1)} \\
    =& \zeta (\mathbb{I}[j - i = \ell ] - 1) + \sum_{B, C \in \gN}  \sum_{\substack{\ell_1, \ell_2 \ge 0\\ \ell_2 - \ell_1 = \ell} } \Pr[B \to CA] \\
    &\quad \cdot \alpha(C, i-\ell_1, i-1) \cdot  \beta (B, j -\ell_2, j) 
\end{align*}
}

With the inclusion of the term $\zeta(\mathbb{I}[j - i = \ell ] - 1)$, we make $(\mK_{A}^{(\tilde{\ell})} \ve_j^{(\tilde{\ell}-1)})^{\top}  \mQ_{A}^{(\tilde{\ell})} \ve_i^{(\tilde{\ell}-1)}$ positive if $j - i = \ell$, and negative if $j - i \ne \ell$. Applying a ReLU activation on top will zero out the unnecessary terms, leaving us with $\beta_1(A, i, i+\ell)$ at each location $i$.

Besides, we also need $2|\gN|$ additional heads for the outside probabilities $\beta(A,i-\ell,i)$. In the end, we use the residual connections to copy the previously computed inside probabilities $\beta(A, i-\ell', i)$ and $\alpha(A, i, i+\ell')$ for $\ell' > \ell$.
\end{proof}

\subsection{Proof of \Cref{thm:soft_attnt}}\label{sec:soft_attnt_proof}

Similar to the proof of \cref{thm:hard_attnt}, the first $L-1$ layers simulate the recursive formulation of the Inside probabilities from eq.~\ref{eq:inside_probability}, and the last $L-1$ layers simulate the recursive formulation of the  outside probabilities from  eq.~\ref{eq:outside_probability}. The model uses embeddings of size $2|\gN| L$ and uses $4L+2$ relative position embeddings.

\paragraph{Notations:} For typographical simplicity, we will divide our embeddings into 2 sub-parts. We will use the first $|\gN| L$ coordinates to store the inside probabilities, and the second $|\gN| L$ coordinates to store the outside probabilities. For every position $i$ and span length $\ell+1$, we store the inside probabilities $\{\alpha(A, i-\ell, i)\}_{A \in \gN}$ after computation in its embedding at coordinates $[|\gN|\ell, |\gN|(\ell+1))$, where the coordinates for embeddings start from $0$. Similarly
we store $\{\beta(A, i, i+\ell)\}_{A \in \gN}$ at $[|\gN|(L+\ell), |\gN|(L+\ell+1))$. For simplicity of presentation, we won't handle cases where $i+\ell$ or $i-\ell$ is outside the range of $1$ to $L$ - those coordinates will be fixed to 0. 


\paragraph{Token Embeddings:} The initial embeddings for each token $w$ will contain $\Pr[A \rightarrow w]$ for all $A \in \gP$. This is to initiate the inside probabilities of all spans of length $1$. 

\paragraph{Relative position embeddings:} We introduce $2L + 1$ relative position vectors $ \{ p_{ t } \in \mathbb{R}^{2 |\gN| L}  \}_{-L \le t \le L},$ that modify the key vectors depending on the relative position of the query and key tokens. Furthermore, we introduce $(2L-1)L$ relative position-dependent biases $ \{ b_{ t, \ell } \in \mathbb{R}  \}_{-L \le t \le L, 1 \le \ell \le 2L-1 }.$ We introduce the structures of the biases in the contexts of their intended uses.

\paragraph{Structure of $\{ p_{ t } \}_{-L \le t \le L}$:} 
For $t < 0$, we define $p_{t}$ such that all coordinates in $[|\gN| (-t-1), |\gN| (-t) )$ are set to $1$, with the rest set to $0$. For $t > 0$, we define $p_{t}$ such that all coordinates in $[|\gN| (L+t-1), |\gN| (L + t) )$ are set to $1$, with the rest set to $0$. $p_0$ is set as all $0s$.

\paragraph{Attention formulation:} At any layer $1 \le \ell \le 2L-1$ except $L$, we define the attention score  $a_{i, j}^h$  between $\ve_i^{(\ell-1)}$ and $\ve_j^{(\ell-1)}$ for any head $h$ with Key and Query matrices $\mK^{(\ell)}_h$ and $\mQ^{(\ell)}_h$ as

{\small
\begin{equation}
    a_{i, j}^h =  \text{ReLU}( \mK^{(\ell)}_h \ve_j^{(\ell-1)} + p_{j - i} - b_{j - i, \ell} )^\top \mQ^{(\ell)}_h \ve_i^{(\ell-1)} \label{eq:soft_attention_appnd}.
\end{equation}
}

For layer $L$, we do not use the relative position embeddings, i.e. we define the attention score  $a_{i, j}^h$  between $\ve_i^{(L-1)}$ and $\ve_j^{(L-1)}$ for any head $h$ with Key and Query matrices $\mK^{(L)}_h$ and $\mQ^{(L)}_h$ as

{\small
\begin{equation}
    a_{i, j}^h =  \text{ReLU}( \mK^{(L-1)}_h \ve_j^{(L-1)} - b_{j - i, L} )^\top \mQ^{(\ell)}_h \ve_i^{(L-1)} \label{eq:soft_attention_appnd_L}.
\end{equation}
}

\paragraph{Inside probabilities:} The contextual embeddings at position $i$ after the computations of any layer $\ell < L$ contains the inside probabilities of all spans of length at most $\ell + 1$ ending at position $i$, i.e.  $\alpha(A, i-k, i)$ for all $A \in \gN$ and $k \le \ell$. The rest of the coordinates contain $0$.

\paragraph{Structure of $\{ b_{ t, \ell } \}_{-L \le t \le L, 1 \le \ell \le L-1}$:} For any $1 \le \ell \le L-1$, for all $t \ge 0$ and $t < -\ell$, we set $b_{t, \ell}$ as $\zeta$ for some large constant $\zeta$. All other biases are set as $1$.




\paragraph{Layer $1 \le \ell < L$: }
At each position $i$, this layer computes the inside probabilities of spans of length $\ell+1$ ending at $i$, using the recursive formulation from eq.~\ref{eq:inside_probability}. 

For every non-terminal $A \in \gN$, we will use a unique attention head to compute $\alpha(A, i - \ell, i)$ at each token $i$. Specifically, the attention head representing non-terminal $A \in \gN$ will represent the following operation at each position $i$:  

{\small
\begin{align}
    &\alpha(A, i-\ell, i) \nonumber \\
    =&  \sum_{B, C \in \gN} \sum_{j=i-\ell}^{i-1} \Pr[A \rightarrow B C] \alpha(B, i-\ell, j) \alpha(C, j+1, i) \nonumber \\
    =& \sum_{j=i-\ell}^{i-1} \sum_{B, C \in \gN}  \Pr[A \rightarrow B C]   \alpha(B, i-\ell, j) \alpha(C, j+1, i). \label{eq:construction-inside-computation-soft-attn}
\end{align}
}

 In the final step, we swapped the order of the summations to observe that the desired computation can be represented as a sum over individual computations at locations $j < i$.  That is, we represent $\sum_{B, C \in \gN}  \Pr[A \rightarrow B C]  \cdot \alpha(B, i-\ell, j) \cdot \alpha(C, j+1, i)$ as the attention score $a_{i,  j}$ for all $i-\ell \le j \le i$, while $\alpha(A, i-\ell, i)$ will be represented as $\sum_{i-\ell \le j < i-1} a_{i,  j}.$



\paragraph{Structure of $\mQ_{A}^{(\ell)}$ and $\mK_A^{(\ell)}$ to compute Eq.~\ref{eq:construction-inside-computation-soft-attn}:} 
\begin{enumerate}
    \item $\mK_{A}^{(\ell)}$ is a rotation matrix such that in $\mK_{A}^{(\ell)} \ve_i^{(\ell)}$, for all $\ell_1 \le \ell$, the inside probabilities $ \{ \alpha(B, i-\ell_1, i) \}_{B \in \gN}$ appears in the coordinates $[ |\gN| (\ell - \ell_1),  |\gN| (\ell - \ell_1+1) )$. Note that $\mK_A^{(\ell)}$ are the same for different $A$, and only depend on $\ell$.
    \item The Query matrix  $\mQ_{A}^{(\ell)}$ is a block diagonal matrix,  such that if we define $\mP_A\in \R^{|\gN|\times |\gN|}$ that contains $\{\Pr[A \to BC]\}_{B, C \in \gN},$ $\mP_A$ appears in the first $\ell$ blocks along the diagonal, i.e. it occurs at all positions starting at $( |\gN| \ell_1,  |\gN| \ell_1 )$ for all $\ell_1 < \ell$. The rest of the blocks are set as $0$s.
\end{enumerate}


\paragraph{Intuition behind $\mQ_{A}^{(\ell)}$, $\mK_{A}^{(\ell)}$, the relative position embeddings and the biases:} For any position $i$ and range $\ell_1 < \ell$,  $\ve_i^{(\ell-1)}$ contains the inside probabilities $\{ \alpha(C, i - \ell_1, i) \}_{C \in \gN}$ in the coordinates $[|\gN| \ell_1, |\gN| (\ell_1+1) )$. With the application of $\mK_{A}^{(\ell)}$, $\mK_{A}^{(\ell)} \ve_i^{(\ell-1)}$ contains the inside probabilities $\{ \alpha(C, i - \ell_1, i) \}_{C \in \gN}$ in the coordinates $[|\gN| (\ell - 1 - \ell_1), |\gN| (\ell - \ell_1) ).$
Hence, if we set 
the block  at position $(|\gN| \ell_1, |\gN| \ell_1)$ in $\mQ_{A}^{(\ell)}$
to $\mP_A$ for some $0 \le \ell_1 < \ell$, with the rest set to $0$, we can get for any two positions $i, j$,

{\small
\begin{align*}
    & (\mK_{A}^{(\ell)} \ve_j^{(\ell-1)})^{\top}  \mQ_{A}^{(\ell)} \ve_i^{(\ell-1)} \\
    = & \sum_{B, C \in \gN}   \Pr[A \to B C] \cdot \alpha(B, i-\ell_1, i) \\
    &\quad \cdot  \alpha (C, j - (\ell - 1 - \ell_1), j).
\end{align*}
}

Setting the first $\ell$ diagonal blocks in $\mQ_{A}^{(\ell)}$ to $\mP_A$ can get for any two positions $i, j$,

{\small
\begin{align*}
    & (\mK_{A}^{(\ell)} \ve_j^{(\ell-1)})^{\top}  \mQ_{A}^{(\ell)} \ve_i^{(\ell-1)} \\
    =& \sum_{\ell_1 \le \ell-1} \sum_{B, C \in \gN}   \Pr[A \to B C] \cdot \alpha(B, i-\ell_1, i) \\
    &\quad \cdot  \alpha (C, j - (\ell - \ell_1 - 1), j).
\end{align*}
}

However, for $\alpha(A, i-\ell, i)$, the attention score above should only contribute with  $\ell_1 = i - j - 1$. Moreover, we also want the above sum to be $0$ if $j \ge i$ or $j \le i - \ell - 1$. Hence, we will use the relative position vector $p_{j - i}$, bias $b_{j-i, \ell}$ and the ReLU activation to satisfy the following conditions:
\begin{enumerate}
    \item $i - \ell \le j \le i-1$. 
    \item The portion containing $\{\alpha (C, j - (\ell - \ell_1 - 1), j)\}_{C \in \gN}$ in $\mK_{A}^{(\ell)} \ve_j^{(\ell-1)}$ is activated only if $\ell_1 = i - j - 1$.
\end{enumerate}


 For any positions $i, j$ and $\ell_1 < \ell$, $\mK_{A}^{(\ell)} \ve_j^{(\ell-1)} + p_{j-i} - b_{j-i, \ell}$ will contain $\{ \alpha (C, j - (\ell-\ell_1-1), j) + \mathbb{I} [\ell_1 = i - j - 1]  - 1 - \zeta \mathbb{I} [  j < i - \ell  \text{ or } j > i - 1 ] \}_{C \in \gN}$ in coordinates $[|\gN| \ell_1, |\gN| (\ell_1 + 1) )$, which will give us

 {\small
\begin{align*}
     & \text{ReLU}(\mK_{A}^{(\ell)} \ve_j^{(\ell-1)} + p_{j-i} - b_{j-i, \ell})^{\top}  \mQ_{A}^{(\ell)} \ve_i^{(\ell-1)} \\
     =& \sum_{B, C \in \gN} \Pr[A \to B C] \cdot \alpha(B, j+1, i) \cdot  \alpha (C, i-\ell, j),
\end{align*}
}
if $i - \ell \le j \le i-1$ and $0$ otherwise. Summing over all locations $j$ gives us $\alpha(A, i-\ell, i)$.

\paragraph{Outside probabilities:}


In addition to all the inside probabilities, the contextual embeddings at position $i$ after the computations of any layer $(2L - 1) - \ell$ ($\ge L$) contain the outside probabilities of all spans of length at least $\ell + 1$ starting at position $i$, i.e.  $\beta(A, i, i + k)$ for all $A \in \gN$ and $k \ge \ell $. The rest of the coordinates contain $0$.

\paragraph{Layer $L$}
In this layer, we initialize the outside probabilities $\beta(\text{ROOT}, 1, L) = 1$ and $\beta(A, 1, L) = 0$ for $A\neq \text{ROOT}$. Furthermore, we move the inside probabilities $\alpha(A,i-k,i-1)$ from position $i-1$ to position $i$ using 1 attention head. For the attention head, $b_{-1, L}$ is set as $0$, while the rest are set as $\zeta$ for some large constant $\zeta$ so that the attention heads only attend to position $i-1$ at any position $i$.

\paragraph{Layer $L + 1 \le \tilde{\ell} := (2L - 1) - \ell \le 2L - 1$:} 
At each position $i$, this layer computes the outside probabilities of spans of length $\ell + 1$ starting at $i$, using the recursive formulation from eq.~\ref{eq:outside_probability}. The recursive formulation for $\beta(A, i, i + \ell)$ for a non-terminal $A \in \gN$ has two terms, given by

{\small
\begin{align}
     \beta(A,i,i+\ell)  =& \beta_1(A,i,i+\ell) + \beta_2(A, i, i+\ell), \text{ with   } \label{eq:beta1_beta2_soft}\\
    \beta_1(A,i,i+\ell) =& \sum_{j=1}^{i-1}  \sum_{C,B \in \gN} \Pr[B \to C A] \nonumber \\
    &\quad\cdot \alpha(C, j, i-1) \beta(B, j, i+\ell), \text{ and } \label{eq:beta1_soft} \\
    \beta_2(A, i, i+\ell) =& \sum_{j=i+\ell+1}^{L} \sum_{B,C \in \gN}  \Pr[B \to A C] \nonumber \\
    &\quad \cdot \alpha(C, i+\ell+1, j) \beta(B, i, j). \label{eq:beta2_soft}
\end{align}
}

For each non-terminal $A \in \gN$, we will use a single unique head to compute $\beta(A, i, i+\ell)$ with query matrix $\mQ_{A}^{(\Tilde{\ell})}$ and key matrix $\mK_{A}^{(\Tilde{\ell})}$. Combining the operations of both $\beta_1$ and $\beta_2$ in a single attention head is the main reason behind the decrease in the number of necessary attention heads, compared to \cref{thm:hard_attnt}.


\paragraph{Structure of $\{ b_{ t, \ell } \}_{-L \le t \le L, L+1 \le \ell \le 2L-1}$:} For any $L+1 \le \ell \le 2L-1$, for $0 \le t \le \ell + 1$, $b_{t, \ell}$ is set as $\zeta$ for some large constant $\zeta$. 
All other biases are set as $1$.

\paragraph{Structure of Query and key matrices:}    
\begin{enumerate}
    \item $\mK_{A}^{(\tilde\ell)}$ is a rotation matrix such that in $\mK_{A}^{(\tilde\ell)} \ve_i^{(\ell)}$, for all $L > \ell_1 > \ell$, the outside probabilities $ \{ \beta(B, i, i+\ell_1) \}_{B \in \gN}$ appears in the coordinates $[ |\gN| (\ell_1 -\ell - 1),  |\gN| (\ell_1 -\ell) )$.  Furthermore, for all $0 \le \ell_1 \le L - \ell - 2$, the inside probabilities $ \{ \alpha(C, i-1-\ell_1, i-1) \}_{C \in \gN}$ appears in the coordinates $[ |\gN| (L + \ell + \ell_1 + 1 ),  |\gN| (L + \ell + \ell_1 + 2) )$. Note that $\mK_A^{(\tilde\ell)}$ is same for all $A$, and only depends on $\ell$.
    
    \item The Query matrix  $\mQ_{A}^{(\tilde\ell)}$ is a block diagonal matrix. If we define $\mP_{A, r} \in \R^{|\gN|\times |\gN|}$ as a matrix that contains $\{\Pr[B \to CA]\}_{B,C \in \gN},$ which is the set of all rules where $A$ appears as the right child, $\mP_{A, r}$ appears at positions $(|\gN| \ell_1, |\gN| \ell_1)$ for all $\ell_1 < L$, which is the set of the first $L$ blocks along the diagonal. Furthermore, if we define $\mP_{A, l} \in \R^{|\gN|\times |\gN|}$ as a matrix that contains $\{\Pr[B \to AC]\}_{B,C \in \gN},$ which is the set of all rules where $A$ appears as the left child, $\mP_{A, l}^{\top}$ appears at positions $(|\gN| \ell_1, |\gN| \ell_1)$ for all $\ell_1 \ge L+\ell+1$, which is a set of $L-\ell-2$ blocks along the diagonal located towards the end.
    
\end{enumerate}

\paragraph{Intuition behind $\mQ_{A}^{(\tilde\ell)}$, $\mK_{A}^{(\tilde\ell)}$, the relative position embeddings and the biases:} Considering any location $i$, we split the computation of $\beta(A, i, i+\ell)$ with the attention head into the computation of $\beta_1$ (eq.~\ref{eq:beta1_soft}) and $\beta_2$ (eq.~\ref{eq:beta2_soft}). For $\beta_1$, we express each term $\sum_{C,B \in \gN} \Pr[B \to C A] \alpha(C, j, i-1) \beta(B, j, i+\ell)$ as the attention score $a_{i, j}$ and then express $\beta_1$ as $\sum_{j \le i-1} a_{i, j}$. Similarly, for $\beta_2$, we express each term $\sum_{B,C \in \gN}  \Pr[B \to A C] \alpha(C, i+\ell+1, j) \beta(B, i, j)$ as the attention score $a_{i, j}$ and then express $\beta_1$ as $\sum_{j \ge i+\ell+1} a_{i, j}$. The relative position vectors and biases help to differentiate the operations on the left and right-hand sides of $i$, as we showcase below.

\subparagraph{Computing $\beta_1$ (eq.~\ref{eq:beta1_soft}):}
For any position $i$ and  $\ell_1 \ge 0$,  $\ve_i^{(\tilde\ell-1)}$ contains the inside probabilities $\{ \alpha(C, i - 1 - \ell_1, i-1) \}_{C \in \gN}$ in the coordinates $[|\gN|\ell_1 , |\gN| (\ell_1 + 1) )$. With the application of $\mK_{A}^{(\tilde\ell)}$, for $\ell_1 > \ell$, $\mK_{A}^{(\tilde\ell)} \ve_i^{(\tilde\ell-1)}$  contains the outside probabilities $\{ \beta(B, i,  i + \ell_1) \}_{B \in \gN}$ in the coordinates $[|\gN| (\ell_1 - \ell - 1), |\gN| (\ell_1 - \ell) ).$
Hence, if we set 
the block  at position $(|\gN| \ell_1, |\gN| \ell_1)$ in $\mQ_{A}^{(\ell)}$
to $\mP_{A, r}$ for some $L > \ell_1 \ge 0$, with the rest set to $0$, we can get for any two positions $i, j$,

{\small
\begin{align*}
    & (\mK_{A}^{(\tilde\ell)} \ve_j^{(\tilde\ell-1)})^{\top}  \mQ_{A}^{(\tilde\ell)} \ve_i^{(\tilde\ell-1)} \\
    =& \sum_{B, C \in \gN}   \Pr[B \to CA] \cdot \alpha(C, i-1-\ell_1, i-1) \\
    &\quad \cdot  \beta (B, j, j + \ell + \ell_1 + 1).
\end{align*}
}

Setting the first $L$ diagonal blocks in $\mQ_{A}^{(\tilde\ell)}$ to $\mP_{A, r}$ can get for any two positions $i, j$,

{\small
\begin{align*}
    & (\mK_{A}^{(\tilde\ell)} \ve_j^{(\tilde\ell-1)})^{\top}  \mQ_{A}^{(\tilde\ell)} \ve_i^{(\tilde\ell-1)} \\
    =& \sum_{\ell_1 \ge 0} \sum_{B, C \in \gN}   \Pr[B \to CA] \cdot \alpha(C, i-1-\ell_1, i-1)\\
    &\quad \cdot  \beta (B, j, j + \ell + \ell_1 + 1).
\end{align*}
}

However, for $\beta_1(A, i, i+\ell)$, the attention score above should only contribute with  $\ell_1 = i - j - 1$. Moreover, we also want the above sum to be $0$ if $j \ge i$. Hence, we will use the relative position vector $p_{j - i}$, bias $b_{j-i, \tilde\ell}$ and the ReLU activation to satisfy the following conditions:
\begin{enumerate}
    \item $j < i$. 
    \item The portion containing $\{\beta (B, j, j + \ell + \ell_1 + 1)\}_{C \in \gN}$ in $\mK_{A}^{(\tilde\ell)} \ve_j^{(\tilde\ell-1)}$ is activated only if $\ell_1 = i - j - 1$.
\end{enumerate}


 For any positions $i, j$ and $0 \le \ell_1 \le L$, $\mK_{A}^{(\tilde\ell)} \ve_j^{(\tilde\ell-1)} + p_{j-i} - b_{j-i, \tilde\ell}$ will contain $\{ \beta (B, j, j + \ell + \ell_1 + 1) + \mathbb{I} [\ell_1 = i - j - 1]  - 1 - \zeta \mathbb{I} [  i \le j \le i + \ell ] \}_{B \in \gN}$ in coordinates $[|\gN| \ell_1, |\gN| (\ell_1 + 1) )$, which will give us

{\small
\begin{align*}
     &\text{ReLU}(\mK_{A}^{(\tilde\ell)} \ve_j^{(\tilde\ell-1)} + p_{j-i} - b_{j-i, \tilde\ell})^{\top}  \mQ_{A}^{(\tilde\ell)} \ve_i^{(\tilde\ell-1)} \\ =& \sum_{C,B \in \gN} \Pr[B \to C A] \alpha(C, j, i-1) \beta(B, j, i+\ell),
\end{align*}
}

iff $j < i$ and $0$ otherwise. Summing over all locations gives us $\beta_1(A, i, i+\ell)$.

\subparagraph{Computing $\beta_2$ (eq.~\ref{eq:beta2_soft}):}
For any position $i$ and  $L > \ell_1 > \ell$,  $\ve_i^{(\tilde\ell-1)}$ contains the outside probabilities $\{ \beta(B, i, i+\ell_1) \}_{B \in \gN}$ in the coordinates $[|\gN| (L+\ell_1) , |\gN| (L + \ell_1 + 1) )$. With the application of $\mK_{A}^{(\tilde\ell)}$, for $L > \ell_1 > \ell$, $\mK_{A}^{(\tilde\ell)} \ve_i^{(\tilde\ell-1)}$  contains the inside probabilities $\{ \alpha(C, i-1-\ell_1,  i-1) \}_{C \in \gN}$ in the coordinates $[|\gN| (L + \ell + \ell_1 + 1), |\gN| (L + \ell + \ell_1 + 2) ).$
Hence, if we set 
the block  at position $(|\gN| \ell_1, |\gN| \ell_1)$ in $\mQ_{A}^{(\tilde\ell)}$
to $\mP_{A, l}^\top$ for some $\ell_1 \ge L + \ell + 1$, with the rest set to $0$, we can get for any two positions $i, j$,

{\small
\begin{align*}
    & (\mK_{A}^{(\tilde\ell)} \ve_j^{(\tilde\ell-1)})^{\top}  \mQ_{A}^{(\tilde\ell)} \ve_i^{(\tilde\ell-1)} \\
    =& \sum_{B, C \in \gN}   \Pr[B \to AC] \cdot \alpha(C,  j-\ell_1+\ell+L, j-1) \\
    &\quad \cdot  \beta (B, i, i + \ell_1 - L).
\end{align*}
}

Setting diagonal blocks at positions $\{ (|\gN|\ell_1, |\gN|\ell_1) \}_{\ell_1 \ge L+\ell+1}$ in $\mQ_{A}^{(\tilde\ell)}$ to $\mP_{A, l}^\top$ can get for any two positions $i, j$,

{\small
\begin{align*}
    & (\mK_{A}^{(\tilde\ell)} \ve_j^{(\tilde\ell-1)})^{\top}  \mQ_{A}^{(\tilde\ell)} \ve_i^{(\tilde\ell-1)}\\
    =& \sum_{ \ell_1 \ge \ell + 1 } \sum_{B, C \in \gN}   \Pr[B \to AC] \cdot \alpha(C, j-\ell_1+\ell, j-1) \\
    &\quad \cdot  \beta (B, i, i + \ell_1).
\end{align*}
}

However, for $\beta_1(A, i, i+\ell)$, the attention score above should only contribute with  $\ell_1 = j - i - 1$. Moreover, we also want the above sum to be $0$ if $j \le i+\ell$. 
We will use the relative position vector $p_{j-i}$, bias $b_{j-i, \tilde\ell}$ and the ReLU activation to satisfy the following conditions:
\begin{enumerate}
    \item $j > i+\ell$. 
    \item The portion containing $\{ \alpha(C, j-\ell_1+\ell, j-1) \}_{C \in \gN}$ in $\mK_{A}^{(\tilde\ell)} \ve_j^{(\tilde\ell-1)}$ is activated only if $\ell_1 = j - i - 1$.
\end{enumerate}


 Thus, for any positions $i, j$ and $0 \le \ell_1 \le L$, $\mK_{A}^{(\tilde\ell)} \ve_j^{(\tilde\ell-1)} + p_{j-i} - b_{j-i, \tilde\ell}$ will contain $\{  \alpha(C, j-\ell_1+\ell, j-1) + \mathbb{I} [\ell_1 = i - j - 1]  - 1 - \zeta \mathbb{I} [ i \le j \le i+\ell ] \}_{C \in \gN}$ in coordinates $[|\gN| \ell_1, |\gN| (\ell_1 + 1) )$, which will give us

 {\small
\begin{align*}
     &\text{ReLU}(\mK_{A}^{(\tilde\ell)} \ve_j^{(\tilde\ell-1)} + p_{j-i} - b_{j-i, \tilde\ell})^{\top}  \mQ_{A}^{(\tilde\ell)} \ve_i^{(\tilde\ell-1)}\\
     =& \sum_{j=i+\ell+1}^{L} \sum_{B,C \in \gN}  \Pr[B \to A C] \alpha(C, i+\ell+1, j) \beta(B, i, j),
\end{align*}
}

iff $j > i + \ell + 1$ and $0$ otherwise. Summing over all locations gives us $\beta_2(A, i, i+\ell)$.

\subparagraph{Computing $\beta_1 + \beta_2$ (eq.~\ref{eq:beta1_beta2_soft}):} From our construction, $\beta_1$ requires the dot product of the inside probabilities stored at the query vector and the outside probabilities stored at the key vector. However,  $\beta_2$ requires the dot product of the outside probabilities stored at the query vector and the inside probabilities stored at the key vector. Since $\beta_1$ and $\beta_2$ are computed on the left and the right-hand side of the query respectively, we use the relative position embeddings to separate the two operations. The vector $p_{j-i}$ activates only the outside probabilities in the key vector when $j > i$ and activates only the inside probabilities in the key vector when $j < i$. Thus, we can compute $\beta_1+\beta_2$ as the sum of the attention scores of a single head, where the computation of $\beta_1$ and $\beta_2$ have been restricted to the left and the right-hand side of the query respectively.

\subsection{Proof of \cref{thm:io-optimal-mlm}} \label{sec:proof_io-optimal-mlm}
\begin{proof}[Proof of \Cref{thm:io-optimal-mlm}]
We first focus on 1-mask predictions, where given an input of tokens $w_1, w_2, \cdots, w_L$, and a randomly selected index $i$, we need to predict the token at position $i$ given the rest of the tokens, i.e. $\Pr\{w | w_{-i}\}$. Under the generative rules of the PCFG model, we have

{\small
\begin{align}
    &\Pr[w | w_{-i}] \nonumber \\
    = & \sum_{A}\Pr[A\to w] \cdot \Pr[A\text{ generates word at pos }i | w_{-i}] \nonumber\\
    = & \sum_{A}\Pr[A\to w]\cdot \frac{\beta(A,i,i)}{\sum_{B}\beta(B,i,i)}. \label{eq:1mask_pcfg}
\end{align}
}
Note that $\Pr[A\to w]$ can be extracted from the PCFG and $\{\beta(B, i, i)\}_{B \in \gN}$ can be computed by the Inside-outside algorithm. Thus, Inside-outside can solve the 1-masking problem optimally.

Now we consider the case where we randomly mask $m\%$ (e.g., 15\%) of the tokens and predict these tokens given the rest. In this setting, if the original sentence is generated from PCFG $\gG = (\gN, \gI, \gP, n, p)$, one can modify the PCFG to get $\gG' = (\gN, \gI, \gP, n+1, p')$ with $n+1$ denote the mask token $text{[MASK]}$ and for each preterminal $A\in\gP$, $p'(A\to \text{[MASK]}) = m\%$ and $p'(A\to w) = (1-m\%)p(A\to w),$ for all $w\neq \text{[MASK]}$. Then, the distribution of the randomly masked sentences follows the distribution of sentences generated from the modified PCFG $\gG'$. Similar to the 1-masking setting, we can use the Inside-outside algorithm to compute the optimal token distribution at a masked position.

\end{proof}

\section{Omitted Details in \Cref{sec:approx-overview}}\label{sec:approx-detailed}
In \Cref{sec:approx-overview}, we claim that it is possible to approximately execute the Inside-Outside algorithm for PCFG learned on \dataset{PTB} dataset, and can drastically reduce the size of our constructed model with minimal impact on the 1-masking predictions and parsing performance (\Cref{thm:approx-low-rank-informal}) by applying two ingredients: restricting the computations to few non-terminals and utilizing the underlying low-rank structure between the non-terminals. 
This section is organized as follows: In \Cref{appendix:small_nonterminal_subset}, we show more intuition and experiment results on why we can restrict the computation of the inside-outside algorithm to a small subset of non-terminals. In \Cref{appendix:approx-low-rank-informal}, we add more discussions on the second ingredient (utilizing the low-rank structure). Then in \Cref{sec:approx-few-nt}, we show the details why restricting the computations of few non-terminals can reduce the size of the attention model. In \Cref{sec:approx-low-rank}, we show the detailed proof of \Cref{thm:approx-low-rank-informal}. Finally in \Cref{sec:approx-exp-details}, we show the experiment details in \Cref{sec:approx-overview}.

\subsection{More discussions on computation with few non-terminals} \label{appendix:small_nonterminal_subset}
We hypothesize that we can focus only on a few non-terminals while retaining most of the performance.

\begin{hypothesis}\label{hyp:small_nonterminal_subset}
     For the PCFG $\gG = (\gN, \gI, \gP, n, p)$ learned on the English corpus, there exists $\tilde\gI\subset\gI,\tilde\gP\subset\gP$ with $|\tilde\gI|\ll |\gI|, |\tilde\gP|\ll |\gP|$, such that simulating Inside-Outside algorithm with $\tilde\gI \cup \tilde\gP$ non-terminals introduces \underline{small} error in the 1-mask perplexity and has \underline{minimal} impact on the parsing performance of the Labeled-Recall algorithm.
\end{hypothesis}

\begin{figure}[!t]
     \centering
    \includegraphics[width=0.8\linewidth]{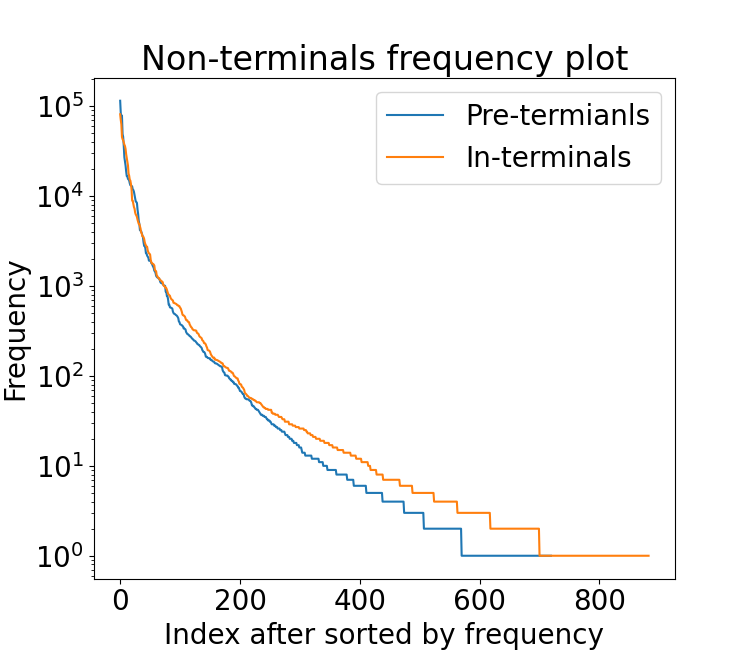}

        \caption{Plot for the frequency distribution of in-terminals ($\gI$) and pre-terminals ($\gP$). We compute the number of times a specific non-terminal appears in a span of a parse tree in the \dataset{PTB} training set. We then sort the non-terminals according to their normalized frequency and then show the frequency vs. index plot.}
        \label{fig:freq-dist}
\end{figure}

To find candidate sets $\tilde\gI,\tilde\gP$ for our hypothesis, we check the frequency of different non-terminals appearing at the head of spans in the parse trees of the \dataset{PTB}~\citep{marcus1993building} training set. We consider the Chomsky-transformed (binarized) parse trees for sentences in the \dataset{PTB} training set, and collect the labeled spans $\{(A, i, j)\}$ from the parse trees of all sentences. For all non-terminals $A$, we compute $\text{freq}(A)$, which denotes the number of times non-terminal $A$ appears at the head of a span. 
\Cref{fig:freq-dist} shows the plot of $\text{freq}(A)$ for in-terminals and pre-terminals, with the order of the non-terminals sorted by the magnitude of $\text{freq}(\cdot)$. We observe that an extremely small subset of non-terminals have high frequency, which allows us to restrict our computation for the inside and outside probabilities to the few top non-terminals sorted by their $\text{freq}$ scores. We select the top frequent non-terminals as possible candidates for forming the set $\tilde\gN$.


We verify the effect of restricting our computation to the frequent non-terminals on the 1-mask perplexity and the unlabeled F1 score of the approximate Inside-Outside algorithm in \Cref{tab:few-nt-pcfg-global}. Recall from \Cref{thm:io-optimal-mlm}, the 1-mask probability distribution for a given sentence $w_1, \cdots, w_L$ at any index $i$ is given by \cref{eq:1mask_pcfg}, and thus we can use \cref{eq:1mask_pcfg} to compute the 1-mask perplexity on the corpus. To measure the impact on 1-mask language modeling, we report the perplexity of the original and the approximate Inside-Outside algorithm on 200 sentences generated from PCFG. 

We observe that restricting the computation to the top-$40$ and $45$ frequent in-terminals and pre-terminals leads to $<6.5\%$ increase in average 1-mask perplexity. 
Furthermore, the Labeled-Recall algorithm observes at most $4.24\%$ drop from the F1 performance of the original PCFG. 
If we further restrict the computation to the top-$20$ and $45$ in-terminals and pre-terminals, we can still get $71.91\%$ sentence F1 score, and the increase in average 1-mask perplexity is less than $8.6\%$. However, restricting the computation to $10$ in-terminals leads to at least $15\%$ drop in parsing performance.

Thus combining \Cref{thm:soft_attnt} and \Cref{tab:few-nt-pcfg-global}, we have the following informal theorem.

\begin{theorem}[Informal]\label{thm:approx-few-nt-informal}
    Given the PCFG $\gG = (\gN, \gI, \gP, n, p)$ learned on the English corpus, there exist  subsets $\tilde\gI\subset\gI,\tilde\gP\subset\gP$ with $|\tilde\gI| = 20, |\tilde\gP| = 45$, and an attention model with soft relative attention modules (\ref{eq:soft_attention}) with embeddings of size $275+40L$, $2L+1$ layers, and $20$ attention heads in each layer, that can simulate the Inside-Outside algorithm restricted to $\tilde\gI,\tilde\gP$
    on all sentences of length at most $L$ generated from $\gG$. The restriction introduces a $9.29\%$ increase in average 1-mask perplexity and $8.71\%$ drop in the parsing performance of the Labeled-Recall algorithm. 
\end{theorem}

If we plug in the average length $L\approx 25$ for sentences in \dataset{PTB}, we can get a model with $20$ attention heads, $1275$ hidden dimension, and $51$ layers. Compared with the construction in \Cref{thm:soft_attnt}, the size of the model is much closer to reality. The proof of \Cref{thm:approx-few-nt-informal} is shown in \Cref{sec:approx-few-nt}.

\subsection{More discussions on low-rank approximation} \label{appendix:approx-low-rank-informal}
We hypothesize that we can  find linear transformation matrices $\{\mW^{(\ell)}\}_{\ell\le L}$ that can reduce the computations while retaining most of the performance, and our hypothesis is formalized as follow:

\begin{hypothesis}
     For the PCFG $\gG = (\gN, \gI, \gP, n, p)$ learned on the English corpus, there exists transformation matrices $\mW^{(\ell)}\in\R^{k^{(\ell)}\times |\tilde\gI|}$ for every $\ell \le L$, such that approximately simulating the Inside-Outside algorithm with $\{\mW^{(\ell)}\}_{\ell\le L}$ introduces \underline{small} error in the 1-mask perplexity and has \underline{minimal} impact on the parsing performance of the Labeled-Recall algorithm.
\end{hypothesis}

\Cref{tab:learned-transformation-global} verifies our hypothesis, and lead to \Cref{thm:approx-low-rank-informal}. Compared with the parsing results from \Cref{thm:approx-few-nt-informal}, the corpus and sentence F1 scores are nearly the same, and we further reduce the number of attention heads in each layer from $20$ to $15$. If we only use $10$ attention heads to approximately execute the Inside-Outside algorithm, we can still get $61.72\%$ corpus F1 and $65.31\%$ sentence F1 on \dataset{PTB} dataset, which is still much better than the Right-branching baseline. \Cref{thm:approx-low-rank-informal} shows that attention models with a size much closer to the real models (like BERT or RoBERTa) still have enough capacity to parse decently well (>70\% sentence F1 on \dataset{PTB}).

It is also worth noting that approximately executing the Inside-Outside algorithm using the transformation matrices $\{\mW^{(\ell)}\}_{\ell\le L}$ is very different from reducing the size of the PCFG grammar, since we use different matrix $\mW^{(\ell)}$ when computing the probabilities for spans with different length. If we choose to learn the same transformation matrix $\mW$ for all the layers $\ell$, the performance drops.

\paragraph{More discussions on the transformation matrix $\mW^{(\ell)}$} We can observe that by introducing the transformation matrix $\mW^{(\ell)}$ generalized the first ingredient that only computes a small set of in-terminals $\tilde\gI$ and pre-terminals $\tilde\gP$, and in theory we can directly learn the transformation matrix $\mW^{(\ell)}$ from the original PCFG without reducing the size at first, i.e., $\mW^{(\ell)}\in\R^{k^{(\ell)}\times |\gI|}$. However empirically, if we directly learn $\mW^{(\ell)}$ from all the in-terminals $\gI$ but not from the top-20 frequent in-terminals $\tilde\gI$, the performance drops. Thus, we choose to learn the matrix $\mW^{(\ell)}$ starting from the most frequent in-terminals $\tilde\gI$. One possible explanation is that the learning procedure is also heuristic, and certainly may not learn the best transformation matrix.

Besides, we use the same transformation matrix $\mW^{(\ell)}$ when computing the inside and outside probabilities, and it is also natural to use different transformation matrices when computing the inside and outside probabilities. Recall that we learn the transformation $\mW^{(\ell)}$ by the Eigenvalue decomposition on matrix $\mX^{(\ell)}$, where $\mX^{(\ell)} = \sum_{s} \mX_s^{(\ell)} / \norm{\mX_s^{(\ell)}}_{\text{F}}$ and $\mX_s^{(\ell)} = \sum_{i,j:j-i=\ell} \vmu_s^{i,j}(\vmu_s^{i,j})^\top$. Then, we can also learn two matrices $\mW^{(\ell)}_{\text{inside}}$ and $\mW^{(\ell)}_{\text{outside}}$ through the Eigenvalue decomposition on matrices $\mX^{(\ell)}_{\text{inside}}$ and $\mX^{(\ell)}_{\text{outside}}$ respectively, where
\begin{align*}
    \mX^{(\ell)}_{\text{inside}} =& \sum_{s} \mX_{s,\text{inside}}^{(\ell)} / \norm{\mX_{s,\text{inside}}^{(\ell)}}_{\text{F}}, \\
    \mX_{s,\text{inside}}^{(\ell)} =& \sum_{i,j:j-i=\ell} \bm\alpha_s^{i,j}(\bm\alpha_s^{i,j})^\top, \\
    \mX^{(\ell)}_{\text{outside}} =& \sum_{s} \mX_{s,\text{outside}}^{(\ell)} / \norm{\mX_{s,\text{outside}}^{(\ell)}}_{\text{F}}, \\
    \mX_{s,\text{outside}}^{(\ell)} =& \sum_{i,j:j-i=\ell} \bm\beta_s^{i,j}(\bm\beta_s^{i,j})^\top.
\end{align*}
However empirically, we also find that the performance drops by using different transformation matrices for inside and outside probabilities computation, which may also be attributed to the non-optimality of our method to learn the transformation matrix.

\subsection{Proof for \cref{thm:approx-few-nt-informal}}\label{sec:approx-few-nt}

Note that in both \Cref{thm:hard_attnt} and \Cref{thm:soft_attnt}, in every layer $1 \le \ell \le L-1$, we use one attention head with parameters $\mK_A^{(\ell)}, \mQ_A^{(\ell)}, \mV_A^{(\ell)}$ to compute all the inside probabilities $\alpha(A, i, j)$ for all spans with length $\ell+1$, i.e. $j-i = \ell$. For layer $L+1 \le \ell \le 2L-1$, the model constructed in \Cref{thm:hard_attnt} uses two attention heads to compute the outside probabilities $\beta(A,i,j)$ for a specific non-terminal $A$ for spans with length $2L - \ell$, and the model constructed in \Cref{thm:soft_attnt} uses one attention heads to compute the outside probabilities $\beta(A,i,j)$ for a specific non-terminal $A$ for spans with length $2L - \ell$. Now to show how restricting the computations to certain non-terminals $\tilde\gI\cup\tilde\gP$ can reduce the size of the constructed models in \Cref{thm:hard_attnt,thm:soft_attnt} we classify the inside and outside probabilities into four categories: (1) the inside probabilities for pre-terminals, $\alpha(A,i,i)$ for $A\in\gP$; (2) the inside probabilities for in-terminals, $\alpha(A,i,j)$ for $A\in\gI$; (3) the outside probabilities for in-terminals, $\beta(A,i,j)$ for $A\in\gI$; and (4) the outside probabilities for pre-terminals, $\beta(A,i,i)$ for $A\in \gP$.

\paragraph{Category (1): the inside probabilities for pre-terminals} Recall that in the constructed model in \Cref{thm:hard_attnt,thm:soft_attnt}, the inside probabilities for pre-terminals $\alpha(A,i,i)$ for $A\in\gP$ is directly initialized from the PCFG rules, and thus do not need attention heads to compute. Thus, we can just use $O(|\gP|)$ dimensions to store all the inside probabilities for pre-terminals $\alpha(A,i,i)$ for $A\in\gP$. Although we can also only initialize the inside probabilities only for the pre-terminals $\tilde\gP$, i.e. initialize $\alpha(A,i,i)$ for $A\in\tilde\gP$ and use less embedding dimensions, empirically the performance will drop and thus we initialize all the probabilities $\alpha(A,i,i)$ for $A\in\gP$. Although we should store the probabilities for pre-terminals larger than the set $\tilde\gP$, there is indeed another technique to reduce the embedding dimensions. Note that since in the future computations, we only compute the probabilities for the in-terminals $\tilde\gI$, and not every pre-terminal $A\in\gP$ can be produced by in-terminals $B\in\tilde\gI$. Thus, we only need to store the pre-terminals $\gP_{\tilde\gI}$ that can be produced from $\tilde\gI$. Empirically, for PCFG learned on \dataset{PTB} dataset, $|\gP| = 720$, but if we choose $|\tilde\gI| = 20$, the number of pre-terminals that can be produced from $\tilde\gI$ drops to $|\gP_{\tilde\gI}| = 268 < 270$.
Specifically for the model in \Cref{thm:soft_attnt}, we need $|\gP_{\tilde\gI}|$ coordinates at each position to store these inside probabilities.

\paragraph{Category (2): the inside probabilities for in-terminals} The computation of the inside probabilities for in-terminals, $\alpha(A,i,j)$ for $A\in\gI$ happens from layer $1$ to layer $L-1$ in the constructed model in \Cref{thm:hard_attnt,thm:soft_attnt}. Note that from layer $1$ to layer $L-1$, the model only computes the probabilities for the in-terminals, since a span with a length larger than 1 cannot be labeled by a pre-terminal. Thus, if we only compute the inside probabilities for in-terminals $|\tilde\gI|$, we can reduce the number of attention heads in layer $1$ to layer $L-1$ from $O(|\gI|)$ to $O(|\tilde\gI|)$ since in \Cref{thm:hard_attnt,thm:soft_attnt} we use a constant number of attention heads to compute the probabilities for a single in-terminal. Specifically for the model in \Cref{thm:soft_attnt}, we only need $|\tilde\gI|$ attention heads from layer $1$ to layer $L-1$; besides, we need $(L-1)|\tilde\gI|$ coordinates at each position to store these inside probabilities.

\paragraph{Category (3): the outside probabilities for in-terminals} The computation of the outside probabilities for in-terminals, $\beta(A,i,j)$ for $A\in\gI$ happens from layer $L$ to layer $L-2$ in the constructed model in \Cref{thm:hard_attnt,thm:soft_attnt}. Note that in layer $L$, we only need to initialize the outside probabilities $\beta(A,1,L)$ for $A\in\gI$, thus do not need attention heads for computation (however we need attention heads to move the inside and outside probabilities in this layer, which cost 2 attention heads). Then from layer $L+1$ to layer $L-2$, the model computes the outside probabilities for the in-terminals $\beta(A,i,j)$ for $A\in\tilde\gI$. Thus if we only compute the outside probabilities for in-terminals $|\tilde\gI|$, we can reduce the number of attention heads in layer $1$ to layer $L-1$ from $O(|\gI|)$ to $O(|\tilde\gI|)$. Specifically for the model in \Cref{thm:soft_attnt}, we only need $|\tilde\gI|$ attention heads from layer $L$ to layer $L-2$; besides, we need $(L-1)|\tilde\gI|$ coordinates at each position to store these outside probabilities for in-terminals $\tilde\gI$.

\paragraph{Category (4): the outside probabilities for pre-terminals} The outside probabilities for pre-terminals $\beta(A,i,i)$ for $A\in\gP$ is only computed in the final layer in \Cref{thm:hard_attnt,thm:soft_attnt}. Thus if we choose to compute the probabilities for only $\tilde\gP$, we can reduce the number of attention heads in layer $2L-1$ from $O(|\gI|)$ to $O(|\tilde\gI|)$. Specifically for the model in \Cref{thm:soft_attnt}, we only need $|\tilde\gP|$ attention heads in layer $L-1$; besides, we need $|\tilde\gP|$ coordinates at each position to store these outside probabilities for in-terminals $\tilde\gP$. Also as mentioned in \Cref{sec:approx-overview}, if $|\tilde\gP| < c|\tilde\gI|$ for some constant $c$, we can also simulate the computations in the last layer with $|\tilde\gP|$ heads by $c$ layers with $|\tilde\gI|$ heads. In particular, if we choose $|\tilde\gP| = 45, |\tilde\gI|=20$, we can use 3 layers with $20$ attention heads in each layer to simulate the last layer with $45$ attention heads in the original construction.

\paragraph{Put everything together: proof of \Cref{thm:approx-few-nt-informal}} We choose $|\tilde\gP| = 45, |\tilde\gI|=20$. We can use $20$ attention heads in each layer, and we now count the number of layers and the embedding dimension we need. The number of layers is easy to compute, since we just need to use $3$ layers with $20$ attention heads to simulate the original $1$ layer with $45$ attention heads, thus the total number of layers is $2L-1 + (3-1) = 2L+1$. As for the embedding dimension, we need
\begin{align*}
    d = & |\gP_{\tilde\gI}| + (L-1)|\tilde\gI| + (L-1)|\tilde\gI| + |\tilde\gP| \\
    \le & 270 + (2L-2)|\tilde\gI| + |\tilde\gP| \\
    =& 275 + 2L|\tilde\gI|\\
    =& 275 + 40L.
\end{align*}



\subsection{Proof for \cref{thm:approx-low-rank-informal}}\label{sec:approx-low-rank}

In this section, we show the details of how to further reduce the number of attention heads using structures across non-terminals, and add more discussion on how we learn the transformation matrices $\{\mW^{(\ell)}\}_{\ell \le L}$

\paragraph{Reducing the number of attention heads} We focus on reducing the number of attention heads to compute the inside and outside probabilities for the in-terminals $\tilde\gI$, since the computation for the outside probabilities for pre-terminals $\tilde\gP$ only happens in the final layer of the constructed model, and thus can use multiple layers to compute as long as $\tilde\gP$ is not too large.

For simplicity, we only show the details of how to reduce the number of attention heads to compute the inside probabilities for in-terminals $\tilde\gI$ in \Cref{thm:soft_attnt}, and the technique can be easily applied to the computation of outside probabilities for in-terminals $\tilde\gI$ in \Cref{thm:soft_attnt}, and the inside and outside probabilities for $\tilde\gI$ in \Cref{thm:hard_attnt}.

Recall from the proof of \Cref{thm:soft_attnt} that we at each layer $\ell$, we use a single attention head $\mK^{(\ell)}_A, \mQ^{(\ell)}_A$ to compute the inside probabilities $\alpha(A,i,j)$ for spans with length $\ell+1$, i.e., $j-i = \ell$. Specifically, for the attention head $\mK^{(\ell)}_A, \mQ^{(\ell)}_A$ at layer $\ell$, we want to compute and store the probability $\alpha(A, i-\ell, i)$ at position $i$. Thus we construct $\mK^{(\ell)}_A, \mQ^{(\ell)}_A$ such that the attention score $a_{i,j}^{A,(\ell)}$ when the position $i$ attends to position $j$ satisfies

{\small
\begin{align*}
     &a_{i,j}^{A,(\ell)} \\
     =& \text{ReLU}(\mK_{A}^{(\ell)} \ve_j^{(\ell-1)} + p_{j-i} - b_{j-i, \ell})^{\top}  \mQ_{A}^{(\ell)} \ve_i^{(\ell-1)} \\
     =& \sum_{B, C \in \gN} \Pr[A \to B C] \cdot \alpha(B, j+1, i) \cdot  \alpha (C, i-\ell, j),
\end{align*}
}

if $i - \ell \le j \le i-1$ and $0$ otherwise. Then, summing over all locations $j$ gives us $\alpha(A, i-\ell, i)$. Also, a key property of $\mK_{A}^{(\ell)}$ is that this key matrix does not depend on the non-terminal $A$, but only depends on $\ell$. Thus, if we have a set of coefficients $\{\omega_A^{(\ell)}\}_{A\in\gI}$, we can compute the linear combination of the inside probability $\sum_{A\in\tilde\gI} \omega_A^{(\ell)}\alpha(A, i-\ell, i)$ using one attention head, since if we choose
\[\mQ^{(\ell)} = \sum_{A\in\tilde\gI} \omega_A^{(\ell)}\mQ_{A}^{(\ell)}, \quad\mK^{(\ell)} = \mK_A^{(\ell)}, \forall A\in\tilde\gI,\]
we have the attention score

{\small
\begin{align*}
     &a_{i,j}^{(\ell)} \\
     = &\text{ReLU}(\mK^{(\ell)} \ve_j^{(\ell-1)} + p_{j-i} - b_{j-i, \ell})^{\top}  \mQ^{(\ell)} \ve_i^{(\ell-1)} \\
     = &\text{ReLU}(\mK^{(\ell)} \ve_j^{(\ell-1)} + p_{j-i} - b_{j-i, \ell})^{\top} \\
     &\cdot \left(\sum_{A\in\tilde\gI} \omega_A^{(\ell)}\mQ_{A}^{(\ell)}\right) \ve_i^{(\ell-1)} \\
     = &\sum_{A\in\tilde\gI} \omega_A^{(\ell)}\\
     &\cdot \text{ReLU}(\mK^{(\ell)} \ve_j^{(\ell-1)} + p_{j-i} - b_{j-i, \ell})^{\top}  \mQ_{A}^{(\ell)} \ve_i^{(\ell-1)} \\
     = &\sum_{A\in\tilde\gI} \omega_A^{(\ell)}\\
     &\cdot \text{ReLU}(\mK_A^{(\ell)} \ve_j^{(\ell-1)} + p_{j-i} - b_{j-i, \ell})^{\top}  \mQ_{A}^{(\ell)} \ve_i^{(\ell-1)} \\
     = &\sum_{A\in\tilde\gI} \omega_A^{(\ell)}\\
     &\cdot\left(\sum_{B, C \in \gN} \Pr[A \to B C] \cdot \alpha(B, j+1, i) \cdot  \alpha (C, i-\ell, j)\right),
\end{align*}
}
if $i - \ell \le j \le i-1$ and $0$ otherwise. Then, summing over all locations $j$ gives us $\sum_{A\in\tilde\gI} \omega_A^{(\ell)}\alpha(A, i-\ell, i)$. Then if we have a transformation matrix $\mW^{(\ell)}\in\R^{k^{(\ell)}\times |\tilde\gI|}$, we can use $k^{(\ell)}$ attention heads to compute $\mW^{(\ell)}\bm\alpha(i-\ell, i)$, where $\bm\alpha(i-\ell, i)\in\R^{|\tilde\gI|}$ is the vector that contains $\alpha(A, i-\ell, i)$ for all $A\in\tilde\gI$. Then after we use $k^{(\ell)}$ attention heads to compute the probabilities $\mW^{(\ell)}\bm\alpha(i-\ell, i)$ and stored them in position $i$'s embeddings, we can then use linear layer on position $i$ to recover the original probabilities by $\bm{\tilde\alpha}(i-\ell, i) = (\mW^{(\ell)})^{\dagger} \mW^{(\ell)}\bm\alpha(i-\ell, i)$, and use $\tilde\alpha(A, i-\ell, i)$ for $A\in\tilde\gI$ for the future computations.

\paragraph{Put everything together: proof of \Cref{thm:approx-low-rank-informal}} We choose $k^{(\ell)} = 15,|\tilde\gP| = 45, |\tilde\gI|=20$. Note that the embedding dimension doesn't change if we apply the approximation technique, and only the number of attention heads reduces from $20$ to $15$. Thus, the embedding dimension is still
\begin{align*}
    d =& |\gP_{\tilde\gI}| + (L-1)|\tilde\gI| + (L-1)|\tilde\gI| + |\tilde\gP| \\
    \le& 270 + (2L-2)|\tilde\gI| + |\tilde\gP|\\
    =& 275 + 2L|\tilde\gI|\\
    =& 275 + 40L.
\end{align*}
Also note that $|\tilde\gP| = 45 = 3\times 15$, and thus we can compute all the outside probabilities for pre-terminals $\tilde\gP$ by $3$ layers where each layer has $15$ attention heads.

\subsection{Experiment details in \Cref{sec:approx-overview}}\label{sec:approx-exp-details}
In this section, we provide the experiment details in \Cref{sec:approx-overview}. We use and modify the code~\citep{Spectral-Parser} to learn the PCFG from the \dataset{PTB} dataset and conduct the experiments with approximated computations. \citet{Spectral-Parser} implements the spectral learning method to learn PCFG~\citep{cohen2012spectral,cohen2014spectral} and is under MIT licence. We follow all the default hyperparameters in \citet{Spectral-Parser}, and we also follow the split of \dataset{PTB}: using \dataset{PTB} section 02-21 as the training set and \dataset{PTB} section 22 as the development set.


\end{document}